\let\latexdocument\document
\let\latexenddocument\enddocument

\documentclass{clv3} 

\let\document\latexdocument
\let\enddocument\latexenddocument

\makeatletter
\AtBeginDocument{%
  \if@filesw
    \immediate\openout\@mainqry=\jobname.qry
  \fi
}
\AtEndDocument{%
   \ifx\@biography\@empty\else{\par\ifbrief\vskip10pt\fi\biofont\noindent\@biography\par}\fi
   \immediate\closeout\@mainqry
      \ifodd\c@page\clearpage\thispagestyle{empty}\null\clearpage\else\clearpage\fi
      
}
\makeatother

\NewCommandCopy{\cnumdef}{\numdef}
\NewCommandCopy{\endcnumdef}{\endnumdef}
\let\numdef\relax \let\endnumdef\relax

\usepackage{xcolor}
\usepackage{latexsym}
\usepackage{amsmath,amssymb}
\usepackage{txfonts}
\usepackage[T1]{fontenc}
\usepackage[utf8]{inputenc}
\usepackage{yfonts}
\usepackage{inconsolata}
\usepackage{multirow}
\usepackage{makecell}
\usepackage{graphicx}
\usepackage{pgf}
\usepackage{booktabs}
\usepackage{colortbl}
\usepackage{microtype}
\usepackage{xfp}
\usepackage{twoopt}
\usepackage{xspace}
\usepackage{verbatim}
\usepackage[size=tiny]{todonotes}
\usepackage{subcaption}
\usepackage{tabularx, booktabs}
\usepackage{floatrow}
\usepackage{hyperref}
\usepackage{soul}
\usepackage{graphics}

\newcommand{\newcrossmark}{\scalebox{0.85}[1]{$\times$}}

\newcommand{\zrf}[1]{\textcolor{black}{#1}}

\definecolor{darkblue}{rgb}{0, 0, 0.5}
\hypersetup{colorlinks=true,citecolor=darkblue, linkcolor=darkblue, urlcolor=darkblue}

\usepackage[normalem]{ulem}

\begin{document}
\pageonefooter{Action editor: Mohit Bansal. Submission received: 2023-07-13; revised version received: 2024-02-02; accepted for publication: 2024-04-30}

\runningtitle{Cross-lingual Cross-temporal Summarization:\\ Dataset, Models, Evaluation}

\runningauthor{}

\title{Cross-lingual Cross-temporal Summarization:\\ Dataset, Models, Evaluation}

\author{Ran Zhang \thanks{School of Business Informatics and Mathematics, B6 26, 68159 Mannheim, Germany. E-mail: ran.zhang@uni-mannheim.de.}}
\affil{Natural Language Learning Group (NLLG),  School of Business Informatics and Mathematics, University of Mannheim}

\author{Jihed Ouni \thanks{Technische Universität Darmstadt, Hochschulstraße 10, Darmstadt 64289, Germany. E-mail: jihed.ouni@stud.tu-darmstadt.de}}
\affil{Fachbereich Informatik, Technische Universität Darmstadt}

\author{ Steffen Eger \thanks{School of Business Informatics and Mathematics, B6 26, 68159 Mannheim, Germany. E-mail: steffen.eger@uni-mannheim.de}}
\affil{Natural Language Learning Group (NLLG), School of Business Informatics and Mathematics, University of Mannheim}
\maketitle

\begin{abstract}

While summarization has been extensively researched in natural language processing (NLP), cross-lingual cross-temporal summarization (CLCTS) is a largely unexplored area that has the potential to improve cross-cultural accessibility and understanding. This paper comprehensively addresses the CLCTS task, including dataset creation, modeling, and evaluation. We (1) build the first CLCTS corpus with 328 instances for hDe-En \zrf{(extended version with 455 instances)} and 289 for hEn-De \zrf{(extended version with 501 instances)}, leveraging historical fiction texts and Wikipedia summaries in English and German; (2) examine the effectiveness of popular transformer end-to-end models with different intermediate finetuning tasks; (3) explore the potential of GPT-3.5 as a summarizer; (4) report evaluations from humans, GPT-4, and several recent automatic evaluation metrics. Our results indicate that intermediate task finetuned end-to-end models generate bad to moderate quality summaries while GPT-3.5, as a zero-shot summarizer, provides moderate to good quality outputs. GPT-3.5 also seems very adept at normalizing historical text. To assess data contamination in GPT-3.5, we design an adversarial attack scheme in which we find that GPT-3.5 performs slightly worse for unseen source documents compared to seen documents. Moreover, it sometimes hallucinates when the source sentences are inverted against its prior knowledge with a summarization accuracy of 0.67 for plot omission, 0.71 for entity swap, and 0.53 for plot negation. Overall, our regression results of model performances suggest that longer, older, and more complex source texts (all of which are more characteristic for historical language variants) are harder to summarize for all models, indicating the difficulty of the CLCTS task. Regarding evaluation, we observe that both GPT-4 and BERTScore correlate moderately with human evaluations, \zrf{implicating great potential for future improvement.}  

\end{abstract}

\section{Introduction}
\label{sec:introduction}

Summarization is a key task in natural language processing (NLP), especially in an age of information overload. The classical approach for summarization is to summarize documents (e.g., news reports) in the same language \cite{zhang2020pegasus, liu-etal-2022-brio, ravaut-etal-2022-summareranker} or, less prominently, in different languages than the source (\emph{cross-lingual summarization}; CLS) \cite{ladhak-etal-2020-wikilingua, cao-etal-2020-jointly, bai-etal-2021-cross, liang-etal-2022-variational}.
Tasks such as the summarization of historical texts to modern languages (\emph{cross-temporal summarization}; CTS) are scarcely explored, in contrast. Such summarizers are beneficial not only for historical researchers and students but also for laypeople with an interest in historical information sources. 

In this work, we go one step further and explore \emph{cross-lingual cross-temporal summarization (CLCTS)}, in which the goal is to summarize a historical document in a different modern language. See Table \ref{data:example} for an example. Such cross-lingual and cross-temporal summarizers would potentially tremendously facilitate cross-cultural accessibility, information sharing, and understanding. CLCTS is a valuable area of research that expands upon the recently popular task of literary translation with summarization. Both tasks share cultural importance and present unique complexities due to the intricate nature of the creative work involved \cite{karpinska-iyyer-2023-large, thai-etal-2022-exploring}. CLCTS is not limited to literary works alone: it can also assist in summarizing non-literature such as historical newspapers, reports, textbooks, etc. For example, CLCTS models can assist in creating (missing) Wikipedia pages for historical works. While the translation of literature may require language that is faithful to the original source for a `close reader', CLCTS could potentially be better suited for a `distant reader', and thus may require less sophisticated output text. Moreover, modern summaries can facilitate the classification of literary works, for example, genre classification \cite{saraswat2022leveraging, agarwal2021genre}. Firstly, summaries are more condensed than (historical) source texts and, therefore, are much easier to model. Secondly, modern texts are better suited for models that operate with modern data such as SBERT \cite{reimers-2020-multilingual-sentence-bert}.

\begin{table}
\fontsize{8pt}{8pt}\selectfont
    \begin{tabular}{p{0.15\textwidth}p{0.75\textwidth}}
    \toprule
    \textbf{Title} & An Inhabitant of Carcosa \\
    \textbf{Author} & Ambrose Bierce \\
    \textbf{Year} & 1886 \\
    \textbf{Source} & The Collected Works of Ambrose Bierce, Volume 3 \\
    \midrule
    \textbf{Text} & ``For there be divers sorts of death—some wherein the body \textit{remaineth}; and in some it \textit{vanisheth} quite away with the spirit ... In one kind of death the spirit also \textit{dieth}, and this it \textit{hath} been known to do while yet the body was in vigor for many years. [...]'' \\
    \midrule
    \textbf{Summary (German)} & ``Ein Mann aus der Stadt Carcosa, der über die Worte des Philosophen Hali über die Natur des Todes nachdenkt, wandert durch eine ihm unbekannte Wildnis. [...]'' \\
    \bottomrule
    \end{tabular}
    \caption{Example from our cross-lingual cross-temporal dataset (historical English to modern German). We highlight words with spelling/morphological changes in italics.}
    \label{data:example}
\end{table}

CLCTS is difficult and challenging for the following reasons: (i) languages change over time along multiple dimensions, including syntax \cite{juzek-etal-2020-exploring, lei_is_2020}, semantics \cite{hamilton-etal-2016-diachronic, giulianelli-etal-2020-analysing}, lexical choice and morphology \cite{gibson_how_2019, joseph2017diachronic}. (ii) Historical documents are often longer, not only because of tendencies of simplification/text length reduction over time \cite{lewis1894history, sherman1893analytics, rudnicka2018variation}, but also because of the genre of historical text, which predominantly includes literary work \cite{wang2017effects, zhu2022investigating}. (iii) The process can be thought of as consisting of several subtasks, previously considered independently in NLP research: machine translation (MT), summarization, and historical (spelling) normalization \cite{bollmann-sogaard-2016-improving, eger2016comparison}. 

In this work, we consider the CLCTS problem comprehensively, including dataset creation, modeling, and evaluation.\footnote{Our code and data are available at \url{https://github.com/zhangr2021/CLCTS}.}
We start with building the first CLCTS dataset. Our corpus contains stories and plays (as prime artefacts of historical cultures), which requires the models to handle long documents; we assemble modern summaries by querying Wikipedia.

In terms of modeling, we consider three approaches: 
(1) extractive summarizers \cite{gu-etal-2022-memsum} 
whose advantage is that they can in principle deal with unlimited input document lengths but which are limited in that they can only copy out sentences from the original source (which is a problem, e.g., when the source is first-person narrative but the summary should be third-person perspective). 
(2) Abstractive summarizers can avoid the issue above but typically have severe limitations in input lengths. Even models exclusively designed for long document summarization by utilizing efficient variants of attention mechanism such as Longformer Encoder-Decoder (LED) \cite{beltagy2020longformer} and Bigbird \cite{zaheer2020bigbird} limit the max input tokens to 16,384 and 4,096, respectively. 
(3) GPT-3 \cite{brown2020language} and its extensions such as ChatGPT are also strong candidates for our task \cite{ goyal_news_2022} as they have so spectacularly impacted research in a multitude of scenarios \cite{Leiter2023ChatGPTAM}, including summarization \cite{yang_exploring_2023, wang_cross-lingual_2023, zhang2023extractivechatgpt}.
Consequently, we also include ChatGPT \zrf{(GPT-3.5 and GPT-4)} in our analysis.

In terms of evaluation, we consider three evaluation methods: (1) automatic evaluation, where we report multiple evaluation metrics additional to the commonly used variants of ROUGE \cite{lin-2004-rouge} in summarization tasks \cite{liang-etal-2022-variational, bai-etal-2021-cross, cao-etal-2020-jointly, 2021summarising} since the performance of individual metrics may vary across datasets, challenging their reliability \cite{bhandari-etal-2020-evaluating, fabbri2021summeval}; 
(2) human evaluation, which reflects the actual quality of summaries according to human judgments and functions as a source of reliability measurement for automatic evaluation metrics;  
(3) ChatGPT evaluation, where we examine the potential of ChatGPT as an alternative to human annotators given the same instruction considering the high cost of human evaluation \cite{gao2023human} and issues of reproducibility \cite{belz-etal-2021-reprogen,belz-etal-2023-missing,chen-etal-2022-reproducibility}.    
Our contributions are:
\begin{itemize}
    \item To our best knowledge, we build the first CLCTS corpus, leveraging historical ficti\zrf{on} texts and Wikipedia summaries.
    \item We examine the effectiveness of popular transformer end-to-end (e2e) models with different tasks for intermediate task finetuning \cite{chang-lu-2021-rethinking-intermediate}.
    \item We explore how ChatGPT performs for CLCTS.
    \item We provide human, \zrf{GPT-4}, and automatic evaluation for the task, showing that our intermediate task finetuned e2e models generate bad to moderate quality summaries; \zrf{GPT-3.5} as a summarizer provides moderate to good quality outputs (without any finetuning); \zrf{GPT-4} as an evaluator correlates with human evaluations to a moderate level. 
\end{itemize}

\section{Related work}

\textbf{Historical text normalization} Historical text normalization, the process of mapping non-standard word tokens to their modern standard forms, is an important technique for analyzing historical texts. Model designs include (1) corpus-based lexical substitution as an effective component for normalization systems \cite{rayson2005vard, baron2008vard2, bollmann2012automatic}; (2) rule-based models, \zrf{such as} phonological rules \cite{porta2013edit} and edit distance measures \cite{adesam2012bokstaffua, bollmann2012automatic}; (3) statistical models which aim to maximize the probability of contemporary word form given its historical form \cite{pettersson2016spelling}; (4) neural network (NN) models which utilize NN architectures such as encoder-decoder LSTMs \cite{bollmann2017learning, robertson2018evaluating} or RNNs \cite{makarov-clematide-2020-semi}. Worth noting is that nearly all existing historical normalizers operate on words in isolation, except for the semi-supervised contextual normalizer from \citet{makarov-clematide-2020-semi}. In our study, we find that ChatGPT \zrf{(GPT-3.5)} has the potential to serve as a (much better) context-aware historical normalizer.  \newline

\noindent\textbf{Cross-temporal natural language generation (NLG)} Especially CTS is an underexplored topic in NLP. \namecite{2021summarising} first introduce the task of historical text summarization (CTS).\footnote{In their study, they regard such historical text summarization as a special case of cross-lingual summarization (CLS). We use the term cross-temporal summarization (CTS) to avoid confusion and highlight the differences between CLS and CTS tasks.} 
They point out the scarcity of CTS datasets and build the HISTSUMM dataset containing 100 historical German and Chinese news documents with reference summaries written by linguistic experts in both languages. Our work differs along several dimensions from \namecite{2021summarising}, we mostly explore e2e models, work on very different datasets, and consider transformation across both time and language --- for example, CLCTS instead of only CTS --- and provide a much more comprehensive evaluation. A related cross-temporal task in NLG is literary machine translation (e.g., \namecite{thai-etal-2022-exploring} build a large multilingual paragraph-level literary machine translation dataset PAR3 including paragraphs from historical novels and short stories with multiple human-written English translations and machine translations from Google \zrf{Translate}). 
Since we focus on various summarizations instead of translations and long documents instead of paragraphs, the translation dataset is not within the scope of our current study. \newline

\noindent \textbf{Cross-lingual summarization} 
One natural solution for CLS task is to utilize a pipeline framework following a translate-then-summarize \cite{yao2015phrase, ouyang-etal-2019-robust} or summarize-then-translate \cite{wan2010cross} paradigm. To tackle the problem of error propagation from pipeline methods, \namecite{zhu-etal-2019-ncls} propose an e2e CLS framework where they incorporate monolingual summarization (MLS) and machine translation (MT) simultaneously into the CLS training process. Following this work, many researchers use related tasks additional to the target task either simultaneously (multi-task learning) or consecutively (intermediate task finetuning) to train e2e CLS summarizers \cite{weller-etal-2022-use}. \namecite{takase-okazaki-2022-multi} combine translation and MLS datasets simultaneously in training a single encoder-decoder model and additionally use a special token in the input (e.g., <Trans> and <Summary>) to define the task. \namecite{bai2022unifying} also use a multi-task learning framework by interpreting MT as a special CLS task where no source content should be discarded. \namecite{ladhak-etal-2020-wikilingua} experiment with two-step finetuning where the model is first finetuned for MT and then CLS and this framework delivers better results than pipeline approaches. Similarly, \namecite{he2023zcode} incorporate intermediate task finetuning with multiple summarization datasets before finetuning on the target task. Considering the effectiveness of intermediate task finetuning especially in low-resource settings \cite{chang-lu-2021-rethinking-intermediate}, we train our e2e models using the same approach by first finetuning with different intermediate tasks such as MLS or CLS before our target CLCTS task. \newline 

\noindent\textbf{Long document summarization} is another area of our concern. 
Current approaches for long document summarization can be classified into three categories, namely extractive \cite{cui-hu-2021-sliding}, abstractive \cite{beltagy2020longformer} and hybrid summarization approaches \cite{pilault-etal-2020-extractiveabs}. Among the proposed models, BART \cite{lewis-etal-2020-bart} and PEGASUS \cite{zhang2020pegasus} are most popular transformer-based pretrained models used for long document summarization for both supervised abstractive \cite{zaheer2020bigbird, huang-etal-2021-efficient} and hybrid models \cite{hybrid-longdoc-divide, manakul-gales-2021-long}. As pointed out by \namecite{koh2022empirical}, abstractive finetuned models built with a combination of pretrained large language models (LLMs) and efficient attention mechanisms are among the most competitive approaches for long document summarization, for example, BART with LED attention (input limits 16,384) \cite{beltagy2020longformer} and PEGASUS with BigBird attention (4,096).\footnote{The very recently proposed retrieval-based approach \zrf{U}nlimiformer \cite{bertsch2023unlimiformer} extends the input limitations of the aforementioned pretrained encoder-decoder transformers to practically unlimited input sequences at test time by offloading the cross-attention computation. } 
We build our model with mBART \cite{tang2020multilingual} pretrained for multilingual MT tasks and combine with LED attention to enable efficient processing for long sequences. Additionally, we build another retrieve-then-summarize pipeline where we retrieve sentences with an extractive summarizer \cite{gu-etal-2022-memsum} and then summarize cross-lingually using ChatGPT. \newline

\noindent\textbf{ChatGPT for summarization} Despite ChatGPT's young age, there is already a wealth of research that explores it for various NLP tasks, including summarization \cite{yang_exploring_2023} and summarization evaluation \cite{chiang2023large, gao2023human}. Disagreement exists concerning ChatGPT's performance. 
Most studies focus on  MLS tasks. 
\namecite{soni2023comparing} find that humans are unable to distinguish between summaries written by humans and those produced by ChatGPT. However, ChatGPT as an extractive summarizer is inferior to existing supervised systems according to ROUGE 
\cite{zhang2023extractivechatgpt}. \namecite{bang2023multitask} also find that the finetuned BART outperforms zero-shot ChatGPT by a large margin for MLS and point out that ChatGPT, like other LLMs, tends to generate hallucinated information beyond the given knowledge (extrinsic hallucination). 
As for CLS, \namecite{wang_cross-lingual_2023} examine the ability of ChatGPT with prompt engineering and claim that it outperforms other models on the task.
They also claim that ChatGPT can perform better with pipeline-like prompt lines
in an interactive manner. It is also worth noticing that all works mentioned above utilize small samples of 50 to 100 random samples (out of 3,000 to 60k observations) from each dataset in their experiments. \newline
 
\noindent\textbf{Summarization evaluation metrics} Depending on the availability of reference summaries, one can utilize \emph{reference-based} metrics (focusing on overlap measure \cite{lin-2004-rouge, popovic-2015-chrf}, embedding similarity \cite{bert-score, zhao2019moverscore}, discourse coherence \cite{zhao-etal-2023-discoscore}, natural language inference (NLI) \cite{chen2022menli} and text generation task such as question answering (QA) \cite{NEURIPS2021_e4d2b6e6, wang-etal-2020-asking}) or \emph{reference-free} metrics \cite{gao-etal-2020-supert, chen-etal-2021-training, liu-etal-2022-reference, laban-etal-2022-summac, belouadi-eger-2023-uscore}. 
In terms of summarization evaluation for ChatGPT or other variants of GPT, the above-mentioned ChatGPT summarization studies all choose ROUGE and yet \namecite{peyrard-2019-studying} point out the low correlation between ROUGE and human judgments when evaluating high-scoring summaries. Furthermore, the study from \namecite{goyal_news_2022} show disagreements between humans and 16 evaluation metrics \footnote{Reference-based metrics: ROUGE variants, METEOR, BLEU, BERTScore, MoverScore, QAEval variants. Reference-free metrics: SUPERT, BLANC, QuestEval, QAFactEval, FactCC, DAE, SummaC.} when ranking prompt-based GPT-3 and finetuned models. To address this concern, we leverage multiple automatic evaluation metrics, especially more recent ones (e.g., MENLI and DiscoScore), as well as human evaluation. Since we focus on long documents, we do not explore reference-free metrics, due to the vast mismatch in length between source texts and summaries.

As for recent developments in evaluation with LLMs, the very recently proposed metric G-Eval \cite{liu2023geval} develops a reference-free evaluation framework based on GPT-4 utilizing both chain-of-thought and a form-filling paradigm. Other studies explore simpler scenarios where LLMs such as ChatGPT are queried with prompts containing task instructions as inputs and directly output the evaluation results. \namecite{shen2023large} 
find that LLM’s evaluation capability depends on the evaluated dimensions where ChatGPT is more effective at evaluating consistency. 
\namecite{gao2023human} conduct experiments with the summarization evaluation datasets SummEval \cite{fabbri2021summeval} and Newsroom \cite{grusky-etal-2018-newsroom} and find that ChatGPT outperforms automatic 
evaluation metrics (ROUGE, BERTScore, BARTScore, MoverScore) on one of the datasets (SummEval). \namecite{chiang2023large} provide more comprehensive experiments with recent LLMs including ChatGPT and find that ChatGPT can not only rate like human experts but also provide explanations for its own decisions. However, they also point out that ChatGPT evaluation is prone to \zrf{giving} lower scores. In our work, we also include \zrf{GPT-4} as an evaluator where we query evaluation directly via \zrf{GPT-4}.
\section{Dataset}
\label{sec:dataset}
In this section, we introduce our CLCTS corpus and additional sources we utilize for our experiments. We build our own corpus which covers English and German in both directions. We utilize multiple monolingual and cross-lingual summarization datasets as additional resources during our intermediate task finetuning process. 

\subsection{CLCTS Corpus}  
Manually creating summaries is very time-consuming and requires expert knowledge \cite{2021summarising}. Thus, we use fairy tales, short stories, and plays where summaries are available in Wikipedia articles. In each language direction, English and German, the dataset contains summary pairs from \textbf{historical documents} in one language to \textbf{modern summaries} in the other language. \newline

\begin{figure}
\includegraphics[width=0.45\textwidth]{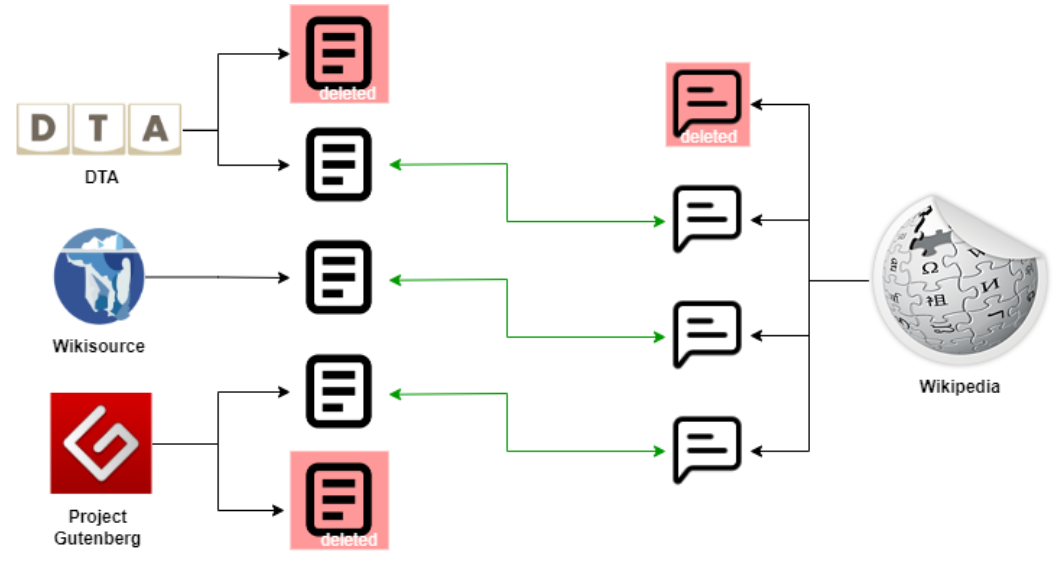}
\caption{Texts matching from DTA, Wikisource, and Project Gutenberg to summaries collected from Wikipedia. Non-paired texts are excluded (highlighted in red).}
\label{ex_collection}
\end{figure}

\noindent \textbf{Dataset collection}
We collect historical documents mainly from three different sources.\footnote{\textit{Wikisource} and \textit{Project Gutenberg} contain both German and English sources, while \textit{Deutsches Textarchiv} contains German sources only.} The links to the sources are collected in Section \ref{apdx:links} (appendix). 
\begin{itemize}
    \item \emph{Deutsches Textarchiv} (DTA; German text archive): This is a basic stock of German-language texts focusing on the early 16th to early 20th centuries. 
    \item \emph{Wikisource}: This is a multilingual online digital library of free-textual content containing historical books in text format and original format stored as images. Some characters are changed from the original historical format to modern digital formats. Proofreading is performed for the texts by users on Wikisource following the proofreading guide.  
    \item \emph{Project Gutenberg}: This is a cultural digital archive with over 60,000 digitized books, including historical texts. It focuses on older works with expired U.S.\ copyrights.
\end{itemize}
Our workflow for building the corpus consists of two stages:  (1) meta-information collection \& historical text extraction
and (2) summary collection \& translation.
In stage (1), we collect meta-information of historical documents using the ``Beautifulsoup'' library.
We extract information such as titles, web links, authors, years, and documents together with information from Wikipedia lists of German fairy tales 
and English short stories. For documents presented in several sources, the oldest version or the version with historical spelling is selected. For stage (2), we utilize the collected information to match the historical documents with their corresponding summaries from Wikipedia. The historical spelling of titles is converted to modern spelling to match the titles in Wikipedia. Figure \ref{ex_collection} illustrates the process of data collection. If no original human correspondent summaries are found in the other language, 
we translate the summaries into the target language using DeepL (we translate 144 out of 328 German summaries to English and 108 out of 289 English summaries to German).\footnote{DeepL API: \url{https://www.deepl.com/pro-api}} An example of our collected dataset is shown in Table \ref{data:example}.

In addition to the summarization datasets, we create a historical German-English translation dataset with 201 translation pairs. This dataset is later utilized in intermediate task finetuning. We retrieve the German documents from the DTA corpus and the historical English from Wikisource. The documents are translated by Margaret Raine Hunt (1831–1912). An example of the translation dataset is given in Table \ref{appdx:Tran_example} (appendix).

We use the \emph{abbreviations} hDe and hEn for historical German and historical English texts to distinguish them from modern texts De or En. Thus, we refer to the historical German to modern English summarization dataset as  ``hDe-En'' and the historical English to modern German summarization dataset as 
``hEn-De''. Similarly, monolingual CTS datasets are called ``hDe-De'' or ``hEn-En''. Names of other modern datasets follow the same rule.

\subsection{External datasets} 
To enrich our finetuning data resources, we utilize external summarization datasets: (1) 
HISTSUMM \cite{2021summarising} where we also translate the summaries into English using DeepL  
to enable cross-lingual finetuning.
(2) We further use Wikilingua for MLS and CLS tasks \cite{ladhak-etal-2020-wikilingua}. Wikilingua is a multilingual summarization dataset that contains 18 languages from WikiHow (an online Wiki-style publication featuring how-to articles), including English and German. The English dataset contains 141,457 document-summary pairs. For finetuning our models, we randomly select 60\% of the pairs (85,874) and we use all of the German dataset (58,341). Wikilingua German entries contain the URL and the section name of the English articles. Thus, cross-lingual summary pairs can be retrieved using this information and we match 20,103 cross-lingual summary pairs in both directions.  
(3) We also utilize the CNN/Daily Mail dataset.
This is a monolingual English summarization dataset containing 311,971 document-summary pairs, including news articles written by CNN and Daily Mail journalists, and summaries from the highlight of the article written by the article author. 
We experiment with a subset of 86,133 text-summary pairs. (4) We include MLSUM \citep{scialom-etal-2020-mlsum}, which is a multilingual summarization dataset containing over 1.5 million document-summary pairs collected from online newspapers such as Süddeutsche Zeitung (German). We use a subset of 66,226 pairs from the German dataset. 

\subsection{Dataset Statistics}
\begin{table*}
\fontsize{6.5pt}{6.5pt}\selectfont
\begin{tabularx}{\textwidth}{lllllllll}
\toprule
\textbf{Dataset} & \textbf{Size}  &  \multicolumn{4}{c}{\textbf{Mean Length}} & \textbf{Compression}& \textbf{Genre} 
& \textbf{Task} \\
& & Doc.  &  Summ. &  $\text{Sentence}_{\text{Doc.}}$ & $\text{Sentence}_{\text{Summ.}}$  & \\
\midrule 
\multicolumn{9}{c}{\textit{CLCTS Corpus}} \\
CLCTS hEn-De & 289 & 9,643.3 & 414.0 & 16.3& 19.2& 23.3&fiction & CLS, CTS\\
CLCTS hDe-En & 328 & 1,398.4  & 324.3 &25.4& 18.6& 4.3&fiction & CLS, CTS \\
CLCTS hEn-En & 276  & 9,847.2 & 488.1 &16.0& 20.2& 20.2&fiction & MLS, CTS \\
CLCTS hDe-De & 328 & 1,398.4  & 217.4 &25.4&16.6&6.4& fiction & MLS, CTS  \\
\multicolumn{9}{c}{\textit{External Corpus}} \\
HISTSUMM hDe-De & 100 &268.1 &18.1& 36.3&12.4&14.8&news & MLS, CTS\\ 
Wikilingua En-En &  85,874 & 425.8 & 39.6 &16.5&6.5&10.8& how-to-guide & MLS \\
Wikilingua De-De & 58,341 &429.8 &42.2 &15.8&6.9& 10.2&how-to-guide & MLS \\ 
Wikilingua De-En & 20,103 & 438.2 & 38.8 &15.8&6.5& 11.3&how-to-guide & CLS\\ 
Wikilingua En-De & 20,103 & 451.7 & 42.7 &16.5&6.9& 10.6&how-to-guide & CLS \\ 
CNN/Daily Mail En-En & 86,133 & 786.7 & 55.1 &20.5&14.5& 14.3& news & MLS \\
MLSUM De-De & 66,226 & 570.3 & 30.4 & 17.8 & 12.7 & 18.8&news & MLS \\
\bottomrule
\end{tabularx}
\caption{Characteristics of different datasets. We compute (1) the mean length of texts of both document and summary; (2) the mean length of sentences from documents ($\text{Sentence}_{\text{Doc.}}$) and sentences from summary ($\text{Sentence}_{\text{Summ.}}$). (3) Compression represents the document-summary level compression ratio (Mean Length Doc. divided by Mean Length Summ.). (4) Task represents the use case of the dataset during intermediate task finetuning: MLS, CLS and CTS represent monolingual, cross-lingual and cross-temporal summarization.}
\label{data:exp-stat}
\end{table*}

In Table \ref{data:exp-stat}, we report statistics of our corpus and other existing summarization datasets. We observe three trends comparing all datasets: 
\begin{itemize}
    \item \emph{Much fewer instances in Historical datasets.}
The CLCTS hEn-De dataset consists of 289 historical short stories and plays, and the CLCTS hDe-En dataset consists of 328 historical fairy tales. \footnote{We limit the maximum length of the document to 16k. Since we collect summaries from Wikipedia, the total number of matched document-summary pairs is limited. Stories with less popularity are less likely to be found on Wikipedia than well-known ones.} The size of HISTSUMM dataset is even smaller, with only 100 historical news articles. In contrast, modern summarization datasets contain several thousand summarization pairs, which is more than 100 times the size of historical datasets. The largest dataset CNN/Daily Mail contains over 300k summarization pairs. 
\item \emph{Longer documents and summaries for CLCTS corpus.} The CLCTS hEn-De dataset has more than 9,643 tokens per document and 414 tokens per summary on average and CLCTS hDe-En has 1,398 per document and 324 per summary. For external datasets, the average length is below 800 tokens per document and below 60 tokens per summary. This is because the CLCTS corpus includes historical documents from the genre fiction, which are naturally longer than news and WikiHow (how-to guide) due to genre differences \citep{rudnicka2018variation} and higher verbosity of more historical languages. 
\item \emph{Larger information loss during summarization for news genre and CLCTS hEn-De measured by compression ratio.} News documents have a higher compression ratio (the mean length of document divided by the mean length of summary) compared to documents from other genres, except for CLCTS hEn-De, which has the highest compression ratio among all datasets. 
The CLCTS hEn-De dataset contains short stories and plays that are much longer than fairytales in CLCTS hDe-En and documents from all other genres as discussed above. For the CLCTS corpus, the compression ratio differs substantially because all summaries of the CLCTS corpus are collected from Wikipedia with the average length of summaries ranging from 200 to 400 tokens regardless of the original document length. 
\end{itemize}
To obtain more information on document-summary pairs, we quantify the divergence of sentence semantics between document and summary by computing the mean sentence-level cosine similarity for all datasets \citep{vyas-etal-2018-identifying}. 
We obtain the sentence embedding via multilingual SBERT \citep{reimers-2020-multilingual-sentence-bert}. We first compute the average similarity of each source to each summary sentence to obtain document level similarity and then average over all documents to obtain dataset level similarity score. 
The results are shown in Table \ref{data:sentence_semantics}. It reveals that cross-lingual sentence semantics between document and summary is more diverged than that of monolingual datasets and among all, both CLCTS hDe-En and hEn-De (which are our target tasks) have lower embedding similarity compared to other datasets. This again indicates the difficulty of cross-lingual and cross-temporal summarization.   
\begin{table}
\fontsize{7pt}{7pt}\selectfont
\begin{tabular}{llll}
\toprule
\multicolumn{2}{c}{\textbf{Monolingual}}  & \multicolumn{2}{c}{\textbf{Cross-lingual}} \\
\midrule
CLCTS hDe-De  & 0.38 & CLCTS hDe-En &0.33\\
CLCTS hEn-En & 0.27  & CLCTS hEn-De & 0.25 \\
HISTSUMM hDe-De &  0.39 & -& - \\
CNN/Daily Mail En-En & 0.32 & - & -\\
MLSUM De-De &  0.37  & -&- \\
Wikilingua De-De  &  0.46 & Wikilingua De-En  & 0.42\\
Wikilingua En-En   & 0.42 & Wikilingua En-De   &  0.42 \\
\bottomrule
\end{tabular}
\caption{\label{data:sentence_semantics}
Divergence of sentence semantics measured by mean cosine similarity of sentence level embeddings between document and summary.}
\end{table} 
We include an analysis of mean sentence length in the next section where we focus on a detailed analysis of historical documents and their comparison to modern documents. 

\subsection{Historical Language Divergence} 
Languages evolve due to social, cultural, and linguistic pressure and this poses challenges to the summarization of historical texts. In this section, we firstly provide an overview of the publication time of the historical documents from HISTSUMM and CLCTS and secondly, analyze the divergence of historical language from syntax (including the length of sentences and dependency distances) and lexicon perspectives. \newline
\begin{figure}[h]
    \centering
    \includegraphics[width=0.5\textwidth]{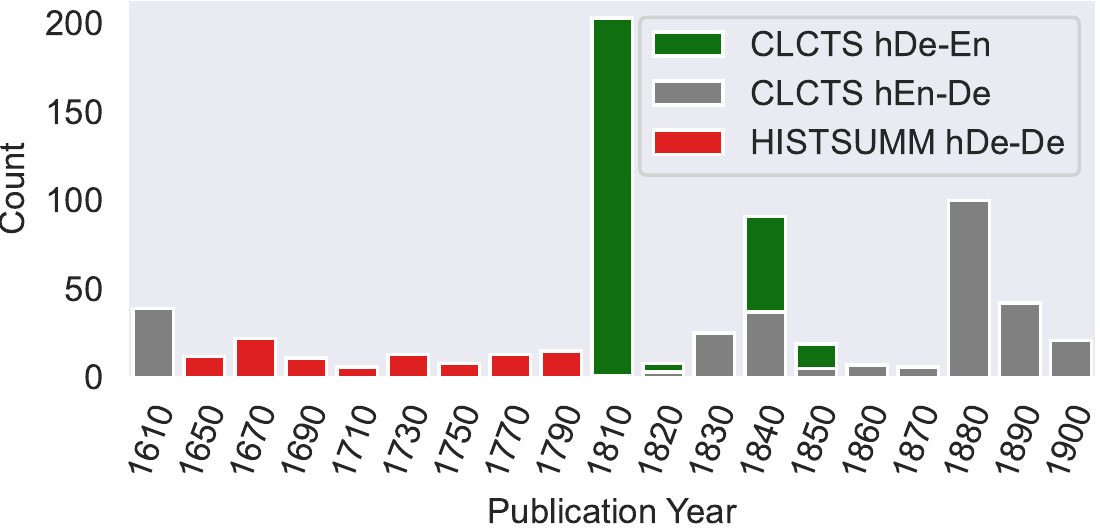}
    \caption{Distribution of publication year.}
    \label{data:year}
\end{figure}

\noindent \textbf{Publication time of historical documents}
The distribution of publication year is shown in Figure \ref{data:year}. HISTSUMM documents (in red) distribute evenly between the period of 1650–1800. CLCTS hEn-De (in gray) contains more documents from the late 1800s and CLCTS hDe-En (in green) from the early and mid 1800s. The documents from CLCTS hEn-De which date back to the 1600s are plays from Shakespeare. For both CLCTS datasets, the majority of historical documents come from the 1800-1900s and comparatively HISTSUMM has overall older documents than our CLCTS corpus. \newline

\noindent \textbf{Divergence in syntax}
\emph{Length of Sentences} The analysis in Table \ref{data:exp-stat} (column 5) reveals that historical German documents in HISTSUMM and CLCTS corpus contain longer sentences compared to other datasets, where HISTSUMM hDe-De has a mean sentence length of 36.3 tokens and CLCTS hDe-En/hDe-De dataset has 25.4 tokens per sentence while modern German documents from Wikilingua and MLSUM  have maximum 20 tokens per sentence on average. This coincides with the finding of decreasing sentence length over time \citep{rudnicka2018variation}, considering that HISTSUMM documents are the oldest. Slightly surprisingly, this is not true for the English documents, where the mean sentence length of historical English documents (CLCTS hEn-De/hEn-En) is almost the same as Wikilingua and is even shorter than documents from the news genre CNN/ Daily Mail. This hints at a genre effect, besides the temporal effect: fiction, especially plays, may have shorter sentences than news and other non-fiction genres due to the higher occurrence of dialogues and two-word sentences \citep{rudnicka2018variation}. In our CLCTS corpus, unlike CLCTS hDe-En/hDe-De containing only fairytales, historical English documents contain plays. When we exclude plays from Shakespeare, the mean sentence length of the CLCTS hEn-De document increases by 4 tokens per sentence to 20.3, which is closer to the value of CNN/Daily Mail. The factual summaries from CLCTS hEn-De/hEn-En in column 6 also contain longer sentences than their fictional documents for similar reasons. 
\begin{figure}[!htbp]
\includegraphics[width=0.5\textwidth]{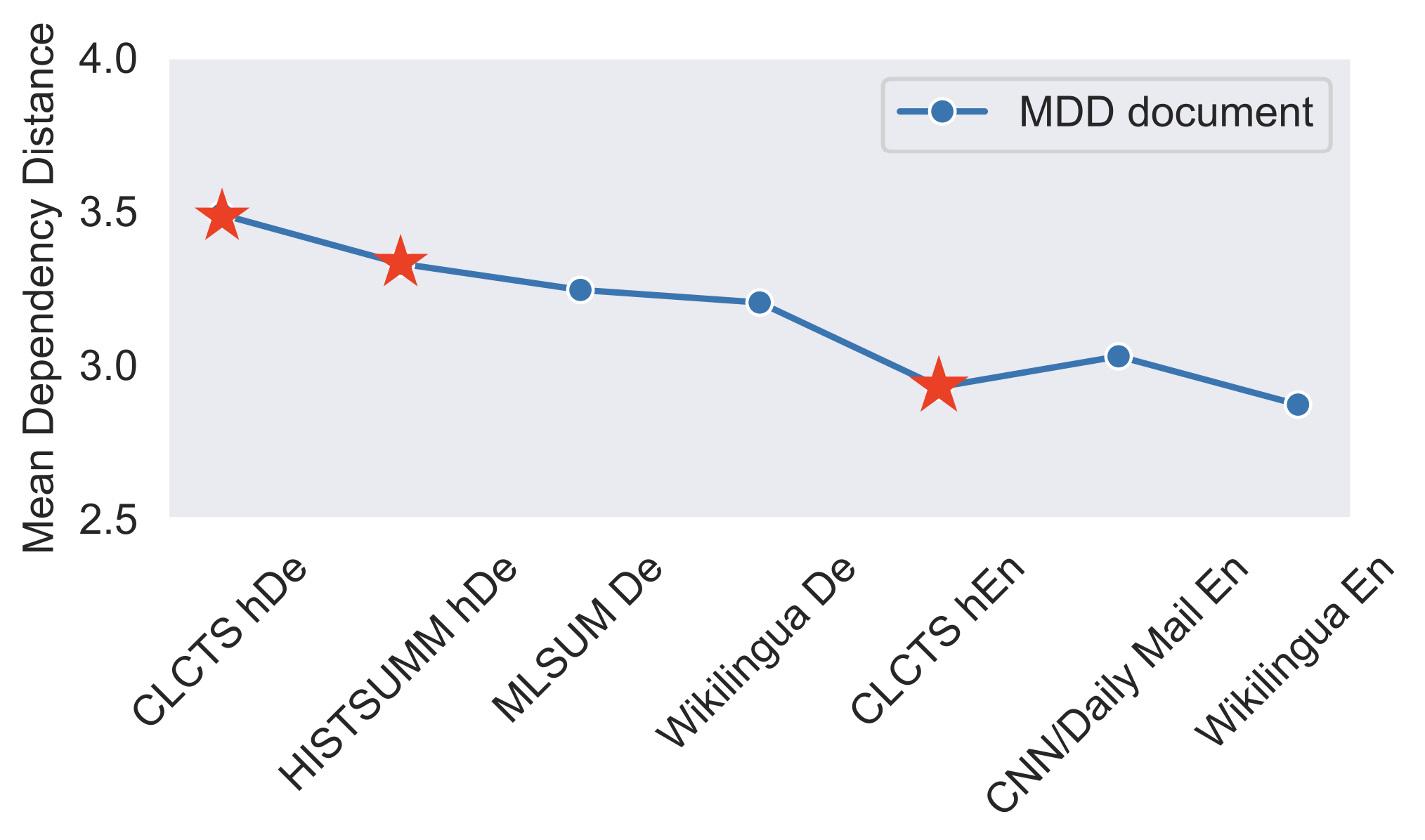}
\caption{Mean dependency distance (MDD) for document sentences in each dataset. Historical documents are marked with red stars.}
\label{data:DD}
\end{figure}

\noindent\emph{Dependency Distances} To further quantify syntactic changes, we analyze the dependency parsing trees for sentences from documents in all datasets and compute the mean dependency distance (MDD) over the entire corpus \cite{liu2008dependency,liu2017dependency}. This measure can reflect the syntactic complexity of sentences. We use stanza \citep{qi-etal-2020-stanza} for parsing the sentences. We plot the MDD in Figure \ref{data:DD} where we observe a downward trend overall. The first four documents from the left are in German and the last three documents are in English (we use dataset plus language, e.g., CLCTS hDe to represent the documents here). The downward trending line reveals an overall higher MDD of German sentences than that of English sentences. If we take a closer look by language, we observe: (1) For German (the first four points from the left), historical sentences (in red star) have a higher value for MDD than that of the modern German text, which suggests more syntactic complexity of historical documents. We especially notice a higher MDD from CLCTS hDe documents (fiction) than that from HISTSUMM hDe (news), which coincides with the finding of longer dependencies in the imaginative genre than those in the informative genre from \citet{wang2017effects}. The modern sentences from MLSUM are also more complex syntactically compared to Wikilingua. (2) For English (top-3 right items), this is not entirely true. News documents from CNN/Daily Mail have again a longer MDD than that of Wikilingua. However, they also have a longer MDD than CLCTS hEn. This may be because of the combined effect of (1) shorter sentences from CLCTS hEn (also from Wikilingua En) as discussed above, (2) more conversational texts with shorter dependencies from CLCTS hEn documents (plays and short stories) \cite{wang2017effects}, and (3) CLCTS hEn documents mostly from the late 1800s which share more similarities to modern English. \newline

\noindent\textbf{Divergence in lexicon} The vocabulary used in historical and modern language varies greatly due to spelling, morphological and semantic changes. For example, in our historical German dataset, the letter ``ß'' often represents ``ss'' in modern texts (e.g., ``daß'' vs. ``dass''). Omitting ``h'', ``t'' or ``e'' is another type of spelling change such as ``Thüre'' (\emph{door}, now ``Türe'' or ``Tür''), ``thun'' (\emph{to do}, now \emph{tun}),  ``Todt'' (\emph{death}, now ``Tod'') and ``gieng'' (passive tense of \emph{to go}, now ``ging''). Such changes in German are a result of two spelling reformations, one in 1901 and one in 1996. In English, we can also observe changes in lexicon such as the word ``you'' written as ``thou'' and the morphological change of the second person ``will'' as ``wilt''. To quantify these divergences, we compute the Jaccard similarity \citep{niwattanakul2013Jaccard} of the lexicons between documents and their summaries in Table \ref{data:divergence}. Historical German has the lowest lexical overlap with modern German, with a Jaccard score of 0.1 for HISTSUMM and 0.19 for CLCTS hDe-De. The pattern is similar for English. 
\begin{table}[h]
\fontsize{7pt}{7pt}\selectfont
\begin{tabular}{ll}
\toprule
\textbf{Dataset} & \textbf{Jaccard}   \\
\midrule
CLCTS hEn-En & 0.234 \\
CNN/Daily Mail En-En & 0.377  \\ 
Wikilingua En-En & 0.342  \\
CLCTS hDe-De & 0.186  \\
HISTSUMM hDe-De & 0.101  \\
MLSUM De-De & 0.335 \\
Wikilingua De-De & 0.314  \\
\bottomrule
\end{tabular}
\caption{\label{data:divergence}
Lexicon overlap measured by Jaccard score}
\end{table} 

\emph{Semantic shift} is another source of language change \citep{eger-mehler-2016-linearity, hamilton-etal-2016-diachronic}. For example, the word ``mistress'' in Shakespeare's work has a similar meaning as ``sweetheart''. The ``car'' mentioned in Shakespeare's work (e.g., ``Phoebus' car'') refers to a special kind of chariot. Another example is the adjective ``gay'' which is used in the sense of ``glad and cheerful'', \zrf{for example}, ``The great city in which he dwelt was very gay, for every day strangers visited the town'' (The Emperor's New Clothes by Christian Andersen in 1888). Similarly, the meaning of the word ``knave'' degenerates from ``servant'' in the story of Brother Grimm to the modern meaning ``a dishonest man'' over time.  

To conclude: (1) Historical German documents from HISTSUMM and CLCTS contain longer sentences with decreasing sentence length over time (temporal effect) while the sentence length of our historical English documents shows barely any difference to modern texts because of the genre effect. (2) German sentences show an overall higher MDD than that of English sentences in our entire corpus due to the joint effect of genre and publication time. (3) The vocabularies of
historical and modern languages differ greatly in terms of spelling, morphological, and semantic changes.

\section{Experimental setup}
\label{sec:setup}
\begin{figure}[h]
    \centering
    \includegraphics[width=0.4\textwidth]{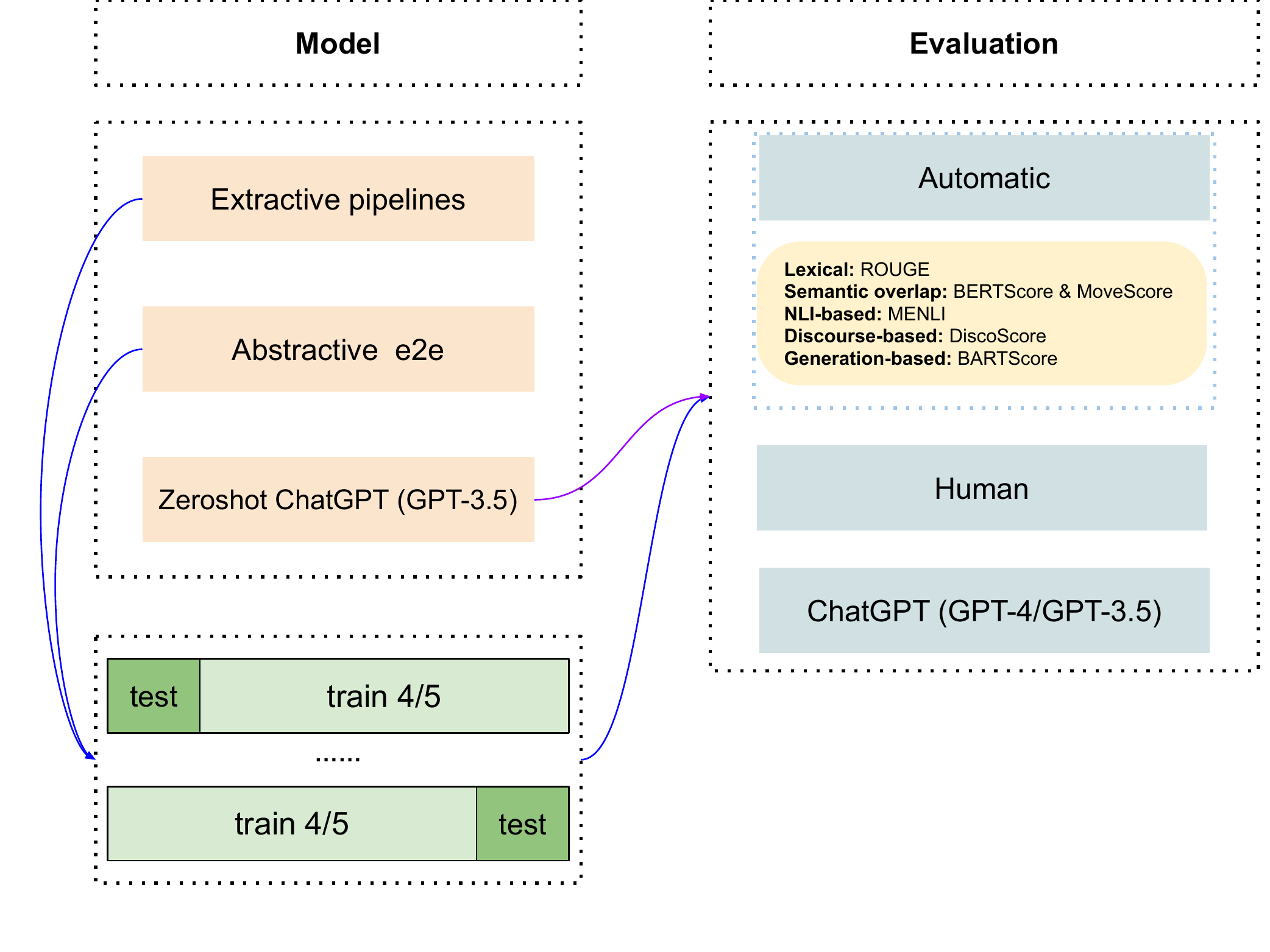}
    \caption{Flow chart illustrating models and evaluation strategies in Section \ref{sec:setup}.}
    \label{fig:flow}
\end{figure}

In this section, we introduce the experimental setup of CLCTS. Figure \ref{fig:flow} illustrates the model designs (detailed in Section \ref{subsec:models}) and evaluation strategies (detailed in Section \ref{subsec:eval}).
\subsection{Models}
\label{subsec:models}
Our experiments consist of three methods: (1) pipeline methods based on extractive summarizers, (2) e2e transformer-based methods with intermediate task finetuning, and (3) zero-shot ChatGPT with different prompt strategies. For finetuned methods (extractive and abstractive), we use 5-fold cross-validation to obtain reliable results for all CLCTS datasets. For ChatGPT, we query summaries for all CLCTS datasets as well to make sure the results are comparable.\footnote{During 5-fold cross-validation, we accumulate the test sets from each fold and in this way, we obtain a final test set consisting of all samples. For example, for direction hDe-En with 328 samples, we accumulate output summaries from each test set consisting of 65-66 samples and overall we collect 328 samples after 5-fold cross-validation. In contrast, for zero-shot ChatGPT, we query directly from ChatGPT and collect 328 summaries at once.} We summarize model details in Table \ref{exp:exp_design}. 
\begin{table*}[!ht]
\fontsize{6pt}{6pt}\selectfont
\begin{tabular}{m{2.5cm}m{6.5cm}m{2.5cm}m{0.5cm} }
\toprule
\textbf{Model name} & \textbf{Methods} &  \textbf{Type}  & \textbf{CV} 
\\
\midrule
\multicolumn{4}{l}{\centering \textbf{MemSum}}\\
-$\text{translation}_{\text{max 25}}$ & extract-then-translate pipeline with 25 maximal extracted sentences per source document & \multirow{2}{*}{extractive pipeline} & \multirow{2}{*}{\checkmark} 
\\
-$\text{Norma-translation}_{\text{max 25}}$  & extract-normalize-translate pipeline with 25 maximal extracted sentences per source document&  &    \\
\multicolumn{4}{c}{
}\\
\multicolumn{4}{l}{\centering \textbf{mLED}}\\
-Base & finetune with CLCTS target task alone & \multirow{3}{*}{abstractive e2e} & \multirow{3}{*}{\checkmark} 
\\
-Historical MT & finetune with CLCTS historical translation dataset \& CLCTS target task & & \\
$\text{-Bidirection}_\text{Prefix}$ & finetune with CLCTS non-target task \& CLCTS target task &    &  \\
\multicolumn{4}{c}{
}\\
\multicolumn{4}{l}{\centering \textbf{mLED}}\\
$\text{-MLS}_{\text{tgt}}$ &  \multirow{4}{*}{finetune with intermediate tasks listed in Table \ref{exp:abs} \& CLCTS target task} & \multirow{4}{*}{abstractive e2e} & \multirow{4}{*}{\checkmark} 
\\
$\text{-MLS}_{\text{src + tgt}}$ &  &  &  
\\
-MLS-CLS &  &  &  
\\
-MLS-CLS-CTS  &  &  &  
\\
\multicolumn{4}{c}{
}\\
\multicolumn{4}{l}{\centering \textbf{ChatGPT as summarizer}}\\
-e2e (Title) prompt & prompt in e2e style using only Title and Author & \multirow{3}{*}{zero-shot abstractive e2e} & \multirow{3}{*}{\newcrossmark} 
\\
-e2e  prompt & prompt in e2e style using source documents & 	\\
-pipeline prompt & prompt in pipeline style using source documents  & \\
-$\text{retrieve-ChatGPT}_{\text{max 100}}$ &  retrieve max 100 key sentences \& prompt in e2e style using retrieved texts & zero-shot abstractive e2e (two-step) & \newcrossmark 
\\
\bottomrule
\end{tabular}
\caption{Summary of model designs. CV stands for 5-fold cross-validation.}
\label{exp:exp_design}
\end{table*}

\subsubsection{Pipeline based on extractive summarizer}
For extractive summarization, we first use an extract-then-translate pipeline. We choose MemSum \citep{gu-etal-2022-memsum} as our model, which utilizes multi-step episodic Markov decision processes to extract important sentences from a given text. We use 5-fold cross-validation to train and test MemSum with our monolingual dataset. We limit the maximal extracted sentences per text to 25 according to the average length of summaries in Section \ref{sec:dataset}. During inference, we sort the sentences according to the original text's order \zrf{and} then translate the extracted sentences to the target language using a multilingual machine translation model developed by Facebook (M2M) \cite{fanbeyond}. We refer to this model as MemSum-translation$_{\text{max 25}}$. We additionally experiment with historical text normalization before translation. We use the normalization method Norma\footnote{\url{https://github.com/comphist/norma}} \cite{bollmann2012automatic}. We choose the combined version that leverages Lookup, Rule-based, and Distance-based normalization which is claimed to perform well for both English and German \cite{bollmann-2019-large}.\footnote{Norma is recommended by the author when only little training data is available} We refer to this model as MemSum-Norma-translation$_{\text{max 25}}$.

\subsubsection{Transformer-based e2e models}
We modify pretrained models together with intermediate task finetuning to build e2e models. We propose a modified multilingual mBART50 model pretrained for machine translation \cite{tang2020multilingual} with an efficient attention mechanism taken from LED \citep{beltagy2020longformer} as model architecture. This architecture is suitable for multilingual long sequences as it benefits from the cross-linguality of mBART and the efficiency of processing long documents from LED. We refer to this model as mLED. We use 5-fold cross-validation to obtain reliable results. Due to resource limitations, our mLED models have a maximum of 4,096 input tokens and the number of tokens for the generated summary ranges from a minimum length of 100 tokens to a maximum length of 512 tokens. \newline

\noindent\textbf{Baselines}
We first provide baselines of mLED finetuned only within the CLCTS corpus: 
(1) 
for baseline 1, we finetune the model only with the target task; \zrf{for example}, for the model in direction hDe-En, we only finetune mLED with our training set of the same direction. We refer to this model as \textbf{mLED-Base}. 
(2)
For baseline 2 (baseline 1 + historical machine translation), we utilize our historical translation dataset. We first finetune with the translation task and then with the target task, which we call \textbf{mLED-Historical MT}. 
(3) Finally, for baseline 3, we finetune with datasets from \emph{both} directions. We add prefixes to indicate the directions. We refer to this model as \textbf{mLED-$\text{Bidirection}_\text{Prefix}$}. \newline

\noindent \textbf{Intermediate task finetuning}
Apart from the final finetuning step with our target CLCTS task, we conduct intermediate task finetuning using external sources including MLS, CLS, and CTS. Table \ref{exp:abs} depicts the detailed design. 

\begin{table}[!h]
\fontsize{7pt}{7pt}\selectfont
\begin{tabular}{lllll}
\toprule
\textbf{Model} & \textbf{MLS} & \textbf{CLS} &\textbf{CTS} \\  \midrule
$\text{MLS}_\text{tgt}$  & CNN/Daily Mail En-En (tgt) & &  \\
\multirow{ 2}{*}{$\text{MLS}_\text{src+tgt}$}   & MLSUM De-De (src) \\
& CNN/Daily Mail En-En (tgt) &  &  \\
MLS-CLS   & Wikilingua De-De \&  En-En & Wikilingua De-En   \\
MLS-CLS-CTS  & Wikilingua De-De \&  En-En & Wikilingua De-En & HISTSUMM hDe-En  \\
\bottomrule
\\
\multicolumn{4}{c}{\textbf{(1) Direction hDe-En}}\\
\\
\toprule
\textbf{Model} & \textbf{MLS} & \textbf{CLS} &\textbf{CTS} \\  \midrule
$\text{MLS}_\text{tgt}$  &  MLSUM De-De (tgt) & &  \\
\multirow{ 2}{*}{$\text{MLS}_\text{src+tgt}$}   & Wikilingua En-En (src) \\
& Wikilingua De-De (tgt) &  &  \\
MLS-CLS  & Wikilingua En-En \& De-De & Wikilingua En-De &  \\
MLS-CLS-CTS  & Wikilingua En-En \& De-De & Wikilingua En-De & HISTSUMM hDe-De \\
\bottomrule
\\
\multicolumn{4}{c}{\textbf{(2) Direction hEn-De}}
\end{tabular}
\caption{Intermediate task finetuning experiments for abstractive e2e models. Src and tgt represent monolingual summarization tasks corresponding to the source or target language of the target CLCTS task. The selected tasks and datasets are listed.}
\label{exp:abs}
\end{table}

\subsubsection{CLCTS with ChatGPT}
We test the summarization ability of ChatGPT with our CLCTS corpus. \newline

\noindent \textbf{Prompt Strategies}
We utilize GPT-3.5-turbo 
via the OpenAI API to query the summary. We set the temperature to 0.7 to allow for some randomness. We adopt (1) a pipeline translate-then-summarize prompt as well as a (2) e2e prompt. We set prompt lines to the same language as the target output since ChatGPT works better in this setting according to our initial experiments.\footnote{During our first round of data collection all the prompt lines were in English. We noticed that ChatGPT tends to output summaries in the wrong target language, especially when the summary is expected to be in German.} 
We truncate the texts that exceed ChatGPT input limits.\footnote{According to our experiment with the GPT2 tokenizer, we find that on average our German texts have 1.73 tokens per word (which is higher than the suggested 1.3 tokens per word ratio from OpenAI) and English 1.07 tokens per word. Therefore, we limit the German text to 2,048 words and English to 3,000 words. Since the 4,096 tokens limits from GPT-3.5-turbo are shared between both prompts and completions (outputs), we need to leave some tokens for the output summary.}
We also utilize other information such as title and author to test if ChatGPT has explicit ``prior'' knowledge of the input document and its corresponding summary. We list all our prompt lines in Table \ref{ChatGPT-prompt}. \newline

\noindent \textbf{Retrieve-then-summarize with ChatGPT}
To tackle input limitations of ChatGPT (e.g., for GPT-3.5-turbo a total of 4,096 tokens are shared between prompt and
completion),\footnote{\url{https://help.openai.com/en/articles/4936856-what-are-tokens-and-how-to-count-them}} we build the retrieve-then-summarize design, where we use MemSum as a \emph{key sentence retriever} and input the retrieved sentences to ChatGPT and request summarization in the target language. For this experiment, we limit the maximal extracted sentences per document to 100 according to the input limits of ChatGPT.
\begin{table}
\fontsize{7pt}{7pt}\selectfont
\begin{tabularx}{\textwidth}{llX}
\toprule
\textbf{Type} & \textbf{Src-Tgt}  &  \textbf{Prompt}  \\
\midrule
e2e & hDe-En &
Please summarize the following text in English : [Text]. \\
e2e & hEn-De &
Bitte fasse den folgenden Text auf Deutsch zusammen: [Text]. \\
e2e (Title) & De-En &
Please give me the summary of the story [Title] written by [Author]. \\
e2e (Title) & En-De & 
Bitte gebe mir die Zusammenfassung der Geschichte [Titel] von [Autor]. \\
pipeline & hDe-En &
Please first translate the following text into English and summarize the translated text: [Text]  \\
pipeline & hEn-De &
Bitte übersetze zuerst den folgenden Text auf Deutsch und fasse den übersetzten Text zusammen: [Text]. \\
\bottomrule
\end{tabularx}
\caption{\label{ChatGPT-prompt}
Prompt lines of ChatGPT CLCTS. 
}
\end{table}

\subsection{Evaluation}
\label{subsec:eval}
To identify the best model, we utilize multiple recently popular reference-based metrics. We also conduct a multi-phase human evaluation to evaluate the model and test the effectiveness of the resulting metrics. During the multi-phase evaluation, we give feedback to annotators after each phase for them to better align with the instructions. We list the details of all three evaluation strategies in Table \ref{tab:eval}. 
\begin{table}[!htbp]
    \centering
    \fontsize{7pt}{7pt}\selectfont
\begin{tabular}{m{1.3cm}m{2.2cm}m{2.5cm}m{5.3cm} }
\toprule
        \textbf{Type} & \textbf{Aspects} & \textbf{Source} & \textbf{Detail}  \\
        \midrule
        & Lexical overlap & ROUGE-variant  \cite{lin-2004-rouge}  & \textit{ROUGE-1} matching the texts through computing the unigram overlap; \textit{ROUGE-L} leveraging the longest common subsequence \\
        \multirow{1}{*}{\centering \textbf{Automatic}} & \multirow{2}{*}{Semantic overlap} & BERTScore \cite{bert-score} & measuring semantic overlap through calculating the token similarity \\
        & & MoverScore \cite{zhao2019moverscore} & measuring semantic overlap utilizing Word Mover’s distance \\
        & Generation-based & BARTScore \cite{NEURIPS2021_e4d2b6e6} &  metric based on text generation using BART\\
        & NLI-based  & MENLI \cite{chen2022menli} & metric based on natural language inference\\
        & Discourse-based & DiscoScore \cite{zhao-etal-2023-discoscore} &  metric using BERT to model discourse coherence\\
        \midrule
    \multirow{1}{*}{\centering \textbf{Human}} & consistency, coherence, fluency, and relevance & SummEval \cite{fabbri2021summeval} & 6 annotators, 130 annotated instances for direction hDe-En; 104 annotated instances for hEn-De.\\
    \midrule
    \textbf{ChatGPT} & prompting & - & prompts consisting of the same instruction for human annotators; 494 annotations for hDe-En \& 407 for hEn-De.\\
    \bottomrule
    \end{tabular}
    \caption{Details of the evaluation strategies.}
    \label{tab:eval}
\end{table} \newline

\noindent\textbf{Automatic Evaluation} 
We consider six scores as automatic metrics to reflect five different aspects for measuring the quality of output summaries. The configurations used for each automatic metric are presented in Table \ref{apdx:config} (appendix). \newline

\noindent\textbf{Human Evaluation}
We use six `expert' annotators to evaluate all models. 
Three of the annotators are female and three are male. All annotators have high competence in English as they are affiliated with the university: they are Master students (3), postgraduates (1), PhD students (1), and NLP faculty members (1). Three of them are native German speakers. Thus, the generated English summaries are evaluated by all six annotators and the generated German summaries are annotated by three native speakers.

We utilize the evaluation strategy by \citet{fabbri2021summeval} where the annotators score the summary from four perspectives: coherence, consistency, fluency, and relevance. The outputs are rated with a score ranging from 1 (worst) to 5 (best) for each dimension and we allow 0.5 increments. The annotators are presented with the source document, one reference summary, and one output summary during annotation.\footnote{The annotation details including instructions and the number of annotated instances per model are given in Section \ref{appdx:annotation} (appendix).} After phase 1, we give feedback to the annotators by providing a comparison of their scores with the average scores. The scores that differ substantially are highlighted and we encourage them to review the differences.
To speed up the process, we did not evaluate the same amount of documents for all the models. 
Nonetheless, we make sure that each model is evaluated for at least 4 identical source documents. For hDe-En, we collect 130 summary level instances and 104 instances for hEn-De. \newline

\noindent\textbf{ChatGPT evaluation}
Since human annotation is costly, we follow related work \cite{gao2023human, chiang2023large} and test ChatGPT as an evaluator for our CLCTS task. We differ in the following aspects: (1) our CLCTS task is harder to evaluate due to the long length of both reference and generated summary. (2) Our CLCTS task involves multilingual evaluation for both English and German. We give the same instructions to ChatGPT as to the human annotators.
For reproducibility purposes, we set the temperature to 0 to reduce the randomness. We collect evaluations for all available generated summaries from documents that have been annotated by humans in the previous step, resulting in 494 ChatGPT annotations for hDe-En and 407 ChatGPT annotations for hEn-De.

\section{Results}\label{sec:results}
\subsection{Automatic evaluation metrics}
The experiment results are shown in Table \ref{exp:model_res} for hDe-En (upper) and hEn-De (lower).
\begin{table*}[!ht]
\fontsize{6pt}{6pt}\selectfont
\begin{tabular}{lllllll}
\toprule
\textbf{Model} & \textbf{ROUGE-1/L}  & \textbf{BERTScore-P/R/F1} &\textbf{BARTScore} &  \textbf{MoverScore } &  \textbf{MENLI-$\text{W}_1$/$\text{W}_{.8}$/$\text{W}_{.3}$/$\text{W}_{.2}$} & \textbf{DiscoScore}\\
\hline
\multicolumn{7}{c}{\centering \textit{Supervised Extractive}}\\
\multicolumn{7}{l}{\centering \textbf{MemSum}}\\
-$\text{translation}_{\text{max 25}}$ & 0.320/0.175&0.521/0.574/0.545&\textbf{-3.311}&0.552&-0.232/0.407/0.463/0.474 & 0.551\\
-$\text{Norma-translation}_{\text{max 25}}$ &	0.315/0.171&0.514/0.567/0.538&\textbf{-3.360}&0.551&-0.248/0.398/0.450/0.461 & 0.536\\
\multicolumn{7}{c}{\centering \textit{Supervised Abstractive - Baseline CLCTS}}\\
\multicolumn{7}{l}{\centering \textbf{mLED}}\\
-Base & 0.391/0.201&0.537/0.561/0.547&-3.524&0.568&\underline{-0.206}/\underline{0.418}/0.469/0.479 & \textbf{\underline{1.838}}\\
-Historical MT & \underline{\textbf{0.393}}/0.200&0.539/\underline{0.566}/\underline{0.551}&\underline{-3.492}&\underline{0.569}&-0.216/0.416/\underline{0.473}/\underline{0.485} & \textbf{1.725}\\
$\text{-Bidirection}_\text{Prefix}$ & 
0.388/0.198&0.537/0.561/0.547&-3.534&0.567&-0.231/0.408/0.465/0.477 & 1.142\\
\multicolumn{7}{c}{\centering \textit{Supervised Abstractive - intermediate finetuning}}\\
\multicolumn{7}{l}{\centering \textbf{mLED}}\\
$\text{-MLS}_{\text{tgt}}$ & 
0.389/0.198&\underline{0.540}/0.556/0.547&-3.567&0.568&-0.255/0.398/0.461/0.473 & 1.426\\
$\text{-MLS}_{\text{src + tgt}}$ &0.386/0.201&0.536/0.547/0.54&-3.593&0.568&-0.252/0.397/0.453/0.464 & 1.595\\
-MLS-CLS & 0.380/0.198&0.530/0.544/0.536&-3.615&0.567&-0.241/0.400/0.448/0.458 & 1.685\\
-MLS-CLS-CTS  &
0.386/\underline{0.202}&0.532/0.548/0.539&-3.590&0.568&-0.246/0.399/0.451/0.462 & 1.635\\
\multicolumn{7}{c}{\centering \textit{Zero-shot Abstractive}}\\
\multicolumn{7}{l}{\centering \textbf{ChatGPT as summarizer}}\\
-e2e (Title) prompt &0.304/0.164&0.527/0.530/0.528&-3.857&0.542&-0.700/0.211/0.367/0.398 & 0.788\\
-e2e prompt &	\textbf{0.399}/\textbf{0.244}&\textbf{0.646}/\textbf{0.607}/\textbf{0.624}&-3.363&\textbf{0.575}&\textbf{-0.194}/\textbf{0.455}/\textbf{0.580}/\textbf{0.605} & 0.982\\
-pipeline prompt&	0.382/0.232&0.637/\textbf{0.597}/\textbf{0.615}&-3.422&\textbf{0.571}&-0.282/0.415/0.554/0.581 & 0.907\\
-$\text{retrieve-ChatGPT}_{\text{max 100}}$ & 0.382/\textbf{0.234}&\textbf{0.639}/\textbf{0.597}/\textbf{0.615}&-3.421&\textbf{0.571}&\textbf{-0.206}/\textbf{0.446}/\textbf{0.566}/\textbf{0.589} & 0.883\\
\bottomrule
\\
\multicolumn{7}{c}{\centering \textbf{(a) Direction hDe-En}}\\
\\
\toprule
\textbf{Model} & \textbf{ROUGE-1/L}  & \textbf{BERTScore-P/R/F1} &\textbf{BARTScore} &  \textbf{MoverScore } &  \textbf{MENLI-\text{$W_1$}/\text{$W_{.8}$}/\text{$W_{.3}$}/\text{$W_{.2}$}} & \textbf{DiscoScore}\\
\hline
\multicolumn{7}{c}{\centering \textit{Supervised Extractive Pipeline}}\\
\multicolumn{7}{l}{\centering \textbf{MemSum}}\\
-$\text{translation}_{\text{max 25}}$ 	&	0.315/0.131&0.830/0.839/0.834&-5.320&0.814&-0.648/0.236/0.387/0.418 & 0.301\\
-$\text{Norma-translation}_{\text{max 25}}$	&	0.311/0.130&0.829/0.838/0.833&-5.350&0.814&-0.652/0.234/0.383/0.413 & 0.291\\
\multicolumn{7}{c}{\centering \textit{Supervised Abstractive - Baseline CLCTS}}\\
\multicolumn{7}{l}{\centering \textbf{mLED}}\\
-Base	&	0.321/0.142&0.853/0.843/0.848&-5.275&0.823&-0.605/0.273/0.462/0.499 & \textbf{1.653}\\
-Historical MT	&	\textbf{0.328}/0.143&0.853/\textbf{0.845}/0.849&-5.301&0.823&-0.600/\textbf{0.276}/0.465/0.503 & 1.411\\
$\text{-Bidirection}_\text{Prefix}$	&	\textbf{0.328}/0.144&0.854/0.844/0.849&-5.276&0.824&\textbf{-0.601}/\textbf{0.276}/0.467/0.505 & 1.157\\
\multicolumn{7}{c}{\centering \textit{Supervised Abstractive - intermediate finetuning}}\\
\multicolumn{7}{l}{\centering \textbf{mLED}}\\
$\text{-MLS}_{\text{tgt}}$	&	0.316/0.141&0.855/0.841/0.848&-5.253&\textbf{0.826}&-0.613/0.269/0.460/0.498 & 1.500\\
$\text{-MLS}_{\text{src+tgt}}$	&	0.321/0.145&0.852/0.841/0.847&-5.311&0.825&-0.618/0.266/0.453/0.491 & \textbf{\underline{1.804}}\\
-MLS-CLS	&	\textbf{0.328}/0.147&0.855/0.844/0.850&\underline{-5.250}&\textbf{0.826}&\textbf{\underline{-0.586}}/\textbf{\underline{0.283}}/\underline{0.472}/\underline{0.510} & 1.050\\
-MLS-CLS-CTS	&	\underline{\textbf{0.341}}/\underline{\textbf{0.153}}&\underline{0.855}/\textbf{\underline{0.845}}/\underline{0.850}&-5.255&\textbf{\underline{0.827}}&-0.624/0.268/0.468/0.508 & 0.764\\
\multicolumn{7}{c}{\centering \textit{Zero-shot Abstractive}}\\
\multicolumn{7}{l}{\centering \textbf{ChatGPT as summarizer}}\\
-e2e (Title) prompt	&	0.288/0.135&0.857/0.840/0.849&-5.009&0.822&-0.745/0.218/0.444/0.489 & 0.590\\
-e2e prompt &	0.282/\textbf{0.150}&\textbf{0.876}/0.843/\textbf{0.859}&-4.949&0.823&-0.655/0.269/\textbf{0.509}/\textbf{0.557} & 0.572\\
-pipeline prompt &	0.285/0.147&\textbf{0.874}/0.843/\textbf{0.858}&\textbf{-4.933}&0.824&-0.636/0.274/0.505/0.551 & 0.797\\
-$\text{retrieve-ChatGPT}_{\text{max 100}}$	&	0.269/0.143&\textbf{0.874}/0.843/\textbf{0.858}&\textbf{-4.914}&0.824&-0.638/0.274/\textbf{0.506}/\textbf{0.552} & 0.605\\
\bottomrule
\\
\multicolumn{7}{c}{\centering \textbf{(b) Direction hEn-De}}\\
\\
\end{tabular}
\caption{Experiment results for CLCTS hEn-De and hDe-En. The best two results among all models are in bold font. The best score for the supervised abstractive models is indicated with an underline. MENLI-W represents MENLI combined with BERTScore-F1 with different weights using formula MENLI-\text{W}$_\text{i}$= $\text{i} \times \text{NLI-D} \text{ + (1-i)} \times \text{BERTScore-F1}$. See Table \ref{apdx:config} (appendix) for more information on the metric configuration.}
\label{exp:model_res}
\end{table*} \newline

\noindent \textbf{Supervised extractive pipelines} 
\emph{Finding 1. Slightly diminishing results from historical text normalization}: We observe a decrease in all metrics scores on MemSum pipeline normalized with Norma for both hDe-En and hEn-De. This is contrary to our hypothesis that spelling normalization of historical texts can boost the translation quality for translators originally trained for modern text. Even though Norma performs well at token-level normalization, it fails to consider the context of the words, where in our use case some historical spellings may correspond to multiple modern meanings depending on the context. An example is the word ``in'' which can function as an adverb in the phrase \zrf{for example}, ``in den Wald'' (in the forest) or the accusative case of the word ``ihn'' (him) in historical texts. A false normalization of such cases results in the meaning change of the entire sentence. \newline

\noindent \textbf{Supervised abstractive e2e} 
\emph{Finding 1. Mixed results on intermediate task finetuning}:   
For \textbf{hDe-En}, intermediate task finetuning with external sources does not improve scores, and sometimes finetuning with more tasks even deteriorates the outcome according to the metrics. This may be because our CLCTS hDe-En dataset, which dates back to the early 1800s, has substantial linguistic differences from the external sources as discussed in Section \ref{sec:dataset}. For \textbf{hEn-De}, the results show the opposite. We observe higher scores for models trained with more intermediate summarization tasks from external sources. Though, according to MENLI variants and BARTScore, it is not always better to add CTS tasks. 
This may be because the CTS tasks from HISTSUMM contain documents that have up to 200 years of difference in publication time and again lead to substantial linguistic differences.

\emph{Finding 2. Historical translation task improves model performance only mildly}:
For \textbf{hDe-En}, one of the best finetuned models is mLED-Historical MT, where we finetune intermediately with our historical translation task before our target CLCTS task, though according to the metrics, the improvement is very small. Similarly, for the other direction \textbf{hEn-De}, the historical translation task improves model scores but only mildly compared to intermediate tasks. Since the historical translation dataset contains only 201 pairs, a larger size of the historical translation dataset may further improve the results. \newline

\noindent\textbf{Zero-shot abstractive ChatGPT}
\emph{Finding 1. A close match between pipeline and e2e prompts:} For \textbf{hDe-En}, we observe slightly better scores for summaries generated with e2e prompts which tell ChatGPT to summarize in the target language directly (see Table \ref{ChatGPT-prompt}). However, in the following Section \ref{section:analysis},
we show that the e2e prompts are prone to invalid outputs.  

\emph{Finding 2. Competitive results from the retrieve-then-summarize design with ChatGPT: }We observe competitive results from retrieve-then-summarize design for both hDe-En and hEn-De. Especially for hDe-En, it is scored as one of the top two models by ROUGE-L, MoverScore, BERTScore and MENLI. Also worth noting is that this design not only performs well but also expands ChatGPT's potential for long document summarization.

\emph{Finding 3. ChatGPT can output summaries from its memory:} For the e2e (Title) prompt experiment, we input zero text from the document but only the title, author, and publication year. Slightly surprisingly, for both directions, the resulting scores are mostly comparable to that of the supervised extractive models if not better. ChatGPT is capable of outputting summaries based solely on meta-information for all of our documents. This raises our question on the actual summarization ability of ChatGPT and therefore in the following Section \ref{section:analysis},
we design multiple tests to investigate in more detail about ChatGPT involving CLCTS task.\newline

\noindent\textbf{Comparison among all approaches}
All the metrics give lower scores to extract-then-translate models (with/without normalization) and ChatGPT e2e (Title) prompt than other models, except for BARTScore, which rates the MemSum pipeline as the highest for hDE-En. However, \emph{mixed results} are observed from the metrics for determining the best models. BERTScore variants in both directions give higher scores to ChatGPT-related models (excluding e2e (Title) prompt). MENLI variants agree with this fully for hDe-En and partially agree for hEn-De. The same is observed from MoverScore and ROUGE-L. BARTScore contradicts with them and scores ChatGPT models higher for hEn-De, though not for hDe-En. Moreover, for hEn-De, ChatGPT is no longer dominating. However, the study from \citet{goyal_news_2022} also shows that current metrics fail to evaluate the zero-shot model output properly. To better understand the performance of models and the metrics, we also consider human evaluation.

According to our automatic metrics, we find (1) extractive pipelines perform the worst among all methods; (2) supervised abstractive e2e methods outperform pipeline methods, where the historical translation task improves model performance mildly for both directions and we speculate that the effect of intermediate task finetuning depends on the linguistic similarity of external sources and target sources; (3) zero-shot abstractive ChatGPT delivers the best results under e2e prompts but can output summaries from its memory based on the experiment results from e2e (Title) prompts. 

\subsection{Human and ChatGPT evaluation} 
\begin{table*}[!htb]
\fontsize{7pt}{7pt}\selectfont
\begin{tabular}{lllll}
\toprule
\textbf{Models} & \textbf{Coh.} & \textbf{Con.} & \textbf{Flu.} & \textbf{Rel.} \\
\midrule
hDe-En & 0.497 & 0.595 & 0.150 & 0.572 \\
hEn-De & 0.353 & 0.493 & 0.433 & 0.605 \\ 
\bottomrule
\end{tabular}
\caption{\label{dis:agree}
Annotation agreement measured by mean Spearman’s correlation coefficient. Coh., Con., Flu, and Rel. represent coherence, consistency, fluency, and relevance respectively. 
}
\end{table*} 
Table \ref{dis:agree} shows the average Spearman’s ranking correlation coefficient among the annotators for coherence, consistency, fluency, and relevance. We obtain moderate agreement for relevance in both directions and moderate agreement for coherence and consistency in hDe-En. The annotation agreement for hEn-De is slightly worse at a weak-moderate level except for relevance. One exception is fluency in hDe-En, where the correlation is poor. 

\begin{table*}[!ht]
\fontsize{6.5pt}{6.5pt}\selectfont
\begin{tabular}{lllll}
\toprule
\textbf{Model} & \textbf{Coherence}  &  \textbf{Consistency}  & \textbf{Fluency} & \textbf{Relevance } \\
\midrule
\multicolumn{5}{c}{\centering \textit{Supervised Extractive}}\\
\multicolumn{5}{l}{\centering \textbf{MemSum}}\\
-$\text{translation}_{\text{max 25}}$  &	2.52/1.60&2.92/1.50&2.98/1.20&3.32/1.80\\
-$\text{Norma-translation}_{\text{max 25}}$	&2.50/1.50&2.92/1.40&2.95/1.10&3.20/1.70\\
\multicolumn{5}{c}{\centering \textit{Supervised Abstractive - Baseline CLCTS}}\\
\multicolumn{5}{l}{\textbf{mLED}}\\
\text{-Base}	&	2.10/1.62&2.08/1.12&2.78/1.50&2.22/1.38\\
\text{-Historical MT}	&	\underline{2.86}/\underline{1.67}&\underline{2.63}/\underline{1.56}&3.01/\underline{1.78}&3.19/\underline{1.67}\\
\text{-$\text{Bidirection}_{\text{Prefix}}$}	&	2.43/1.30&2.28/1.10&2.54/1.30&2.58/1.20\\
\multicolumn{5}{c}{\centering \textit{Supervised Abstractive - intermediate finetuning}}\\
\multicolumn{5}{l}{\textbf{mLED}}\\
\text{$\text{-MLS}_{\text{tgt}}$} 	& 2.75/1.22&2.56/1.11&\underline{3.08}/1.22&\underline{3.39}/1.22\\
\text{$\text{-MLS}_{\text{src+tgt}}$} 	&	2.38/1.00&2.30/1.00&2.78/1.10&2.75/1.00\\
\text{-MLS-CLS}	&	2.20/1.00&1.90/1.00&2.48/1.00&2.42/1.00\\
\text{-MLS-CLS-CTS}	& 2.11/1.10&1.92/1.00&2.30/1.10&2.28/1.10\\
\multicolumn{5}{c}{\centering \textit{Zero-shot Abstractive - ChatGPT}}\\
\multicolumn{5}{l}{\textbf{ChatGPT as summarizer}}\\
-e2e (Title) prompt &2.55/1.70&3.05/1.20&4.10/3.70&2.02/1.40\\
-e2e prompt & 4.14/\textbf{3.30}&3.98/\textbf{2.40}&4.18/3.80&3.97/2.70\\
-pipeline prompt &	\textbf{4.35}/\textbf{3.30}&\textbf{4.30}/\textbf{2.40}&\textbf{4.55}/\textbf{4.00}&\textbf{4.35}/\textbf{3.00}\\
-$\text{retrieve-ChatGPT}_{\text{max 100}}$	&4.08/2.90&4.08/2.10&4.15/3.50&4.25/2.70\\
\bottomrule
\\
\multicolumn{5}{c}{\centering \textbf{(a) Direction hDe-En}}\\
\\
\toprule
\textbf{Model} & \textbf{Coherence}  &  \textbf{Consistency}  & \textbf{Fluency} & \textbf{Relevance} \\
\midrule
\multicolumn{5}{c}{\centering \textit{Supervised Extractive}}\\
\multicolumn{5}{l}{\centering \textbf{MemSum}}\\
-$\text{translation}_{\text{max 25}}$	&2.47/1.50&2.34/1.25&2.62/1.38&2.19/1.38\\
-$\text{Norma-translation}_{\text{max 25}}$&2.53/1.38&2.12/1.25&2.75/1.38&2.22/1.25\\
\multicolumn{5}{c}{\centering \textit{Supervised Abstractive - Baseline CLCTS}}\\
\multicolumn{5}{l}{\textbf{mLED}}\\
\text{-Base}	&	2.01/1.00&1.56/1.00&2.32/1.00&1.65/1.00\\
\text{-Historical MT} &	1.98/\underline{1.38}&1.78/\underline{1.25}&2.74/\underline{1.38}&1.78/\underline{1.25}\\
\text{-$\text{Bidirection}_{\text{Prefix}}$}	&	\underline{2.75}/\underline{1.38}&\underline{2.03}/1.12&\underline{2.78}/1.25&1.97/\underline{1.25}\\
\multicolumn{5}{c}{\centering \textit{Supervised Abstractive - intermediate finetuning}}\\
\multicolumn{5}{l}{\textbf{mLED}}\\
\text{$\text{-MLS}_{\text{tgt}}$} 	&	1.99/1.12&1.72/1.12&2.43/1.12&1.91/1.12\\
\text{$\text{-MLS}_{\text{src+tgt}}$} 	&	1.83/1.00&1.70/1.00&2.33/1.00&1.66/1.00\\
\text{-MLS-CLS}		&	1.80/1.00&1.71/1.00&2.39/1.00&1.98/1.00\\
\text{-MLS-CLS-CTS}	&	2.41/1.00&1.88/1.00&2.34/1.00&\underline{2.00}/1.00\\
\multicolumn{5}{c}{\centering \textit{Zero-shot Abstractive - ChatGPT}}\\
\multicolumn{5}{l}{\textbf{ChatGPT as summarizer}}\\
-e2e (Title) prompt	&	3.31/2.00&2.25/1.00&3.28/3.50&2.31/1.38\\
-e2e prompt & 3.31/\textbf{3.00}&2.80/2.22&3.38/3.56&2.98/2.33\\
-pipeline prompt&	3.34/2.86&\textbf{3.19}/2.00&\textbf{3.47}/3.57&\textbf{3.31}/2.57\\
-$\text{retrieve-ChatGPT}_{\text{max 100}}$	&\textbf{3.38}/\textbf{3.00}&2.94/\textbf{2.38}&3.34/\textbf{3.75}&3.06/\textbf{2.62}\\
\bottomrule
\\
\multicolumn{5}{c}{\centering \textbf{(b) Direction hEn-De}}\\
\\
\toprule
\textbf{Models} & \textbf{Coh.} & \textbf{Con.} & \textbf{Flu.} & \textbf{Rel.} \\
\midrule
hDe-En & 0.559 & 0.458 & 0.755 & 0.599   \\
hEn-De & 0.657 & 0.650 & 0.539 & 0.672 \\ 
\bottomrule
\\
\multicolumn{5}{c}{\centering \textbf{(c) Annotation agreement for human and ChatGPT}}\\
\end{tabular}
\caption{\label{dis:gpt4_annotation}
Average human and ChatGPT (GPT-4-1106-preview) ratings for the CLCTS dataset (in Table (a) and (b), the scores from human and ChatGPT are separated by a slash (/), i.e., human-annotation/ChatGPT-annotation). Table (c) is the document level annotation agreement between humans and ChatGPT. The best result of all models is in bold font. The best score for the supervised abstractive model is indicated with an underline.}
\end{table*}
We report the mean value of human and ChatGPT (GPT-4-1106-preview) annotations for each model in Table \ref{dis:gpt4_annotation}.\footnote{We utilize both GPT-3.5-turbo and GPT-4-1106-preview for annotation. GPT-4-1106-preview correlates better with human annotators than GPT-3.5-turbo. We report the annotation results from GPT-4-1106-preview in Table \ref{dis:gpt4_annotation} in the main text. The annotation results from GPT-3.5-turbo are reported in Table \ref{dis:annotation} in Section \ref{apdx:chatgpt_eval} (appendix).} The best scores for \emph{supervised abstractive models} in both directions range between 2 to mildly above 3 by human annotators and below 2 by ChatGPT, which indicates the quality of outputs is bad to mediocre. However, the highest scores for ChatGPT as a summarizer are over 4 (very good) for hDe-En and 3 to 4 (mediocre to good) for hEn-De by human annotators. Overall, the quality of outputs for hEn-De is lower than that for hDe-En. This is also expected since the length of historical English documents is much longer with 9,643 words on average which results in more information loss due to text truncation because of computational limitations\footnote{We also experimented with mBART finetuning with unlimiformer under both low-cost training and computationally expensive long-range training method. Since the low-cost training method only sees full texts at test time and long-range training suffers from the same GPU memory constraints as mLED, we do not observe improvements in model performance compared to baseline mLED.} and the cross-lingual embedding distance from Table \ref{data:sentence_semantics} also suggests more difficulty of hEn-De compared to hDe-En.

For both directions, humans give higher scores to summaries generated by ChatGPT, especially ChatGPT pipeline prompt for hDe-En with an increase of annotation score to a great margin. Unlike the evaluation metrics, humans prefer pipeline prompt output over e2e prompt output for hDe-En and regarding consistency, fluency and relevance for hEn-De. For comparing finetuned models, humans give higher scores for the model finetuned with historical translation in coherence and consistency similar to evaluation metrics for hDe-En. The ratings for hEn-De differ where human ratings prefer mLED-Bidirection$_{\text{Prefix}}$. Since the overall quality of generated summaries is worse for hEn-De, the comparisons among outputs may be more difficult for both humans and evaluation metrics.\newline

\noindent \textbf{Mixed results for ChatGPT (GPT-4-1106-preview) as an evaluator.} 
Table \ref{dis:annotation} (c) shows the annotation agreement between humans and ChatGPT annotation. For both directions, the agreement is moderate to good overall with a slightly better agreement for direction hEn-De (except for fluency). Similarly to the findings from \citet{chiang2023large}, we observe lower Likert scores from ChatGPT evaluation compared to human evaluation except for summaries generated by ChatGPT itself. However, the output for hEn-De is more concerning where we observe more ratings of score 1. Another point worth noticing is that even though in our instruction, we allow for 0.5 increments, we do not observe such ratings from ChatGPT output. 

To sum up, ChatGPT as an evaluator (1) is prone to lower scores than humans;\footnote{We include one sample annotation from the gold standard annotations during instruction.} (2) cannot distinguish summaries of low quality (hEn-De); (3) does not fully follow the instruction of 0.5 score increments. 

\subsection{Example outputs}\label{sec:exmaples}
In this section, we provide four sample outputs (for direction hDe-En) from both abstractive e2e models and ChatGPT together with the corresponding reference texts \zrf{(see Section \ref{apdx:sample} for a more detailed analysis)}. We use two documents as examples and for each document, we list one generated summary of mediocre to good level quality and one of low quality according to human annotation. The results are shown in Table \ref{result:example}. Based on the error taxonomy from \namecite{goyal-etal-2022-snac}, we highlight \textcolor{orange}{factual inconsistency} and \textcolor{purple}{language errors (unclear coreference of pronouns, nonsensical text and untranslated expression)} by different colors. 

For low quality generation (summary 1.2 and summary 2.2), we observe a substantial amount of factual inconsistency with the source text (1) and nonsensical text (3). In summary 2.2, we also notice the untranslated expression ``Nimmersatt" repeatedly appears in the text. Unclear coreference of pronouns is another issue that affects the readability of the outputs. For example, in the sentence from summary 1.2 ``the priest asks him if \textcolor{purple}{he (2)} can come to the wedding''. The pronoun ``he'' is ambiguous as it could refer to the boy or the priest himself. In contrast, summary 1.1 and summary 2.1 contain fewer errors, especially the ChatGPT output which only contains one factual inconsistency. In summary 1.1, we find two factual inconsistencies with the source text where the boy ``shares'' the food instead of ``cooking'' it himself and he ``asks'' the image instead of ``asking for'' it. Additionally, we also notice several other errors such as unclear coreferences of pronouns, nonsensical texts, and spelling mistakes (such as ``could't''). 
\begin{table}[h]
\fontsize{7pt}{7pt}\selectfont
    \begin{tabular}{p{\textwidth}}
    \toprule
    \textbf{Reference Texts}  \\
    ``A poor peasant boy hears a sermon that the way to heaven always leads straight ahead. He goes and comes to the service in a big city church, thinks he is in heaven and cannot be sent away. He sees people praying in front of Mary with the baby Jesus and thinks it is God. He shares his food with him, and the image grows thicker. Once, when he is bedridden for eight days, he goes and apologizes to the image. It invites him to the wedding next Sunday. The priest, who has been watching him, is not allowed to go, but gives him communion on Sunday, when the boy falls down dead.''\\
    \textbf{1.1 Generated Text from model mLED-Base, quality level 3-4} \\
    ``A poor peasant boy wants to go to Heaven, but the parson tells him to go straight. When he arrives in the church, he thinks God is there, and says, "Listen, dear God, I will bring you half of my food every day." Every day he \textcolor{orange}{cooks (1)} half of his food, and the image begins to eat. A couple of weeks later, when people realize that the image is getting bigger and bigger, he stays in church and sees his bread being shared with the mother God. He gets sick and can't get up for eight days. The parson goes after him and hears him say, "God, \textcolor{purple}{don't take it so long (3)}, I was sick and \textcolor{red}{could't} stand up." \textcolor{purple}{He answers (2)}, "That doesn't matter\textcolor{red}{,} I've seen your good will\textcolor{red}{,} that's enough, and next Sunday you are to come to my wedding." The boy rejoices and \textcolor{orange}{asks for the image (1)}. "No, it's you alone," says he. \textcolor{purple}{He (2)} first wants to prepare and gives him the supper. The next Sunday, when supper is coming to him, he falls down and is dead and is \textcolor{red}{to eternal wedding}."
    ''\\
    \textbf{1.2 Generated Text from model mLED-MLS-CLS-CTS , quality level 1-2} \\
    ``A poor peasant in the church hears \textcolor{orange}{God (1)} admonishing him to go to Heaven, and he goes to God's church. \textcolor{orange}{The priest tells him not to go out, but he wants to keep him (1)}. He stays in \textcolor{red}{church} and when he sees the mother God statue, which is cut from wood, \textcolor{purple}{the little Jesus (3),} \textcolor{orange}{he says that it's God, which means that he will eat half of his food every day (1)}. After a few weeks, people marvel at the statue and see how he spreads bread with \textcolor{orange}{God (1)}. After eight days, the priest asks him if \textcolor{purple}{he (2)} can come to the wedding. \textcolor{purple}{The boy replies, "No, you're only me, I told you that you told me that you were you alone. I'll give him dinner first, which he says he will give him, and then he dies, then he'll never be able to marry again," so he has to go back to the church and asks if he could come with God alone. After the wedding, the boy asks the parson to give him the dinner, but God says he's dead and he is dead, and God says that he must marry him again (3)}.'' \\
    \bottomrule
    \\
    \multicolumn{1}{c}{\centering \textbf{Example 1}}\\
    \\
    \toprule
    \textbf{Reference Texts}  \\
    ``The wolf is the stronger of the two and dominates the fox. The fox would like to be rid of him. The wolf forces the fox to get him food. He gets him a lamb and leaves. But the wolf is even hungrier and tries to get a lamb for himself. But he is so clumsy that the farmers notice and beat him. The wolf says to the fox that he has tricked him. The fox replied, `Why are you such a glutton!' The next day the fox has to get pancakes for the wolf. They go to the house together and the fox sneaks around the house to carefully get six pancakes from the plate. The wolf again doesn't have enough and wants to get more. But he again doesn't manage and the farmer's wife beats him again. The following day, the two go to the butcher. They jump through a window hatch into the cellar. The wolf eats a lot. The fox, on the other hand, eats little and keeps looking to see if he still fits through the window hatch. Suddenly the butcher comes, the red fox jumps out quickly, only the wolf gets stuck and is beaten to death by the butcher.''\\
    \textbf{2.1 Generated Text from model ChatGPT-e2e prompt, quality level 4-5} \\
    ``The wolf and the fox were walking in the woods together. The wolf demanded that the fox bring him food or he would eat him. The fox suggested they steal a lamb from a nearby farm. The wolf ate the lamb and wanted more, but when he tried to steal another, he was caught and beaten by the farmers. The next day, the fox suggested they steal pancakes from a farm, but the wolf ate too many and was caught again. On the third day, the fox suggested they steal salted meat from a cellar, but the wolf ate too much and got stuck in the cellar. The \textcolor{orange}{farmers (1)} found him and killed him, while the fox escaped and was happy to be rid of the greedy wolf.''\\
    \textbf{2.2 Generated Text from model mLED\text{$\text{-MLS}_{\text{src+tgt}}$}, quality level 1-2} \\
    ``\textcolor{orange}{The wolf wants to catch the fox because he is the weakest and would be the master. (1)} \textcolor{orange}{The fox tells the wolf to get a lamb. (1)} \textcolor{purple}{The lamb is not happy, (3)} but the other wants to have it too. The farmers run to get the lamb, and they hit \textcolor{purple}{him (2)}. \textcolor{purple}{When they come to rescue them, the witch tells them they are a nimmersatt (3, 4)}. They go to a farmer's house, where \textcolor{orange}{they (1)} cook pancakes, and \textcolor{orange}{the sheep eat the pancakes again. (1)} The sheep shout loudly and \textcolor{red}{calls} for help, \textcolor{purple}{but when she comes out, she says that they have been slaughtered by a man in the forest, and he says that he wants to help out. (3)} The \textcolor{red}{wolfe lets go} and jumps through the hole, but \textcolor{purple}{he does not want to see anything. He jumps out of the hole and kills the shepherd, who does not know what to do. (3)} On the third day, \textcolor{purple}{he (2)} is happy that he is rid of the old \textcolor{purple}{Nimmersatt (4)}, \textcolor{purple}{who has been rid of a man who is slaughtering meat in the cellar, and who is the first to get rid of him. (3)}'' \\
    \bottomrule
    \\
    \multicolumn{1}{c}{\centering \textbf{Example 2}}\\
    \end{tabular}
    \caption{Example of generated summaries and the reference texts. Sentences highlighted in \textcolor{orange}{orange} contain factual inconsistency with the source text (1). Unclear coreference of pronouns (2), nonsensical text (3), and untranslated expression (4) are highlighted in \textcolor{purple}{purple}. Spelling, grammar, and punctuation mistakes are highlighted in \textcolor{red}{red}.}
    \label{result:example}
\end{table} 
\section{Discussion and analysis}
\label{section:analysis}
\subsection{How well do different metrics perform at segment level?}
\label{anslysis:corr}
\begin{figure*}[!htb]
\centering
   \includegraphics[width=0.5\linewidth]{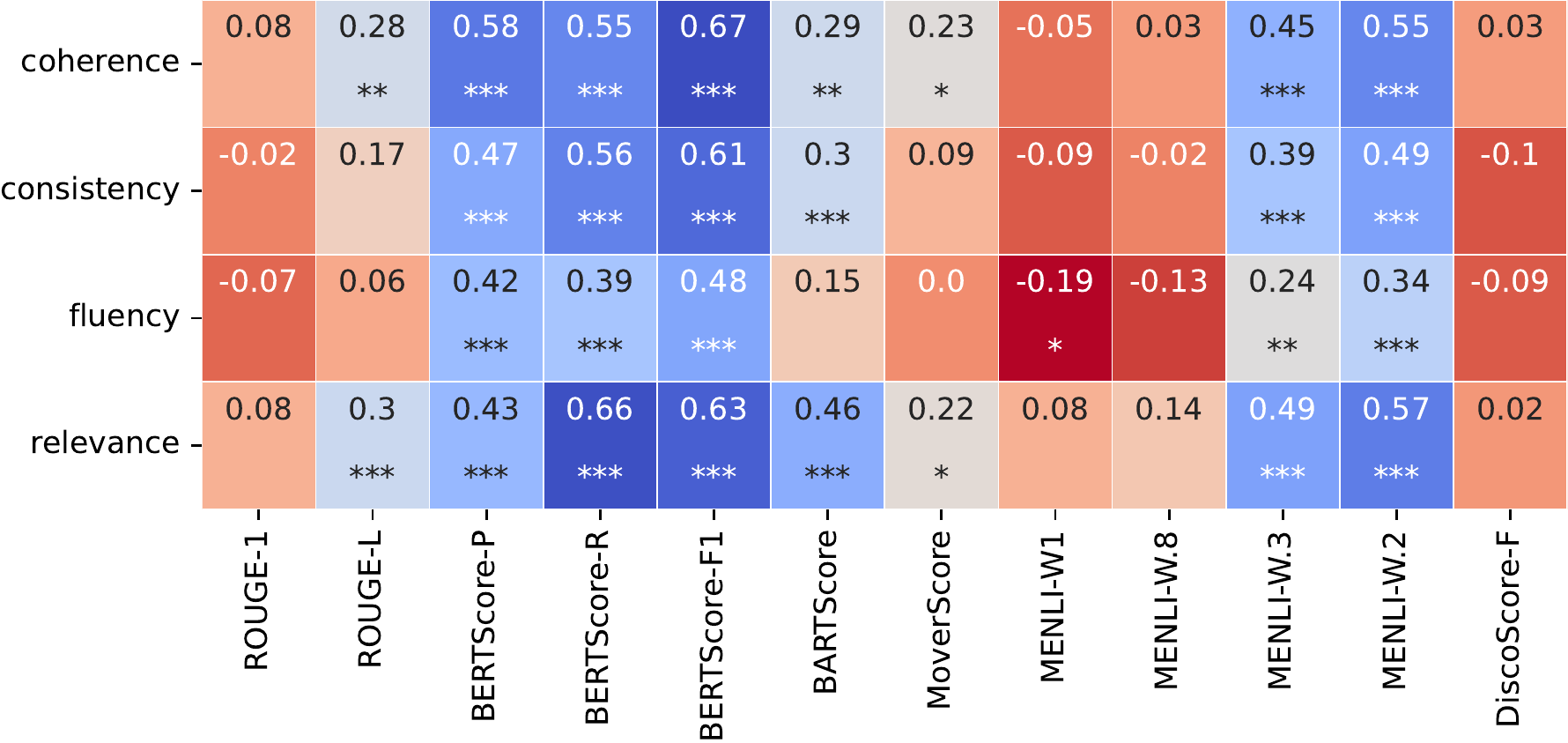}\\ 
   \centering \textbf{(1) Direction hDe-En}\\ 
   \vspace*{1mm}
   \includegraphics[width=0.5\linewidth]{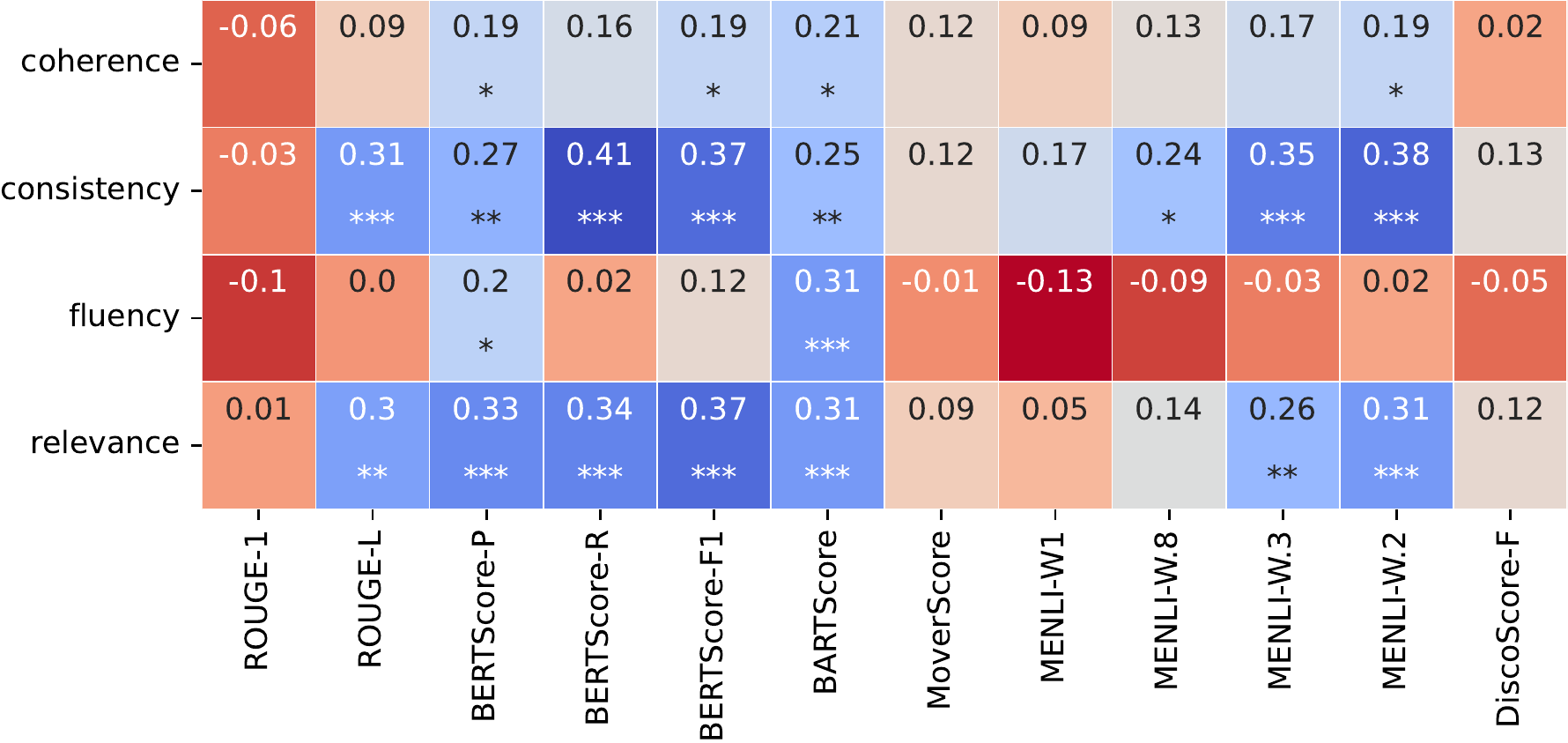} \\
   \centering \textbf{(2) Direction hEn-De}
    \caption{Segment level Spearman's ranking correlation between human annotation and evaluation metrics. The corresponding p-value is obtained through a two-tailed Student's t-test at significance levels 0.05 (*), 0.01 (**), and 0.001(***).}
    \label{fig:segment}
\end{figure*}

Figure \ref{fig:segment} shows Spearman's ranking correlation between human annotation and evaluation metrics at the segment level. \zrf{Recall that the correlation results between GPT-4 and human annotation are provided in Table \ref{dis:gpt4_annotation} (see Table \ref{dis:annotation} for GPT-3.5)}.  \newline

\noindent\textbf{Decent correlation from BERTScore variants but worse for German}
For hDe-En, we observe a weak to moderate level of correlation with human annotation for BERTScore, BARTScore, $\text{MENLI-W}_{.3}$, and $\text{MENLI-W}_{.2}$. MoverScore correlates weakly in coherence and relevance and the same is 
true for ROUGE-L. ROUGE-1 has low correlations with human annotation and the same applies for $\text{MENLI-W}_{.8}$ and DiscoScore. The situation is slightly worse for hEn-De where we observe a lower level of correlation in general compared to hDe-En and less significant correlations from BARTScore and BERTScore variants according to the Student’s t-test especially for coherence. ROUGE-L correlates moderately for direction hEn-De in consistency and relevance. Overall, BERTScore variants correlate slightly better than other metrics with humans in hDe-En for coherence (0.67), consistency (0.61), and relevance (0.66), and in hEn-De for consistency (0.41) and relevance (0.37).  

\subsection{Statistical analysis on model performance}
\label{regress}
To quantify the effect of document level characteristics on model performance and contribute to explainability of our models, we analyze how BERTScore-F1 as our best automatic metric overall changes depending on four document features reported in Section \ref{sec:dataset}, namely the mean length of documents (Length), mean dependency distance (MDD), publication year (Year) and divergence of sentence semantics between document and summary by computing the mean sentence-level cosine similarity (Similarity). We apply normalization to numerical features including BERTScore-F1, Length, MDD and Similarity to have zero mean and unit variance.\footnote{We exclude the concern of multicollinearity for numerical variables by computing variance inflation factors (VIF) and the details are reported in Section \ref{appdx:reg} (appendix).} 
We use publication year as a categorical variable where we categorize it by two time periods for hDe-En, namely 1800-1850 (Year:1800-1850) and after 1850 (Year:1850+), and for hEn-De we add another grouping which includes all documents before 1800 (Year:-1800). To examine model effects, we also encode a categorical variable to represent different models (Model). The regression formula is thus:  
\begin{equation}
    \text{BERTScore-F1}_i = \beta_0 + \beta_S\text{Similarity}_i +\beta_L\text{Length}_i + \beta_D\text{MDD}_i + \beta_{Y_j}\text{Year}_{i} + \beta_{M_k}\text{Model}_{i} + \varepsilon_i
    \label{formular}
\end{equation}
where $i$ represents documents, $j$ represents year groups, and $k$ represents models defined in Section \ref{sec:setup} and 
\begin{math}
\varepsilon_i \in \mathcal{N}(0, \sigma)    
\end{math} 
is an error term. We fit Equation (\ref{formular}) using ordinary least squares via statsmodels \cite{seabold2010statsmodels} separately for both directions (hDe-En and hEn-De). \\ \\
\begin{table}
\fontsize{7pt}{7pt}\selectfont
\begin{tabular}{llllll}
\toprule
 \multirow{2}{*}{\textbf{Direction}} & \multicolumn{3}{c}{\textbf{Numerical}} & \multicolumn{2}{c}{\textbf{Categorical}} \\
 & $\beta_S$ & $\beta_L$ &  $\beta_D$ & $\beta_{Year:1800-1850}$ & $\beta_{Year:1850+}$\\
\midrule
hDe-En  & 0.06$^{***}$ & -0.14$^{***}$ & 0.03$^*$ & base &-0.05  \\
hEn-De & 0.24$^{***}$ & -0.15$^{***}$ & -0.08$^{***}$ & -0.003 & 0.69$^{***}$  \\ 
\bottomrule
\end{tabular}
\caption{\label{reg:feature}
Regression results for document features (where $\beta_S$, $\beta_L$, $\beta_D$, and $\beta_{Year:*}$ represent coefficients for variable Similarity, Length, MDD, and Year respectively). The corresponding p-value is obtained through a two-tailed Student's t-test for coefficients at significance levels 0.05 (*), 0.01 (**), and 0.001(***). ``Base'' indicates the base group which is represented by
the intercept. For hEn-De, the base group for publication year is Year grouping before 1800 (Year:-1800).}
\end{table} 

\noindent\textbf{Effect of document features}
We report the regression results of document features in Table \ref{reg:feature}.\footnote{The adjusted $R^2$ is 0.425 for direction hDe-En and 0.551 for direction hEn-De.} For both directions, \textbf{embedding similarity} measured by $\beta_S$ has a significantly positive effect on model performance which indicates the model performs better with higher embedding similarity between document and summary. In contrast, \textbf{the mean length of document} (measured by $\beta_L$) shows a significantly negative relation to model performance where more tokens in the source document result in worse performance. This is expected since longer inputs increase the difficulty of summarization and also increase the chance of information loss due to input truncation. The effect of \textbf{mean dependency distance} ($\beta_D$) varies. We obtain a significantly negative effect for hEn-De which indicates that sentences with more syntactic complexity have a negative influence on model performance. A positive effect for hDe-En is obtained but with a low level of significance. The effect of \textbf{document publication year} is not significant for hDe-En where the model performance differs insignificantly between base group Year:1800-1850 and group Year:1850+. For hEn-De, this also holds for documents published in 1800-1850 compared to the base group with the publication year before 1800. However, we observe a significantly positive effect on model performance if we compare the group Year: 1850+ to base group Year:-1800. This effect also coincides with our finding from Section \ref{sec:dataset} that English documents from the late 1800s share more similarities to modern English, and thus have a positive impact on model performance.    

Also worth noticing is that the main contributor among document features to model performance varies between directions. For hDe-En, the main influence comes from the length of document and mildly from embedding similarity while for hEn-De, the main contributor is the publication year (group Year:1850+) where embedding similarity and the length of document rank the 2nd and 3rd among document features respectively. Overall, however, our findings are largely consistent and they are meaningful: as we argued in the introduction, older texts are more difficult to handle, and the same is true for longer and more syntactically complex source documents; more generally, the more divergence between source documents and reference, the harder is the summarization task.
\newline

\noindent\textbf{Effect of model choice}
The effect of model choice coincides with our findings from Section \ref{sec:results} where for both directions, we obtain a significant gain with ChatGPT models (except for ChatGPT title prompt) and in contrast, extractive models (extract-then-translate) are significantly worse than the baseline model \textbf{mLED-Base}. For hDe-En, the ChatGPT title prompt model performs significantly worse than the baseline \textbf{mLED-Base}, though not significant for hEn-De. We also notice that for hDe-En, models with intermediate finetuning provide significantly worse performance compared to the baseline as discussed in Section \ref{sec:results} and for hEn-De, the improvement of model performance adding intermediate finetuning tasks is not significant. We include the complete results in Section \ref{appdx:reg} (appendix). \newline

\noindent\textbf{Effect of training set size}
As discussed in Section \ref{sec:dataset} and Table \ref{data:exp-stat}, the size of our corpora is much smaller compared to other datasets such as CNN/Daily Mail, Wikilingua, and MLSUM. To investigate whether the limited amount of data for finetuning is the main cause of bad performance, we harvest additional instances for both directions. Specifically, we expand the existing datasets for hDe-En from 328 to 455 instances (38.6\% increase) and for hEn-De from 289 to 501 instances (73.4\% increase). The newly added sources for hDe-En mainly include 95 chapters from the philosophical novel \textit{Wilhelm Meister's Apprenticeship} published in 1795, 12 books of epic poetry \textit{Reineke Fuchs} published in 1794, and 15 scenes from Friedrich Schiller's play \textit{The robbers} published in 1781. For hEn-De, the new sources consist of 210 historical translations of fairy tales from different parts of the world published between 1885 and 1910 and 2 short stories.\footnote{The publication time of the newly acquired datasets for both directions closely aligns with that of the original dataset discussed in Section \ref{sec:dataset}.} We retrain the CLCTS baseline models using the expanded datasets for both directions, employing the same methods described in Section \ref{sec:setup}. To ensure comparability, we perform automatic evaluations on the same test sets. Table \ref{apdx:expansion} shows the results 
for both directions. For hDe-En, the baseline models exhibit marginal improvements, with a mean increase of 0.004 in ROUGE-1, 0.010 in ROUGE-L, and 0.008 in BERTScore-F1. Conversely for hEn-De, the model performances stay invariant on average according to ROUFE-L and BERTScore-F1 with a slight decrease according to ROUGE-1 (-0.004). We speculate that the decrease is because the original CLCTS hEn-De datasets contain more diverse sources (fairy tales, short stories, and plays) while the newly acquired training set primarily contains fairy tales. Moreover, as demonstrated in Section \ref{anslysis:corr}, the correlation between automatic evaluation metrics and human annotation is worse for hEn-De, which may also affect the evaluation. Overall, our experiments with a larger training set size only show marginal improvements to the existing baseline models. Based on this observation, we hypothesize that a greater volume of data might be necessary to effectively address the intricate nature of the CLCTS task. 

\subsection{Analysis on ChatGPT summarization}
\subsubsection{The effect of different prompts and target language}
\emph{A case study with ChatGPT-e2e vs.\ ChatGPT-pipeline}. Though in Section \ref{sec:results} we observe better metric and human annotation results from ChatGPT e2e prompt, we notice that ChatGPT under this prompt is prone to output summaries in the wrong target language (invalid output). In such cases, we query the summary with the same prompt until a valid output is returned. 
We calculate the occurrences of \emph{invalid} outputs after 2 rounds of queries and report them in Table \ref{dis:imbalance}, where we observe that using e2e prompts has a higher chance to yield invalid outputs. What is also worth noticing is when the output is expected in German, we observe more invalid cases among all three prompts which means that ChatGPT tends to output English texts when our expected language is in German. 
\begin{table}[!h]
\fontsize{7pt}{7pt}\selectfont
\begin{tabular}{lll}
\toprule
\textbf{Model} & \textbf{Tgt.lang} & \textbf{Invalid/Obs.} \\
\midrule
ChatGPT-title  & De & 1/328  \\
ChatGPT-pipeline & De &	34/328  \\ 
ChatGPT-e2e & De &	57/328   \\ 
ChatGPT-title  & En & 0/289  \\
ChatGPT-pipeline & En &	2/289  \\ 
ChatGPT-e2e & En &	8/289   \\ 
\bottomrule
\end{tabular}
\caption{\label{dis:imbalance}
Counts of invalid outputs from ChatGPT. Tgt.lang is the target language. Invalid/Obs. represents invalid cases/ total observation. }
\end{table} 
\subsubsection{ChatGPT for historical text translation}
We conduct an additional analysis and observe a performance gain of 5\% average over all metrics by simply switching the translator to ChatGPT. Part of the reason is that ChatGPT can translate better with prior knowledge, for example, ChatGPT keeps the word ``Schimmel'' (a special type of horse) unlike other translators translating to the meaning ``mold''. 
What is also worth noticing is that ChatGPT can better process historical variants such as historical spellings (e.g., \textfrak{*a, *o, *u}) and expressions which is very suitable for our setup. This could fill the gap of historical spelling normalization tools as discussed in Section \ref{sec:results}. 
\subsubsection{ChatGPT knows it all?} Recently, the prevalence of LLMs has raised the concern of data contamination \cite{balloccu-etal-2024-leak,sainz-etal-2023-nlp}. Since our corpus-building process is highly dependent on Wikipedia (and popular stories presumably discussed on other places on the web), we want to know to what extent ChatGPT has an unfair advantage. We ask ChatGPT to output summaries based on the author, title, and publication year. ChatGPT reports summaries to all requests in both language directions, though the output may contain irrelevant information besides the actual summarization.\footnote{For example, ``The story emphasizes the power of kindness and the importance of inner beauty. It also highlights the value of perseverance and the rewards that can come from staying true to oneself.''} 

However, despite the overall outstanding performance of ChatGPT summarization according to humans, ChatGPT evaluation, and automatic evaluation metrics, ChatGPT as a summarizer can also be prone to hallucinations like other NLG systems \cite{bang2023multitask, ji2023survey}. For example, in the summary of the famous story \textit{Cinderella}, we notice the golden slippers in the original document become glass slippers which is the best-known version worldwide but is incorrect based on the input text. \newline

\noindent\textbf{Adversarial attacks for ChatGPT}
Knowing that ChatGPT has specific prior knowledge of our CLCTS task, we go a step further and design three adversarial experiments inspired by the work from \citet{chen2022menli} where we gradually reduce the amount of specific prior knowledge in the input documents (from seen documents to pseudo-unseen documents to unseen documents). We aim to answer (1) does ChatGPT summarize truly based on the input text given its specific prior knowledge? (2) Can ChatGPT summarize against its specific prior knowledge? (3) How well can ChatGPT summarize facing unseen documents? 

\begin{figure}
    \centering
    \includegraphics[width=0.45\textwidth]{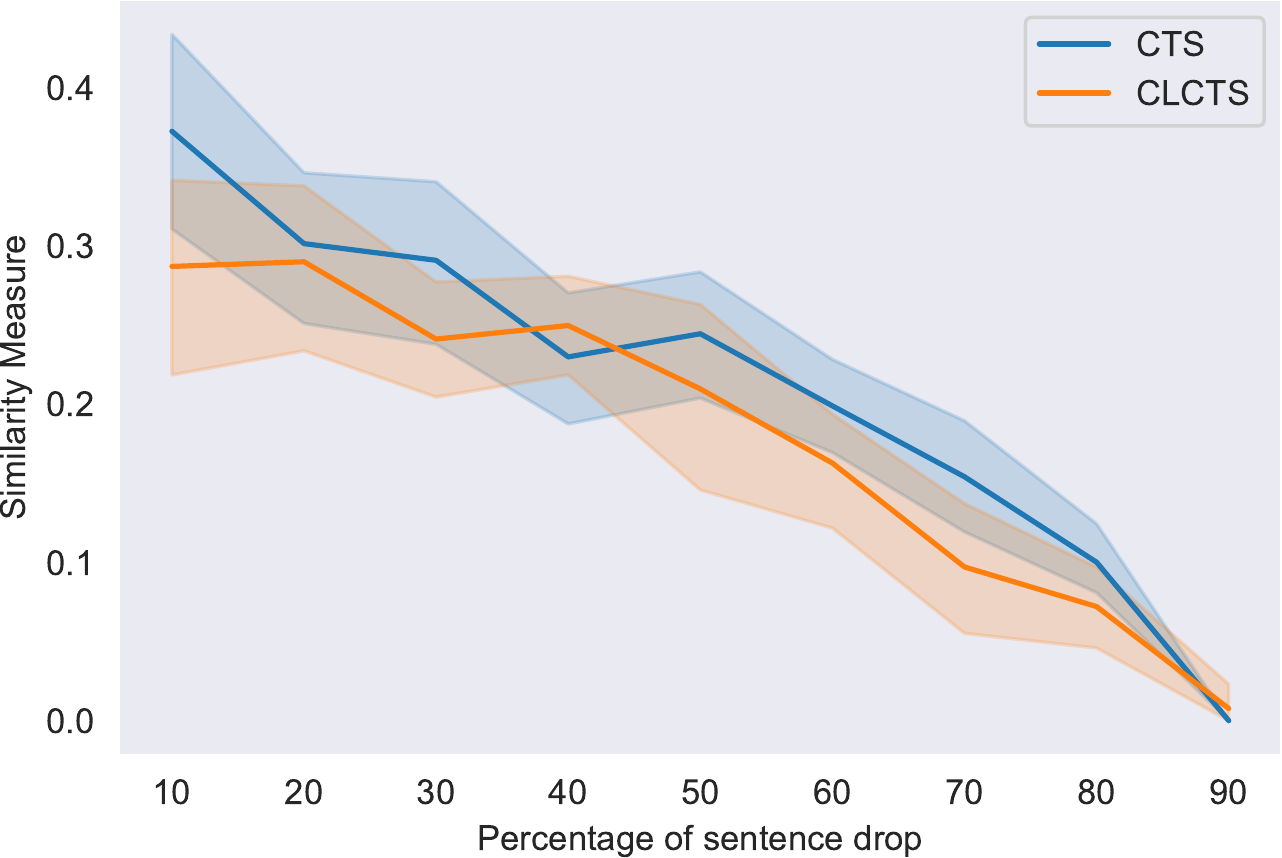}
    \caption{Similarity measure of generated summaries compared to the original summary over different percentages of sentence drop. We show an average score over all documents and present 95\% confidence interval.}
    \label{analysis:ratio}
\end{figure}

For question (1), we design a \emph{sentence omission task}. By dropping an increasing amount of sentences from the original documents, we control the information reduction of ChatGPT input for summarization. We randomly select 11 historical English documents from our CLCTS corpus containing 100-150 sentences and construct new documents by dropping sentences of varying percentages. We then use ChatGPT to summarize the new documents. To quantify the information change, we use the summary of the original documents (generated by ChatGPT) as the reference and evaluate generated summaries from documents after omission using automatic evaluation metrics.\footnote{We set the temperature to 0 since our goal is not to obtain the best summaries but to test ChatGPT summarization performance under gradual information loss.} We average the resulting metric scores after scaling and use it as our similarity measure. The result is shown in Figure \ref{analysis:ratio}. We observe an overall downward trend of similarity score of both CTS and CLCTS which reflects a certain level of information loss accordingly. 

For question (2), we first construct pseudo-unseen documents by entity swap, negation, and omission.\footnote{For each attack, we construct 6 documents. We list examples of adversarial attacks in Section \ref{apdx:example_adversarial}. For each document, we query the summaries 3 times at temperatures 0, 0.7, and 1 for both English and German. After excluding summaries in the wrong target language, we have 300 valid generated summaries.} 
\emph{Entity swap:} 
We swap multiple entities from the original story with new entities. For example, in the story of Cinderella, we switch the hazel tree into a cactus, and (to make it funny) we replace the prince with a watermelon (to ensure coherence, we also add background such as ``They live in the Fruit Kingdom.''). 
\emph{Negation:}
We alter the important aspects of the plot to make a `negated story'. For example, in the story of Cinderella, the stepmother and stepsisters are mistreated by the rich man's evil daughter and the prince eventually marries one of the stepsisters. 
\emph{Omission}:
For omission, we omit important plots. For example, Cinderella neither attends the ball nor marries the prince. We make sure each pseudo-unseen document only contains one kind of attack. We manually annotate the resulting output where we focus on whether the attack is completed accurately or not (e.g., if during negation, Cinderella is still mistreated by the stepmother --- as in the original story --- then the task is judged as a failure).\footnote{We obtain the inter-annotator agreement between two annotators of 0.7 according to Cohen’s kappa and 0.87 according to percentage accuracy based on 15 samples. We list examples of failure cases in Section \ref{failure} (appendix).} We compute the accuracy and report it in Table \ref{analysis:attack}.   
\begin{table}[h]
\fontsize{7pt}{7pt}\selectfont
\begin{tabular}{llll}
\toprule
\textbf{Attack} & \textbf{CTS} & \textbf{CLCTS} \\
\hline
Omission  & 0.79 & 0.67  \\
Entity Swap & 0.83 & 0.71 \\ 
Negation & 0.76 &	0.53 \\ 
\bottomrule
\end{tabular}
\caption{\label{analysis:attack}
Accuracy of different attacks for CTS and CLCTS tasks for ChatGPT.}
\end{table} 
We observe from the results that (1) ChatGPT can better handle omission and entity swap than negation. This may be because compared to omission and entity swap, negation attacks alter the underlying logical connections of the story and thus require more reasoning to complete this attack under the assumption that ChatGPT is affected by its explicit knowledge during summarization. (2) Higher accuracy for CTS than CLCTS task. For all three attacks, we observe a higher accuracy of completing the task for CTS. This is understandable since CLCTS also involves summarization cross-lingually which is more difficult. 

To further investigate question (3), we create 15 unseen documents from fiction chapters whose last updates are in the year 2022 or later\footnote{We include chapters from 5 Chinese novels and 10 English novels.}. For modern fiction chapters, we translate and rewrite the stories in historical languages (by adding historical language elements such as morphological changes) 
and write reference summaries in modern languages. Both documents and summaries are proofread by a co-author of this work. Similar to Section \ref{sec:results}, we evaluate the resulting summaries by three human annotators from four perspectives. The results are shown in Table \ref{ana:unseen}. 
We notice that the overall rating for CTS and CLCTS ranges between 3 and 4 indicating that ChatGPT summarizes unseen documents at a mediocre level which is similar to the results from seen documents. This rating over all four 
dimensions for hEn-De is slightly worse than the results from our previous experiment using CLCTS corpus (we list the result of the same direction in row 3 in Table \ref{ana:unseen}). We also query and annotate English summaries of unseen documents (CTS hEn-En) where we observe the overall performance is better than CLCTS hEn-De for unseen documents. This is expected since the CTS task does not involve cross-lingual summarization.  
However, we observe a decrease in summary quality of 0.75 points on average compared to the annotation results from CLCTS hDe-En task from our experiment. This is slightly surprising, since both target languages are in English, and especially CLCTS hDe-En has a more complex task setting. This may again indicate the benefits of the specific prior knowledge from ChatGPT.
\begin{table}[!htbp]
\fontsize{7pt}{7pt}\selectfont
\begin{tabular}[\textwidth]{lllll}
\toprule
\textbf{Models} & \textbf{Coh.} & \textbf{Con.} & \textbf{Flu.} & \textbf{Rel.} \\
\midrule
unseen CLCTS hEn-De & 3.24	& 2.79	& 3.06	& 2.85 \\ 
(our CLCTS hEn-De &  3.31 &  2.80 &  3.38 &  2.98)\\ 
\midrule
unseen CTS hEn-En & 3.44 & 3.12 & 3.44 & 3.29 \\
( our CLCTS hDe-En &   4.14 &  3.98 &  4.18 &  3.97) \\ 
\bottomrule
\end{tabular}
\caption{\label{ana:unseen}
Average human ratings for ChatGPT summarization on unseen documents with respect to coherence, consistency, fluency, and relevance. 
}
\end{table} 
Overall, we argue that ChatGPT can clearly outperform finetuned models for CLCTS tasks in our experiments. However, it does not ``magically'' solve all the issues, as per our experiments: (1) it profits from its prior knowledge to a certain degree where it may also blend its prior knowledge with the observed data and ChatGPT is better at omission and entity swap than negating against its prior knowledge; (2) the benefit from prior knowledge slightly inflates its assessed quality where ChatGPT generates summaries of mediocre quality from unseen source documents with a slightly worse performance over all four evaluating dimensions compared to seen documents for hEn-De; (3) the performance is language dependent where compared to German outputs (hEn-De), English outputs (hEn-En) yield slightly better ratings.
\section{Concluding remarks}\label{sec:conclusion}
In this work, we build the first CLCTS corpus and a historical translation dataset with additional information for future explorations. We study the characteristics of this corpus both alone and in comparison to other summarization corpora and examine the effectiveness of pipeline models, popular e2e transformer-based abstractive models leveraging intermediate task finetuning and ChatGPT for the CLCTS task. Even though these methods have been shown effective for single-task settings such as CLS or long document summarization, \textbf{they fail to deliver good summaries in our combined task setting (cross-lingual, cross-temporal, and long document)}. We showcase the potential of GPT-3.5 in CLCTS as a zero-shot summarizer. It provides moderate to good quality outputs and seems very adept as a context-aware spelling normalization tool. However, it profits from its prior knowledge to a certain degree as tested by adversarially attacked and unseen source documents. Overall, we observe better performance for plot omission and entity swap than plot negation against its prior knowledge. Moreover, GPT-3.5 performs slightly worse for unseen source documents compared to seen documents. 

As for evaluation, ChatGPT (both GPT-3.5 and GPT-4) as an evaluator can correlate with human evaluations at a moderate to good level but is prone to giving lower scores. Additionally, we find \textbf{a moderate correlation of BERTScore with human annotation (slightly worse for German)}. 

By regression analysis with BERTScore-F1, we quantify the effects of document features on model performance where we find positive impacts from documents with more recent publication years and document-summary embedding similarity while we observe a negative relation between model performance and the length of document. This helps explain model performance and affirms our hypotheses about the difficulty of the CLCTS task as discussed in the introduction (language change and divergence, etc.).

Future work can address the limitations revealed by our experiments. For example, researchers can explore other recent models of great potential, such as \zrf{U}nlimiformer under sufficient GPU memory \cite{bertsch2023unlimiformer} and Longnet \cite{ding2023longnet}. Other training frameworks such as pretraining under multi-task learning are also promising alternatives. Incorporating tasks such as non-parallel historical corpora for pretraining and the PAR3 cross-temporal dataset for machine translation \cite{thai-etal-2022-exploring} may further provide benefits. Finally, 
future work can focus on extending the size and diversity of our CLCTS datasets, including more diverse language pairs.\footnote{\zrf{Our current work builds the CLCTS corpus for the high-resource language pair German and English. Main difficulties working with other languages include: (1) Even more Wikipedia summaries may be missing for other languages (German and English are two of the highest resource languages); (2) it may be more difficult to obtain historical sources of other languages; (3) human evaluation of other languages may be less accessible and thus even more costly; (4) automatic metrics including GPT4 may also be worse as we have observed worse performance for German than for English. While exploring more distant language pairs is thus immensely interesting, (5) our results further indicate that transfer would be even more difficult for all models involved.
}}

\section{Acknowledgments}
We thank all annotators for their hard work and time. 
The NLLG group gratefully acknowledges support from the Federal Ministry of Education and Research (BMBF) via the grant ``Metrics4NLG'' and the German Research Foundation (DFG) via the Heisenberg grant EG375/5-1.

\clearpage
\starttwocolumn
\bibliography{compling_style}

\begin{thebibliography}{103}
\expandafter\ifx\csname natexlab\endcsname\relax\def\natexlab#1{#1}\fi

\bibitem[{Adesam, Ahlberg, and Bouma(2012)}]{adesam2012bokstaffua}
Adesam, Yvonne, Malin Ahlberg, and Gerlof Bouma. 2012.
\newblock bokstaffua, bokstaffwa, bokstafwa, bokstaua, bokstawa... towards
  lexical link-up for a corpus of old swedish.
\newblock In \emph{KONVENS}, pages 365--369.

\bibitem[{Agarwal, Vijay et~al.(2021)}]{agarwal2021genre}
Agarwal, Divya, Devika Vijay, et~al. 2021.
\newblock Genre classification using character networks.
\newblock In \emph{2021 5th International Conference on Intelligent Computing
  and Control Systems (ICICCS)}, pages 216--222, IEEE.

\bibitem[{Bai, Gao, and Huang(2021)}]{bai-etal-2021-cross}
Bai, Yu, Yang Gao, and Heyan Huang. 2021.
\newblock Cross-lingual abstractive summarization with limited parallel
  resources.
\newblock In \emph{Proceedings of the 59th Annual Meeting of the Association
  for Computational Linguistics and the 11th International Joint Conference on
  Natural Language Processing (Volume 1: Long Papers)}, pages 6910--6924,
  Association for Computational Linguistics, Online.

\bibitem[{Bai et~al.(2022)Bai, Huang, Fan, Gao, Zhu, Zhan, Chi, and
  Chen}]{bai2022unifying}
Bai, Yu, Heyan Huang, Kai Fan, Yang Gao, Yiming Zhu, Jiaao Zhan, Zewen Chi, and
  Boxing Chen. 2022.
\newblock Unifying cross-lingual summarization and machine translation with
  compression rate.
\newblock In \emph{Proceedings of the 45th International ACM SIGIR Conference
  on Research and Development in Information Retrieval}, pages 1087--1097.

\bibitem[{Balloccu et~al.(2024)Balloccu, Schmidtov{\'a}, Lango, and
  Dusek}]{balloccu-etal-2024-leak}
Balloccu, Simone, Patr{\'\i}cia Schmidtov{\'a}, Mateusz Lango, and Ondrej
  Dusek. 2024.
\newblock Leak, cheat, repeat: Data contamination and evaluation malpractices
  in closed-source {LLM}s.
\newblock In \emph{Proceedings of the 18th Conference of the European Chapter
  of the Association for Computational Linguistics (Volume 1: Long Papers)},
  pages 67--93, Association for Computational Linguistics, St. Julian{'}s,
  Malta.

\bibitem[{Bang et~al.(2023)Bang, Cahyawijaya, Lee, Dai, Su, Wilie, Lovenia, Ji,
  Yu, Chung, Do, Xu, and Fung}]{bang2023multitask}
Bang, Yejin, Samuel Cahyawijaya, Nayeon Lee, Wenliang Dai, Dan Su, Bryan Wilie,
  Holy Lovenia, Ziwei Ji, Tiezheng Yu, Willy Chung, Quyet~V. Do, Yan Xu, and
  Pascale Fung. 2023.
\newblock A multitask, multilingual, multimodal evaluation of {C}hat{GPT} on
  reasoning, hallucination, and interactivity.

\bibitem[{Baron and Rayson(2008)}]{baron2008vard2}
Baron, Alistair and Paul Rayson. 2008.
\newblock Vard2: A tool for dealing with spelling variation in historical
  corpora.
\newblock In \emph{Postgraduate conference in corpus linguistics}.

\bibitem[{Belouadi and Eger(2023)}]{belouadi-eger-2023-uscore}
Belouadi, Jonas and Steffen Eger. 2023.
\newblock {US}core: An effective approach to fully unsupervised evaluation
  metrics for machine translation.
\newblock In \emph{Proceedings of the 17th Conference of the European Chapter
  of the Association for Computational Linguistics}, pages 358--374,
  Association for Computational Linguistics, Dubrovnik, Croatia.

\bibitem[{Beltagy, Peters, and Cohan(2020)}]{beltagy2020longformer}
Beltagy, Iz, Matthew~E Peters, and Arman Cohan. 2020.
\newblock Longformer: The long-document transformer.
\newblock \emph{arXiv preprint arXiv:2004.05150}.

\bibitem[{Belz et~al.(2021)Belz, Shimorina, Agarwal, and
  Reiter}]{belz-etal-2021-reprogen}
Belz, Anya, Anastasia Shimorina, Shubham Agarwal, and Ehud Reiter. 2021.
\newblock The {R}epro{G}en shared task on reproducibility of human evaluations
  in {NLG}: Overview and results.
\newblock In \emph{Proceedings of the 14th International Conference on Natural
  Language Generation}, pages 249--258, Association for Computational
  Linguistics, Aberdeen, Scotland, UK.

\bibitem[{Belz, Thomson, and Reiter(2023)}]{belz-etal-2023-missing}
Belz, Anya, Craig Thomson, and Ehud Reiter. 2023.
\newblock Missing information, unresponsive authors, experimental flaws: The
  impossibility of assessing the reproducibility of previous human evaluations
  in {NLP}.
\newblock In \emph{The Fourth Workshop on Insights from Negative Results in
  NLP}, pages 1--10, Association for Computational Linguistics, Dubrovnik,
  Croatia.

\bibitem[{Bertsch et~al.(2024)Bertsch, Alon, Neubig, and
  Gormley}]{bertsch2023unlimiformer}
Bertsch, Amanda, Uri Alon, Graham Neubig, and Matthew Gormley. 2024.
\newblock Unlimiformer: Long-range transformers with unlimited length input.
\newblock \emph{Advances in Neural Information Processing Systems}, 36.

\bibitem[{Bhandari et~al.(2020)Bhandari, Gour, Ashfaq, Liu, and
  Neubig}]{bhandari-etal-2020-evaluating}
Bhandari, Manik, Pranav~Narayan Gour, Atabak Ashfaq, Pengfei Liu, and Graham
  Neubig. 2020.
\newblock Re-evaluating evaluation in text summarization.
\newblock In \emph{Proceedings of the 2020 Conference on Empirical Methods in
  Natural Language Processing (EMNLP)}, pages 9347--9359, Association for
  Computational Linguistics, Online.

\bibitem[{Bollmann(2012)}]{bollmann2012automatic}
Bollmann, Marcel. 2012.
\newblock automatic normalization of historical texts using distance measures
  and the norma tool.
\newblock In \emph{Proceedings of the second workshop on annotation of corpora
  for research in the humanities (ACRH-2), Lisbon, Portugal}, pages 3--14.

\bibitem[{Bollmann(2019)}]{bollmann-2019-large}
Bollmann, Marcel. 2019.
\newblock A large-scale comparison of historical text normalization systems.
\newblock In \emph{Proceedings of the 2019 Conference of the North {A}merican
  Chapter of the Association for Computational Linguistics: Human Language
  Technologies, Volume 1 (Long and Short Papers)}, pages 3885--3898,
  Association for Computational Linguistics, Minneapolis, Minnesota.

\bibitem[{Bollmann, Bingel, and S{\o}gaard(2017)}]{bollmann2017learning}
Bollmann, Marcel, Joachim Bingel, and Anders S{\o}gaard. 2017.
\newblock Learning attention for historical text normalization by learning to
  pronounce.
\newblock In \emph{Proceedings of the 55th Annual Meeting of the Association
  for Computational Linguistics (Volume 1: Long Papers)}, pages 332--344.

\bibitem[{Bollmann and S{\o}gaard(2016)}]{bollmann-sogaard-2016-improving}
Bollmann, Marcel and Anders S{\o}gaard. 2016.
\newblock Improving historical spelling normalization with bi-directional
  {LSTM}s and multi-task learning.
\newblock In \emph{Proceedings of {COLING} 2016, the 26th International
  Conference on Computational Linguistics: Technical Papers}, pages 131--139,
  The COLING 2016 Organizing Committee, Osaka, Japan.

\bibitem[{Brown et~al.(2020)Brown, Mann, Ryder, Subbiah, Kaplan, Dhariwal,
  Neelakantan, Shyam, Sastry, Askell et~al.}]{brown2020language}
Brown, Tom, Benjamin Mann, Nick Ryder, Melanie Subbiah, Jared~D Kaplan,
  Prafulla Dhariwal, Arvind Neelakantan, Pranav Shyam, Girish Sastry, Amanda
  Askell, et~al. 2020.
\newblock Language models are few-shot learners.
\newblock \emph{Advances in neural information processing systems},
  33:1877--1901.

\bibitem[{Cao, Liu, and Wan(2020)}]{cao-etal-2020-jointly}
Cao, Yue, Hui Liu, and Xiaojun Wan. 2020.
\newblock Jointly learning to align and summarize for neural cross-lingual
  summarization.
\newblock In \emph{Proceedings of the 58th Annual Meeting of the Association
  for Computational Linguistics}, pages 6220--6231, Association for
  Computational Linguistics, Online.

\bibitem[{Chang and Lu(2021)}]{chang-lu-2021-rethinking-intermediate}
Chang, Ting-Yun and Chi-Jen Lu. 2021.
\newblock Rethinking why intermediate-task fine-tuning works.
\newblock In \emph{Findings of the Association for Computational Linguistics:
  EMNLP 2021}, pages 706--713, Association for Computational Linguistics, Punta
  Cana, Dominican Republic.

\bibitem[{Chen, Li, and King(2021)}]{chen-etal-2021-training}
Chen, Wang, Piji Li, and Irwin King. 2021.
\newblock A training-free and reference-free summarization evaluation metric
  via centrality-weighted relevance and self-referenced redundancy.
\newblock In \emph{Proceedings of the 59th Annual Meeting of the Association
  for Computational Linguistics and the 11th International Joint Conference on
  Natural Language Processing (Volume 1: Long Papers)}, pages 404--414,
  Association for Computational Linguistics, Online.

\bibitem[{Chen, Belouadi, and Eger(2022)}]{chen-etal-2022-reproducibility}
Chen, Yanran, Jonas Belouadi, and Steffen Eger. 2022.
\newblock Reproducibility issues for {BERT}-based evaluation metrics.
\newblock In \emph{Proceedings of the 2022 Conference on Empirical Methods in
  Natural Language Processing}, pages 2965--2989, Association for Computational
  Linguistics, Abu Dhabi, United Arab Emirates.

\bibitem[{Chen and Eger(2023)}]{chen2022menli}
Chen, Yanran and Steffen Eger. 2023.
\newblock Menli: Robust evaluation metrics from natural language inference.
\newblock \emph{Transactions of the Association for Computational Linguistics},
  11:804--825.

\bibitem[{Chiang and Lee(2023)}]{chiang2023large}
Chiang, Cheng-Han and Hung-Yi Lee. 2023.
\newblock Can large language models be an alternative to human evaluations?

\bibitem[{Cui and Hu(2021)}]{cui-hu-2021-sliding}
Cui, Peng and Le~Hu. 2021.
\newblock Sliding selector network with dynamic memory for extractive
  summarization of long documents.
\newblock In \emph{Proceedings of the 2021 Conference of the North American
  Chapter of the Association for Computational Linguistics: Human Language
  Technologies}, pages 5881--5891, Association for Computational Linguistics,
  Online.

\bibitem[{Ding et~al.(2023)Ding, Ma, Dong, Zhang, Huang, Wang, and
  Wei}]{ding2023longnet}
Ding, Jiayu, Shuming Ma, Li~Dong, Xingxing Zhang, Shaohan Huang, Wenhui Wang,
  and Furu Wei. 2023.
\newblock Longnet: Scaling transformers to 1,000,000,000 tokens.

\bibitem[{Eger, vor~der Br{\"u}ck, and Mehler(2016)}]{eger2016comparison}
Eger, Steffen, Tim vor~der Br{\"u}ck, and Alexander Mehler. 2016.
\newblock A comparison of four character-level string-to-string translation
  models for (ocr) spelling error correction.
\newblock \emph{The Prague bulletin of mathematical linguistics}, 105(1):77.

\bibitem[{Eger and Mehler(2016)}]{eger-mehler-2016-linearity}
Eger, Steffen and Alexander Mehler. 2016.
\newblock On the linearity of semantic change: Investigating meaning variation
  via dynamic graph models.
\newblock In \emph{Proceedings of the 54th Annual Meeting of the Association
  for Computational Linguistics (Volume 2: Short Papers)}, pages 52--58,
  Association for Computational Linguistics, Berlin, Germany.

\bibitem[{Fabbri et~al.(2021)Fabbri, Kry{\'s}ci{\'n}ski, McCann, Xiong, Socher,
  and Radev}]{fabbri2021summeval}
Fabbri, Alexander~R, Wojciech Kry{\'s}ci{\'n}ski, Bryan McCann, Caiming Xiong,
  Richard Socher, and Dragomir Radev. 2021.
\newblock Summeval: Re-evaluating summarization evaluation.

\bibitem[{Fan et~al.(2021)Fan, Bhosale, Schwenk, Ma, El-Kishky, Goyal, Baines,
  Celebi, Wenzek, Chaudhary, Goyal, Birch, Liptchinsky, Edunov, Grave, Auli,
  and Joulin}]{fanbeyond}
Fan, Angela, Shruti Bhosale, Holger Schwenk, Zhiyi Ma, Ahmed El-Kishky,
  Siddharth Goyal, Mandeep Baines, Onur Celebi, Guillaume Wenzek, Vishrav
  Chaudhary, Naman Goyal, Tom Birch, Vitaliy Liptchinsky, Sergey Edunov,
  Edouard Grave, Michael Auli, and Armand Joulin. 2021.
\newblock Beyond english-centric multilingual machine translation.
\newblock \emph{J. Mach. Learn. Res.}, 22(1).

\bibitem[{Gao et~al.(2023)Gao, Ruan, Sun, Yin, Yang, and Wan}]{gao2023human}
Gao, Mingqi, Jie Ruan, Renliang Sun, Xunjian Yin, Shiping Yang, and Xiaojun
  Wan. 2023.
\newblock Human-like summarization evaluation with chatgpt.
\newblock \emph{arXiv preprint arXiv:2304.02554}.

\bibitem[{Gao, Zhao, and Eger(2020)}]{gao-etal-2020-supert}
Gao, Yang, Wei Zhao, and Steffen Eger. 2020.
\newblock {SUPERT}: Towards new frontiers in unsupervised evaluation metrics
  for multi-document summarization.
\newblock In \emph{Proceedings of the 58th Annual Meeting of the Association
  for Computational Linguistics}, pages 1347--1354, Association for
  Computational Linguistics, Online.

\bibitem[{Gibson et~al.(2019)Gibson, Futrell, Piantadosi, Dautriche, Mahowald,
  Bergen, and Levy}]{gibson_how_2019}
Gibson, Edward, Richard Futrell, Steven~P. Piantadosi, Isabelle Dautriche, Kyle
  Mahowald, Leon Bergen, and Roger Levy. 2019.
\newblock How {Efficiency} {Shapes} {Human} {Language}.
\newblock \emph{Trends in Cognitive Sciences}, 23(5):389--407.

\bibitem[{Gidiotis and Tsoumakas(2020)}]{hybrid-longdoc-divide}
Gidiotis, Alexios and Grigorios Tsoumakas. 2020.
\newblock A divide-and-conquer approach to the summarization of long documents.
\newblock \emph{IEEE/ACM Transactions on Audio, Speech, and Language
  Processing}, 28:3029--3040.

\bibitem[{Giulianelli, Del~Tredici, and
  Fern{\'a}ndez(2020)}]{giulianelli-etal-2020-analysing}
Giulianelli, Mario, Marco Del~Tredici, and Raquel Fern{\'a}ndez. 2020.
\newblock Analysing lexical semantic change with contextualised word
  representations.
\newblock In \emph{Proceedings of the 58th Annual Meeting of the Association
  for Computational Linguistics}, pages 3960--3973, Association for
  Computational Linguistics, Online.

\bibitem[{Goyal, Li, and Durrett(2022{\natexlab{a}})}]{goyal_news_2022}
Goyal, Tanya, Junyi~Jessy Li, and Greg Durrett. 2022{\natexlab{a}}.
\newblock News summarization and evaluation in the era of gpt-3.

\bibitem[{Goyal, Li, and Durrett(2022{\natexlab{b}})}]{goyal-etal-2022-snac}
Goyal, Tanya, Junyi~Jessy Li, and Greg Durrett. 2022{\natexlab{b}}.
\newblock {SN}a{C}: Coherence error detection for narrative summarization.
\newblock In \emph{Proceedings of the 2022 Conference on Empirical Methods in
  Natural Language Processing}, pages 444--463, Association for Computational
  Linguistics, Abu Dhabi, United Arab Emirates.

\bibitem[{Grusky, Naaman, and Artzi(2018)}]{grusky-etal-2018-newsroom}
Grusky, Max, Mor Naaman, and Yoav Artzi. 2018.
\newblock {N}ewsroom: A dataset of 1.3 million summaries with diverse
  extractive strategies.
\newblock In \emph{Proceedings of the 2018 Conference of the North {A}merican
  Chapter of the Association for Computational Linguistics: Human Language
  Technologies, Volume 1 (Long Papers)}, pages 708--719, Association for
  Computational Linguistics, New Orleans, Louisiana.

\bibitem[{Gu, Ash, and Hahnloser(2022)}]{gu-etal-2022-memsum}
Gu, Nianlong, Elliott Ash, and Richard Hahnloser. 2022.
\newblock {M}em{S}um: Extractive summarization of long documents using
  multi-step episodic {M}arkov decision processes.
\newblock In \emph{Proceedings of the 60th Annual Meeting of the Association
  for Computational Linguistics (Volume 1: Long Papers)}, pages 6507--6522,
  Association for Computational Linguistics, Dublin, Ireland.

\bibitem[{Hamilton, Leskovec, and
  Jurafsky(2016)}]{hamilton-etal-2016-diachronic}
Hamilton, William~L., Jure Leskovec, and Dan Jurafsky. 2016.
\newblock Diachronic word embeddings reveal statistical laws of semantic
  change.
\newblock In \emph{Proceedings of the 54th Annual Meeting of the Association
  for Computational Linguistics (Volume 1: Long Papers)}, pages 1489--1501,
  Association for Computational Linguistics, Berlin, Germany.

\bibitem[{He et~al.(2023)He, Peng, Wang, Liu, Xu, Hassan, Shi, Zhu, Xiong,
  Zeng, Gao, and Huang}]{he2023zcode}
He, Pengcheng, Baolin Peng, Song Wang, Yang Liu, Ruochen Xu, Hany Hassan,
  Yu~Shi, Chenguang Zhu, Wayne Xiong, Michael Zeng, Jianfeng Gao, and Xuedong
  Huang. 2023.
\newblock {Z}-code++: A pre-trained language model optimized for abstractive
  summarization.
\newblock In \emph{Proceedings of the 61st Annual Meeting of the Association
  for Computational Linguistics (Volume 1: Long Papers)}, pages 5095--5112,
  Association for Computational Linguistics, Toronto, Canada.

\bibitem[{Huang et~al.(2021)Huang, Cao, Parulian, Ji, and
  Wang}]{huang-etal-2021-efficient}
Huang, Luyang, Shuyang Cao, Nikolaus Parulian, Heng Ji, and Lu~Wang. 2021.
\newblock Efficient attentions for long document summarization.
\newblock In \emph{Proceedings of the 2021 Conference of the North American
  Chapter of the Association for Computational Linguistics: Human Language
  Technologies}, pages 1419--1436, Association for Computational Linguistics,
  Online.

\bibitem[{Ji et~al.(2023)Ji, Lee, Frieske, Yu, Su, Xu, Ishii, Bang, Madotto,
  and Fung}]{ji2023survey}
Ji, Ziwei, Nayeon Lee, Rita Frieske, Tiezheng Yu, Dan Su, Yan Xu, Etsuko Ishii,
  Ye~Jin Bang, Andrea Madotto, and Pascale Fung. 2023.
\newblock Survey of hallucination in natural language generation.
\newblock \emph{ACM Computing Surveys}, 55(12):1--38.

\bibitem[{Joseph(2017)}]{joseph2017diachronic}
Joseph, Brian~D. 2017.
\newblock Diachronic morphology.
\newblock \emph{The handbook of morphology}, pages 349--373.

\bibitem[{Juzek, Krielke, and Teich(2020)}]{juzek-etal-2020-exploring}
Juzek, Tom~S, Marie-Pauline Krielke, and Elke Teich. 2020.
\newblock Exploring diachronic syntactic shifts with dependency length: the
  case of scientific {E}nglish.
\newblock In \emph{Proceedings of the Fourth Workshop on Universal Dependencies
  (UDW 2020)}, pages 109--119, Association for Computational Linguistics,
  Barcelona, Spain (Online).

\bibitem[{Karpinska and Iyyer(2023)}]{karpinska-iyyer-2023-large}
Karpinska, Marzena and Mohit Iyyer. 2023.
\newblock Large language models effectively leverage document-level context for
  literary translation, but critical errors persist.
\newblock In \emph{Proceedings of the Eighth Conference on Machine
  Translation}, pages 419--451, Association for Computational Linguistics,
  Singapore.

\bibitem[{Koh et~al.(2022)Koh, Ju, Liu, and Pan}]{koh2022empirical}
Koh, Huan~Yee, Jiaxin Ju, Ming Liu, and Shirui Pan. 2022.
\newblock An empirical survey on long document summarization: Datasets, models,
  and metrics.
\newblock \emph{ACM computing surveys}, 55(8):1--35.

\bibitem[{Laban et~al.(2022)Laban, Schnabel, Bennett, and
  Hearst}]{laban-etal-2022-summac}
Laban, Philippe, Tobias Schnabel, Paul~N. Bennett, and Marti~A. Hearst. 2022.
\newblock {S}umma{C}: Re-visiting {NLI}-based models for inconsistency
  detection in summarization.
\newblock \emph{Transactions of the Association for Computational Linguistics},
  10:163--177.

\bibitem[{Ladhak et~al.(2020)Ladhak, Durmus, Cardie, and
  McKeown}]{ladhak-etal-2020-wikilingua}
Ladhak, Faisal, Esin Durmus, Claire Cardie, and Kathleen McKeown. 2020.
\newblock {W}iki{L}ingua: A new benchmark dataset for cross-lingual abstractive
  summarization.
\newblock In \emph{Findings of the Association for Computational Linguistics:
  EMNLP 2020}, pages 4034--4048, Association for Computational Linguistics,
  Online.

\bibitem[{Lei and Wen(2020)}]{lei_is_2020}
Lei, Lei and Ju~Wen. 2020.
\newblock Is dependency distance experiencing a process of minimization? {A}
  diachronic study based on the {State} of the {Union} addresses.
\newblock \emph{Lingua}, 239:102762.

\bibitem[{Leiter et~al.(2024)Leiter, Zhang, Chen, Belouadi, Larionov, Fresen,
  and Eger}]{Leiter2023ChatGPTAM}
Leiter, Christoph, Ran Zhang, Yanran Chen, Jonas Belouadi, Daniil Larionov,
  Vivian Fresen, and Steffen Eger. 2024.
\newblock Chatgpt: A meta-analysis after 2.5 months.
\newblock \emph{Machine Learning with Applications}, 16:100541.

\bibitem[{Lewis(1894)}]{lewis1894history}
Lewis, Edwin~Herbert. 1894.
\newblock \emph{The history of the English paragraph}.
\newblock 2. AMS Press.

\bibitem[{Lewis et~al.(2020)Lewis, Liu, Goyal, Ghazvininejad, Mohamed, Levy,
  Stoyanov, and Zettlemoyer}]{lewis-etal-2020-bart}
Lewis, Mike, Yinhan Liu, Naman Goyal, Marjan Ghazvininejad, Abdelrahman
  Mohamed, Omer Levy, Veselin Stoyanov, and Luke Zettlemoyer. 2020.
\newblock {BART}: Denoising sequence-to-sequence pre-training for natural
  language generation, translation, and comprehension.
\newblock In \emph{Proceedings of the 58th Annual Meeting of the Association
  for Computational Linguistics}, pages 7871--7880, Association for
  Computational Linguistics, Online.

\bibitem[{Liang et~al.(2022)Liang, Meng, Zhou, Xu, Chen, Su, and
  Zhou}]{liang-etal-2022-variational}
Liang, Yunlong, Fandong Meng, Chulun Zhou, Jinan Xu, Yufeng Chen, Jinsong Su,
  and Jie Zhou. 2022.
\newblock A variational hierarchical model for neural cross-lingual
  summarization.
\newblock In \emph{Proceedings of the 60th Annual Meeting of the Association
  for Computational Linguistics (Volume 1: Long Papers)}, pages 2088--2099,
  Association for Computational Linguistics, Dublin, Ireland.

\bibitem[{Lin(2004)}]{lin-2004-rouge}
Lin, Chin-Yew. 2004.
\newblock {ROUGE}: A package for automatic evaluation of summaries.
\newblock In \emph{Text Summarization Branches Out}, pages 74--81, Association
  for Computational Linguistics, Barcelona, Spain.

\bibitem[{Liu(2008)}]{liu2008dependency}
Liu, Haitao. 2008.
\newblock Dependency distance as a metric of language comprehension difficulty.
\newblock \emph{Journal of Cognitive Science}, 9(2):159--191.

\bibitem[{Liu, Xu, and Liang(2017)}]{liu2017dependency}
Liu, Haitao, Chunshan Xu, and Junying Liang. 2017.
\newblock Dependency distance: A new perspective on syntactic patterns in
  natural languages.
\newblock \emph{Physics of life reviews}, 21:171--193.

\bibitem[{Liu et~al.(2023)Liu, Iter, Xu, Wang, Xu, and Zhu}]{liu2023geval}
Liu, Yang, Dan Iter, Yichong Xu, Shuohang Wang, Ruochen Xu, and Chenguang Zhu.
  2023.
\newblock {G}-eval: {NLG} evaluation using gpt-4 with better human alignment.
\newblock In \emph{Proceedings of the 2023 Conference on Empirical Methods in
  Natural Language Processing}, pages 2511--2522, Association for Computational
  Linguistics, Singapore.

\bibitem[{Liu et~al.(2022)Liu, Liu, Radev, and Neubig}]{liu-etal-2022-brio}
Liu, Yixin, Pengfei Liu, Dragomir Radev, and Graham Neubig. 2022.
\newblock {BRIO}: Bringing order to abstractive summarization.
\newblock In \emph{Proceedings of the 60th Annual Meeting of the Association
  for Computational Linguistics (Volume 1: Long Papers)}, pages 2890--2903,
  Association for Computational Linguistics, Dublin, Ireland.

\bibitem[{Liu, Jia, and Zhu(2022)}]{liu-etal-2022-reference}
Liu, Yizhu, Qi~Jia, and Kenny Zhu. 2022.
\newblock Reference-free summarization evaluation via semantic correlation and
  compression ratio.
\newblock In \emph{Proceedings of the 2022 Conference of the North American
  Chapter of the Association for Computational Linguistics: Human Language
  Technologies}, pages 2109--2115, Association for Computational Linguistics,
  Seattle, United States.

\bibitem[{Makarov and Clematide(2020)}]{makarov-clematide-2020-semi}
Makarov, Peter and Simon Clematide. 2020.
\newblock Semi-supervised contextual historical text normalization.
\newblock In \emph{Proceedings of the 58th Annual Meeting of the Association
  for Computational Linguistics}, pages 7284--7295, Association for
  Computational Linguistics, Online.

\bibitem[{Manakul and Gales(2021)}]{manakul-gales-2021-long}
Manakul, Potsawee and Mark Gales. 2021.
\newblock Long-span summarization via local attention and content selection.
\newblock In \emph{Proceedings of the 59th Annual Meeting of the Association
  for Computational Linguistics and the 11th International Joint Conference on
  Natural Language Processing (Volume 1: Long Papers)}, pages 6026--6041,
  Association for Computational Linguistics, Online.

\bibitem[{Niwattanakul et~al.(2013)Niwattanakul, Singthongchai, Naenudorn, and
  Wanapu}]{niwattanakul2013Jaccard}
Niwattanakul, Suphakit, Jatsada Singthongchai, Ekkachai Naenudorn, and
  Supachanun Wanapu. 2013.
\newblock Using of jaccard coefficient for keywords similarity.
\newblock In \emph{Proceedings of the international multiconference of
  engineers and computer scientists}, volume~1, pages 380--384.

\bibitem[{Ouyang, Song, and McKeown(2019)}]{ouyang-etal-2019-robust}
Ouyang, Jessica, Boya Song, and Kathy McKeown. 2019.
\newblock A robust abstractive system for cross-lingual summarization.
\newblock In \emph{Proceedings of the 2019 Conference of the North {A}merican
  Chapter of the Association for Computational Linguistics: Human Language
  Technologies, Volume 1 (Long and Short Papers)}, pages 2025--2031,
  Association for Computational Linguistics, Minneapolis, Minnesota.

\bibitem[{Peng et~al.(2021)Peng, Zheng, Lin, and Siddharthan}]{2021summarising}
Peng, Xutan, Yi~Zheng, Chenghua Lin, and Advaith Siddharthan. 2021.
\newblock Summarising historical text in modern languages.
\newblock In \emph{Proceedings of the 16th Conference of the European Chapter
  of the Association for Computational Linguistics: Main Volume}, pages
  3123--3142, Association for Computational Linguistics, Online.

\bibitem[{Pettersson(2016)}]{pettersson2016spelling}
Pettersson, Eva. 2016.
\newblock \emph{Spelling normalisation and linguistic analysis of historical
  text for information extraction}.
\newblock Ph.D. thesis, Acta Universitatis Upsaliensis.

\bibitem[{Peyrard(2019)}]{peyrard-2019-studying}
Peyrard, Maxime. 2019.
\newblock Studying summarization evaluation metrics in the appropriate scoring
  range.
\newblock In \emph{Proceedings of the 57th Annual Meeting of the Association
  for Computational Linguistics}, pages 5093--5100, Association for
  Computational Linguistics, Florence, Italy.

\bibitem[{Pilault et~al.(2020)Pilault, Li, Subramanian, and
  Pal}]{pilault-etal-2020-extractiveabs}
Pilault, Jonathan, Raymond Li, Sandeep Subramanian, and Chris Pal. 2020.
\newblock On extractive and abstractive neural document summarization with
  transformer language models.
\newblock In \emph{Proceedings of the 2020 Conference on Empirical Methods in
  Natural Language Processing (EMNLP)}, pages 9308--9319, Association for
  Computational Linguistics, Online.

\bibitem[{Popovi{\'c}(2015)}]{popovic-2015-chrf}
Popovi{\'c}, Maja. 2015.
\newblock chr{F}: character n-gram {F}-score for automatic {MT} evaluation.
\newblock In \emph{Proceedings of the Tenth Workshop on Statistical Machine
  Translation}, pages 392--395, Association for Computational Linguistics,
  Lisbon, Portugal.

\bibitem[{Porta, Sancho, and G{\'o}mez(2013)}]{porta2013edit}
Porta, Jordi, Jos{\'e}-Luis Sancho, and Javier G{\'o}mez. 2013.
\newblock Edit transducers for spelling variation in old spanish.
\newblock In \emph{Proc. of the workshop on computational historical
  linguistics at NODALIDA 2013. NEALT Proc. Series}, volume~18, pages 70--79.

\bibitem[{Qi et~al.(2020)Qi, Zhang, Zhang, Bolton, and
  Manning}]{qi-etal-2020-stanza}
Qi, Peng, Yuhao Zhang, Yuhui Zhang, Jason Bolton, and Christopher~D. Manning.
  2020.
\newblock {S}tanza: A python natural language processing toolkit for many human
  languages.
\newblock In \emph{Proceedings of the 58th Annual Meeting of the Association
  for Computational Linguistics: System Demonstrations}, pages 101--108,
  Association for Computational Linguistics, Online.

\bibitem[{Ravaut, Joty, and Chen(2022)}]{ravaut-etal-2022-summareranker}
Ravaut, Mathieu, Shafiq Joty, and Nancy Chen. 2022.
\newblock {S}umma{R}eranker: A multi-task mixture-of-experts re-ranking
  framework for abstractive summarization.
\newblock In \emph{Proceedings of the 60th Annual Meeting of the Association
  for Computational Linguistics (Volume 1: Long Papers)}, pages 4504--4524,
  Association for Computational Linguistics, Dublin, Ireland.

\bibitem[{Rayson, Archer, and Smith(2005)}]{rayson2005vard}
Rayson, Paul, Dawn Archer, and Nicholas Smith. 2005.
\newblock Vard versus word: A comparison of the ucrel variant detector and
  modern spellcheckers on english historical corpora.
\newblock \emph{Corpus Linguistics 2005}.

\bibitem[{Reimers and Gurevych(2020)}]{reimers-2020-multilingual-sentence-bert}
Reimers, Nils and Iryna Gurevych. 2020.
\newblock Making monolingual sentence embeddings multilingual using knowledge
  distillation.
\newblock In \emph{Proceedings of the 2020 Conference on Empirical Methods in
  Natural Language Processing (EMNLP)}, pages 4512--4525.

\bibitem[{Robertson and Goldwater(2018)}]{robertson2018evaluating}
Robertson, Alexander and Sharon Goldwater. 2018.
\newblock Evaluating historical text normalization systems: How well do they
  generalize?
\newblock In \emph{Proceedings of the 2018 Conference of the North American
  Chapter of the Association for Computational Linguistics: Human Language
  Technologies, Volume 2 (Short Papers)}, pages 720--725.

\bibitem[{Rudnicka(2018)}]{rudnicka2018variation}
Rudnicka, Karolina. 2018.
\newblock Variation of sentence length across time and genre.
\newblock \emph{Diachronic corpora, genre, and language change}, pages
  220--240.

\bibitem[{Sainz et~al.(2023)Sainz, Campos, Garc{\'\i}a-Ferrero, Etxaniz,
  de~Lacalle, and Agirre}]{sainz-etal-2023-nlp}
Sainz, Oscar, Jon Campos, Iker Garc{\'\i}a-Ferrero, Julen Etxaniz, Oier~Lopez
  de~Lacalle, and Eneko Agirre. 2023.
\newblock {NLP} evaluation in trouble: On the need to measure {LLM} data
  contamination for each benchmark.
\newblock In \emph{Findings of the Association for Computational Linguistics:
  EMNLP 2023}, pages 10776--10787, Association for Computational Linguistics,
  Singapore.

\bibitem[{Saraswat and Srishti(2022)}]{saraswat2022leveraging}
Saraswat, Mala and Srishti. 2022.
\newblock Leveraging genre classification with rnn for book recommendation.
\newblock \emph{International Journal of Information Technology},
  14(7):3751--3756.

\bibitem[{Scialom et~al.(2020)Scialom, Dray, Lamprier, Piwowarski, and
  Staiano}]{scialom-etal-2020-mlsum}
Scialom, Thomas, Paul-Alexis Dray, Sylvain Lamprier, Benjamin Piwowarski, and
  Jacopo Staiano. 2020.
\newblock {MLSUM}: The multilingual summarization corpus.
\newblock In \emph{Proceedings of the 2020 Conference on Empirical Methods in
  Natural Language Processing (EMNLP)}, pages 8051--8067, Association for
  Computational Linguistics, Online.

\bibitem[{Seabold and Perktold(2010)}]{seabold2010statsmodels}
Seabold, Skipper and Josef Perktold. 2010.
\newblock Statsmodels: Econometric and statistical modeling with python.
\newblock In \emph{Proceedings of the Python in Science Conference}, page~57,
  SciPy.

\bibitem[{Shen et~al.(2023)Shen, Cheng, Nguyen, You, and Bing}]{shen2023large}
Shen, Chenhui, Liying Cheng, Xuan-Phi Nguyen, Yang You, and Lidong Bing. 2023.
\newblock Large language models are not yet human-level evaluators for
  abstractive summarization.
\newblock In \emph{Findings of the Association for Computational Linguistics:
  EMNLP 2023}, pages 4215--4233, Association for Computational Linguistics,
  Singapore.

\bibitem[{Sherman(1893)}]{sherman1893analytics}
Sherman, Lucius~Adelno. 1893.
\newblock \emph{Analytics of literature: A manual for the objective study of
  English prose and poetry}.
\newblock Ginn.

\bibitem[{Soni and Wade(2023)}]{soni2023comparing}
Soni, Mayank and Vincent Wade. 2023.
\newblock Comparing abstractive summaries generated by chatgpt to real
  summaries through blinded reviewers and text classification algorithms.
\newblock \emph{arXiv preprint arXiv:2303.17650}.

\bibitem[{Takase and Okazaki(2022)}]{takase-okazaki-2022-multi}
Takase, Sho and Naoaki Okazaki. 2022.
\newblock Multi-task learning for cross-lingual abstractive summarization.
\newblock In \emph{Proceedings of the Thirteenth Language Resources and
  Evaluation Conference}, pages 3008--3016, European Language Resources
  Association, Marseille, France.

\bibitem[{Tang et~al.(2020)Tang, Tran, Li, Chen, Goyal, Chaudhary, Gu, and
  Fan}]{tang2020multilingual}
Tang, Yuqing, Chau Tran, Xian Li, Peng-Jen Chen, Naman Goyal, Vishrav
  Chaudhary, Jiatao Gu, and Angela Fan. 2020.
\newblock Multilingual translation with extensible multilingual pretraining and
  finetuning.

\bibitem[{Thai et~al.(2022)Thai, Karpinska, Krishna, Ray, Inghilleri, Wieting,
  and Iyyer}]{thai-etal-2022-exploring}
Thai, Katherine, Marzena Karpinska, Kalpesh Krishna, Bill Ray, Moira
  Inghilleri, John Wieting, and Mohit Iyyer. 2022.
\newblock Exploring document-level literary machine translation with parallel
  paragraphs from world literature.
\newblock In \emph{Proceedings of the 2022 Conference on Empirical Methods in
  Natural Language Processing}, pages 9882--9902, Association for Computational
  Linguistics, Abu Dhabi, United Arab Emirates.

\bibitem[{Vyas, Niu, and Carpuat(2018)}]{vyas-etal-2018-identifying}
Vyas, Yogarshi, Xing Niu, and Marine Carpuat. 2018.
\newblock Identifying semantic divergences in parallel text without
  annotations.
\newblock In \emph{Proceedings of the 2018 Conference of the North {A}merican
  Chapter of the Association for Computational Linguistics: Human Language
  Technologies, Volume 1 (Long Papers)}, pages 1503--1515, Association for
  Computational Linguistics, New Orleans, Louisiana.

\bibitem[{Wan, Li, and Xiao(2010)}]{wan2010cross}
Wan, Xiaojun, Huiying Li, and Jianguo Xiao. 2010.
\newblock Cross-language document summarization based on machine translation
  quality prediction.
\newblock In \emph{Proceedings of the 48th Annual Meeting of the Association
  for Computational Linguistics}, pages 917--926.

\bibitem[{Wang, Cho, and Lewis(2020)}]{wang-etal-2020-asking}
Wang, Alex, Kyunghyun Cho, and Mike Lewis. 2020.
\newblock Asking and answering questions to evaluate the factual consistency of
  summaries.
\newblock In \emph{Proceedings of the 58th Annual Meeting of the Association
  for Computational Linguistics}, pages 5008--5020, Association for
  Computational Linguistics, Online.

\bibitem[{Wang et~al.(2023)Wang, Liang, Meng, Zou, Li, Qu, and
  Zhou}]{wang_cross-lingual_2023}
Wang, Jiaan, Yunlong Liang, Fandong Meng, Beiqi Zou, Zhixu Li, Jianfeng Qu, and
  Jie Zhou. 2023.
\newblock Zero-shot cross-lingual summarization via large language models.
\newblock In \emph{Proceedings of the 4th New Frontiers in Summarization
  Workshop}, pages 12--23, Association for Computational Linguistics,
  Singapore.

\bibitem[{Wang and Liu(2017)}]{wang2017effects}
Wang, Yaqin and Haitao Liu. 2017.
\newblock The effects of genre on dependency distance and dependency direction.
\newblock \emph{Language Sciences}, 59:135--147.

\bibitem[{Weller, Seppi, and Gardner(2022)}]{weller-etal-2022-use}
Weller, Orion, Kevin Seppi, and Matt Gardner. 2022.
\newblock When to use multi-task learning vs intermediate fine-tuning for
  pre-trained encoder transfer learning.
\newblock In \emph{Proceedings of the 60th Annual Meeting of the Association
  for Computational Linguistics (Volume 2: Short Papers)}, pages 272--282,
  Association for Computational Linguistics, Dublin, Ireland.

\bibitem[{Yang et~al.(2023)Yang, Li, Zhang, Chen, and
  Cheng}]{yang_exploring_2023}
Yang, Xianjun, Yan Li, Xinlu Zhang, Haifeng Chen, and Wei Cheng. 2023.
\newblock Exploring the limits of chatgpt for query or aspect-based text
  summarization.
\newblock \emph{arXiv preprint arXiv:2302.08081}.

\bibitem[{Yao, Wan, and Xiao(2015)}]{yao2015phrase}
Yao, Jin-ge, Xiaojun Wan, and Jianguo Xiao. 2015.
\newblock Phrase-based compressive cross-language summarization.
\newblock In \emph{Proceedings of the 2015 conference on empirical methods in
  natural language processing}, pages 118--127.

\bibitem[{Yuan, Neubig, and Liu(2021)}]{NEURIPS2021_e4d2b6e6}
Yuan, Weizhe, Graham Neubig, and Pengfei Liu. 2021.
\newblock Bartscore: Evaluating generated text as text generation.
\newblock In \emph{Advances in Neural Information Processing Systems},
  volume~34, pages 27263--27277, Curran Associates, Inc.

\bibitem[{Zaheer et~al.(2020)Zaheer, Guruganesh, Dubey, Ainslie, Alberti,
  Ontanon, Pham, Ravula, Wang, Yang et~al.}]{zaheer2020bigbird}
Zaheer, Manzil, Guru Guruganesh, Kumar~Avinava Dubey, Joshua Ainslie, Chris
  Alberti, Santiago Ontanon, Philip Pham, Anirudh Ravula, Qifan Wang, Li~Yang,
  et~al. 2020.
\newblock Big bird: Transformers for longer sequences.
\newblock \emph{Advances in neural information processing systems},
  33:17283--17297.

\bibitem[{Zhang, Liu, and Zhang(2023)}]{zhang2023extractivechatgpt}
Zhang, Haopeng, Xiao Liu, and Jiawei Zhang. 2023.
\newblock Extractive summarization via {C}hat{GPT} for faithful summary
  generation.
\newblock In \emph{Findings of the Association for Computational Linguistics:
  EMNLP 2023}, pages 3270--3278, Association for Computational Linguistics,
  Singapore.

\bibitem[{Zhang et~al.(2020{\natexlab{a}})Zhang, Zhao, Saleh, and
  Liu}]{zhang2020pegasus}
Zhang, Jingqing, Yao Zhao, Mohammad Saleh, and Peter Liu. 2020{\natexlab{a}}.
\newblock Pegasus: Pre-training with extracted gap-sentences for abstractive
  summarization.
\newblock In \emph{International Conference on Machine Learning}, pages
  11328--11339, PMLR.

\bibitem[{Zhang et~al.(2020{\natexlab{b}})Zhang, Kishore, Wu, Weinberger, and
  Artzi}]{bert-score}
Zhang, Tianyi, Varsha Kishore, Felix Wu, Kilian~Q. Weinberger, and Yoav Artzi.
  2020{\natexlab{b}}.
\newblock Bertscore: Evaluating text generation with bert.
\newblock In \emph{International Conference on Learning Representations}.

\bibitem[{Zhao et~al.(2019)Zhao, Peyrard, Liu, Gao, Meyer, and
  Eger}]{zhao2019moverscore}
Zhao, Wei, Maxime Peyrard, Fei Liu, Yang Gao, Christian~M Meyer, and Steffen
  Eger. 2019.
\newblock Moverscore: Text generation evaluating with contextualized embeddings
  and earth mover distance.
\newblock In \emph{Proceedings of the 2019 Conference on Empirical Methods in
  Natural Language Processing and the 9th International Joint Conference on
  Natural Language Processing (EMNLP-IJCNLP)}, pages 563--578.

\bibitem[{Zhao, Strube, and Eger(2023)}]{zhao-etal-2023-discoscore}
Zhao, Wei, Michael Strube, and Steffen Eger. 2023.
\newblock {D}isco{S}core: Evaluating text generation with {BERT} and discourse
  coherence.
\newblock In \emph{Proceedings of the 17th Conference of the European Chapter
  of the Association for Computational Linguistics}, pages 3865--3883,
  Association for Computational Linguistics, Dubrovnik, Croatia.

\bibitem[{Zhu, Liu, and Pang(2022)}]{zhu2022investigating}
Zhu, Haoran, Xueying Liu, and Nana Pang. 2022.
\newblock Investigating diachronic change in dependency distance of modern
  english: A genre-specific perspective.
\newblock \emph{Lingua}, 272:103307.

\bibitem[{Zhu et~al.(2019)Zhu, Wang, Wang, Zhou, Zhang, Wang, and
  Zong}]{zhu-etal-2019-ncls}
Zhu, Junnan, Qian Wang, Yining Wang, Yu~Zhou, Jiajun Zhang, Shaonan Wang, and
  Chengqing Zong. 2019.
\newblock {NCLS}: Neural cross-lingual summarization.
\newblock In \emph{Proceedings of the 2019 Conference on Empirical Methods in
  Natural Language Processing and the 9th International Joint Conference on
  Natural Language Processing (EMNLP-IJCNLP)}, pages 3054--3064, Association
  for Computational Linguistics, Hong Kong, China.

\end{thebibliography}

\clearpage
\section{Appendix}
\label{appendix}
\subsection{Links to the sources }
\label{apdx:links}
\begin{itemize}
    \item DTA: \url{https://www.deutschestextarchiv.de/}
    \label{DTA}
    \item Wikisource: \url{https://www.wikisource.org/}
    \label{wikisource}
    \item Examples of proofreading: \url{https://de.wikisource.org/w/index.php?title=Seite:Kinder_und_Hausm\%C3\%A4rchen_(Grimm)_1812_I_008.jpg}
    \label{proof}
    \item Project Gutenberg: \url{https://www.gutenberg.org/}
    \item Beautifulsoup library: \url{https://pypi.org/project/beautifulsoup4/}.
    \item list of German fairytale: \url{https://de.wikipedia.org/wiki/Liste\_von\_MÃďrchen#Deutsche\_MÃďrchen}
    \item list of English short stories: \url{https://en.wikipedia.org/wiki/Category:Short_stories} 
    \item WikiLingua Repository: 
\url{https://github.com/esdurmus/Wikilingua}
    \item CNN/Daily Mail Dataset: \url{https://huggingface.co/datasets/cnn_dailymail}
\end{itemize}

\subsection{Example of translation dataset}
Table \ref{appdx:Tran_example} showcases an example of translation dataset. 
\begin{table}[!htb]
\begin{tabularx}{\textwidth}{X}
\toprule
\textbf{German Text: Läuschen und Flöhchen}\\
``Ein Läuschen und ein Flöhchen die lebten zusammen in einem \textit{Haushalte} und brauten das Bier in einer Eierschale. Da fiel das Läuschen hinein und verbrannte sich. Darüber \emph{fieng} das Flöhchen an laut zu schreien. Da sprach die kleine \textit{Stubenthüre} ‘was schreist du, Flöhchen?’ ‘Weil Läuschen sich verbrannt hat.’
Da \emph{fieng} das \textit{Thürchen} an zu knarren. Da sprach ein Besenchen in der Ecke ‘was knarrst du, \textit{Thürchen}?’ ‘Soll ich nicht knarren?’'' [...] \\
\textbf{English Text: The Louse and the Flea}\\
``A louse and a flea kept house together and were brewing beer in an egg-shell. Then the little louse fell in and burnt herself. On this the little flea began to scream loudly. \textit{Then said} the little room-door, "Little flea, why \textit{art} \emph{thou} screaming?" "Because the louse has burnt herself."
Then the little door began to creak. On this a little broom in the corner said, "Why \textit{art} \emph{thou} creaking, little door?" "\emph{Have I not} reason to creak?" ''[...]\\
\bottomrule
\end{tabularx}
\caption{Example of translation dataset. We highlight words with spelling/morphological changes in italics.}
\label{appdx:Tran_example}
\end{table}

\subsection{Configuration of evaluation metrics}
\begin{table}[H]
\begin{tabular}{p{1.5cm}p{1cm}p{3cm}}
\toprule
\textbf{Metrics}  & \textbf{Eval lang} &  \textbf{configurations}  \\
\midrule
\multirow{2}{*}{BERTScore} & English & microsoft/deberta-xlarge-mnli\\
& German &  xlm-roberta-large\\
\multirow{2}{*}{MoverScore} & English & distilbert-base-uncased \\
&German & xlm-roberta-large\\
\multirow{2}{*}{BARTScore} & English & facebook/bart-large-cnn \\
&German & facebook/mbart-large-50-many-to-many-mmt \\
\multirow{2}{*}{MENLI} & English & NLI-D default, microsoft/deberta-large-mnli \\
&German & NLI-D cross-lingual, MoritzLaurer/mDeBERTa-v3-base-mnli-xnli\\
\multirow{2}{*}{DiscoScore} & English & Conpono \\
&German & bert-base-multilingual-cased\\
\bottomrule
\end{tabular}
\caption{\label{apdx:config}
Configuration of evaluation metrics
}
\end{table}

\subsection{Results for Baseline CLCTS trained with the expanded datasets.}
\label{apdx:extended_data}
Table \ref{apdx:expansion} shows the results for baseline CLCTS hEn-De and hDe-En using the expanded datasets.

\begin{table*}[!htb]
\fontsize{8pt}{8pt}\selectfont
\begin{tabular}{llll}
\toprule
\textbf{Model} & \textbf{ROUGE-1}  & \textbf{ROUGE-L}  &   \textbf{BERTScore-F1} \\
\midrule
\multicolumn{4}{c}{\centering \textit{Supervised Abstractive - Baseline CLCTS}}\\
\multicolumn{4}{l}{\textbf{mLED}}\\
\text{-Base} & 0.389 (-0.002) & 0.206 (0.004) & 0.552 (0.005)\\
\text{-Historical MT}	& 0.397 (0.003) & 0.213 (0.013)	 & 0.561 (0.009)
\\
\text{-$\text{Bidirection}_{\text{Prefix}}$}&  0.398 (0.010) & 0.211 (0.013)	&0.558 (0.011)	 \\ 
Avg. change & 0.004	& 0.010 & 0.008 \\
\bottomrule
\\
\multicolumn{4}{c}{\centering \textbf{(a) Direction hDe-En}}\\
\\
\toprule
\textbf{Model} & \textbf{ROUGE-1}  & \textbf{ROUGE-L}  &  \textbf{BERTScore-F1} \\
\midrule
\multicolumn{4}{c}{\centering \textit{Supervised Abstractive - Baseline CLCTS}}\\
\multicolumn{4}{l}{\textbf{mLED}}\\
\text{-Base} & 0.315 (-0.007)	&0.141 (-0.001)&	0.847 (-0.001)\\
\text{-Historical MT} &0.329 (0.001)	&0.145 (0.002)	&0.849 (0.001)\\
\text{-$\text{Bidirection}_{\text{Prefix}}$}& 0.323 (-0.005) & 0.144 (-0.001) &	0.848 (-0.001) \\ 
Avg. change & -0.004 & 0.000 & 0.000
\\
\bottomrule
\\
\multicolumn{4}{c}{\centering \textbf{(b) Direction hEn-De}}\\
\end{tabular}
\caption{\label{apdx:expansion}
Results for Baseline CLCTS hEn-De and hDe-En using the expanded datasets (hDe-En from 328 to 455 instances and hEn-De from 289 to 501 instances). The corresponding changes compared to the results in the main text (i.e., new results minus the old results) are reported in brackets.}
\end{table*}

\subsection{Annotation details}
\label{appdx:annotation}
We utilize the same evaluation criteria as SummEval \cite{fabbri2021summeval}. Annotators are required to read the instructions listed below carefully. Then, they are provided with a source document, a corresponding reference summary, and a generated summary for evaluation. All annotators are blind to the model information (i.e., annotators do not know which output comes from which model). The three authors of this paper (out of six annotators) participated in the annotation as well and they have as much information as the other three annotators. None of the annotators have apriori preferences for any of the models examined. The selection of model outputs for each source document is random and we shuffle the documents before they are presented to the annotators. We ask the annotators to rate the summaries on a Likert scale from 1 to 5 (higher better, including 0.5 increments) along the four dimensions, namely, coherence, consistency, fluency, and relevance.
\\ \\
The following guidelines are presented to the annotators before annotation: \newline

\noindent\textbf{Information \& Instructions}\\
\noindent In this task, you will evaluate automatically generated summaries of historical short stories. The source document and a reference summary from Wikipedia are provided. 
\begin{itemize}
    \item You should evaluate the texts based on their coherence, consistency, fluency, and relevance.
    \item Rate each item with a score from 1 (worst) to 5 (best). 
    \item In case of uncertainty, the grades can be increased by 0.5 steps (e.g., 3.5).
    \item A gold standard was defined based on the test phase of the evaluation. Please consider the gold standard, see the corresponding tab.
\end{itemize}
\noindent\textbf{Definitions (SummEval)}
\begin{itemize}
    \item \textbf{Coherence}: The rating measures the quality of all sentences collectively. The sentences must fit together and sound natural. Consider the quality of the summary as a whole.
    \item \textbf{Consistency}: The rating measures whether the facts in the summary are consistent with the facts in the reference summary. Consider whether the summary reproduces all facts accurately and does not make up untrue information. 
    \item \textbf{Fluency}: This rating measures the quality of individual sentences, whether they are well-written and grammatically correct. Consider the quality of individual sentences.
    \item \textbf{Relevance}: The rating measures how well the summary captures the key points of the article. Consider whether all and only the important aspects are contained in the summary. 	
\end{itemize}

\textbf{Number of annotated instances by models.}
Table \ref{tab:anno_number} demonstrates the number of annotated instances for each model. Each instance is annotated by one or several annotators.
\begin{table}[!hbt]
\fontsize{8pt}{8pt}\selectfont
\begin{tabular}{lll}
\toprule
\textbf{Model} & \textbf{hDe-En}  &  \textbf{hEn-De}  \\
\midrule
\multicolumn{3}{c}{\centering \textit{Supervised Extractive}}\\
\multicolumn{3}{l}{\centering \textbf{MemSum}}\\
-$\text{translation}_{\text{max 25}}$ & 10 & 8 \\
-$\text{Norma-translation}_{\text{max 25}}$	&10 &  8\\
\multicolumn{3}{c}{\centering \textit{Supervised Abstractive - Baseline CLCTS}}\\
\multicolumn{3}{l}{\textbf{mLED}}\\
\text{-Base} &	10 & 8	\\
\text{-Historical MT}		& 10 &8 \\
\text{-$\text{Bidirection}_{\text{Prefix}}$} &	10 & 8 \\
\multicolumn{3}{c}{\centering \textit{Supervised Abstractive - intermediate finetuning}}\\
\multicolumn{3}{l}{\textbf{mLED}}\\
\text{$\text{-MLS}_{\text{tgt}}$} 	& 10 & 8 \\
\text{$\text{-MLS}_{\text{src+tgt}}$} 	& 10 &	8 \\
\text{-MLS-CLS}	&	10 &  8 \\
\text{-MLS-CLS-CTS}	&  10 & 8 \\
\multicolumn{3}{c}{\centering \textit{Zero-shot Abstractive - ChatGPT}}\\
\multicolumn{3}{l}{\textbf{ChatGPT as summarizer}}\\
-e2e (Title) prompt 	& 10 & 8 \\
-e2e prompt & 10 & 8 \\
-pipeline prompt 	& 10 & 8 \\ 
-$\text{retrieve-ChatGPT}_{\text{max 100}}$	& 10  & 8 \\
\midrule
Total & 130 & 104 \\
\bottomrule
\end{tabular}
\caption{Number of annotated instances per model. Each instance is annotated by one or several annotators.}
\label{tab:anno_number}
\end{table}

\subsection{ChatGPT evaluation: annotation results from GPT-3.5-turbo}
\label{apdx:chatgpt_eval}
Table \ref{dis:annotation} reports the annotation results from GPT-3.5-turbo. 
\begin{table*}[!ht]
\fontsize{8pt}{8pt}\selectfont
\begin{tabular}{lllll}
\toprule
\textbf{Model} & \textbf{Coherence}  &  \textbf{Consistency}  & \textbf{Fluency} & \textbf{Relevance } \\
\hline
\multicolumn{5}{c}{\centering \textit{Supervised Extractive}}\\
\multicolumn{5}{l}{\centering \textbf{MemSum}}\\
-$\text{translation}_{\text{max 25}}$  &	2.52/2.00&2.92/2.10&2.98/1.80&3.32/1.60\\
-$\text{Norma-translation}_{\text{max 25}}$	&2.50/1.90&2.92/1.60&2.95/1.90&3.20/1.20 \\
\multicolumn{5}{c}{\centering \textit{Supervised Abstractive - Baseline CLCTS}}\\
\multicolumn{5}{l}{\textbf{mLED}}\\
\text{-Base}	&	2.10/1.88&2.08/1.88&2.78/2.00&2.22/1.62\\
\text{-Historical MT}	&	\underline{2.86}/1.89&\underline{2.63}/\underline{2.33}&3.01/2.00&3.19/\underline{2.00}\\
\text{-$\text{Bidirection}_{\text{Prefix}}$}	&	2.43/1.70&2.28/1.40&2.54/1.90&2.58/1.20\\
\multicolumn{5}{c}{\centering \textit{Supervised Abstractive - intermediate finetuning}}\\
\multicolumn{5}{l}{\textbf{mLED}}\\
\text{$\text{-MLS}_{\text{tgt}}$} 	& 2.75/\underline{2.00}&2.56/1.89&\underline{3.08}/\underline{2.22}&\underline{3.39}/1.67\\
\text{$\text{-MLS}_{\text{src+tgt}}$} 	&	2.38/1.50&2.30/1.30&2.78/1.50&2.75/1.10\\
\text{-MLS-CLS}	&	2.20/1.40&1.90/1.30&2.48/1.40&2.42/1.10\\
\text{-MLS-CLS-CTS}	& 2.11/1.60&1.92/1.30&2.30/1.60&2.28/1.20\\
\multicolumn{5}{c}{\centering \textit{Zero-shot Abstractive - ChatGPT}}\\
\multicolumn{5}{l}{\textbf{ChatGPT as summarizer}}\\
-e2e (Title) prompt&2.55/3.00&3.05/2.20&4.10/3.70&2.02/2.00\\
-e2e prompt & 4.14/\textbf{4.00}&3.98/\textbf{4.60}&4.18/\textbf{4.10}&3.97/\textbf{4.50}\\
-pipeline prompt &	\textbf{4.35}/\textbf{4.00}&\textbf{4.30}/4.30&\textbf{4.55}/4.00&\textbf{4.35}/4.20\\
-$\text{retrieve-ChatGPT}_{\text{max 100}}$	&4.08/3.60&4.08/3.60&4.15/3.70&4.25/3.90\\
\bottomrule
\\
\multicolumn{5}{c}{\centering \textbf{(a) Direction hDe-En}}\\
\\
\toprule
\textbf{Model} & \textbf{Coherence}  &  \textbf{Consistency}  & \textbf{Fluency} & \textbf{Relevance } \\
\hline
\multicolumn{5}{c}{\centering \textit{Supervised Extractive}}\\
\multicolumn{5}{l}{\centering \textbf{MemSum}}\\
-$\text{translation}_{\text{max 25}}$	&2.47/1.50&2.34/1.38&2.62/1.50&2.19/1.00\\
-$\text{Norma-translation}_{\text{max 25}}$&2.53/1.62&2.12/1.25&2.75/1.62&2.22/1.00\\
\multicolumn{5}{c}{\centering \textit{Supervised Abstractive - Baseline CLCTS}}\\
\multicolumn{5}{l}{\textbf{mLED}}\\
\text{-Base}	&	2.01/1.25&1.56/1.00&2.32/1.50&1.65/1.00\\
\text{-Historical MT} &	1.98/2.00&1.78/\underline{1.69}&2.74/\underline{1.94}&1.78/\underline{1.38}\\
\text{-$\text{Bidirection}_{\text{Prefix}}$}	&	\underline{2.75}/\underline{1.69}&\underline{2.03}/1.25&\underline{2.78}/1.62&1.97/1.12\\
\multicolumn{5}{c}{\centering \textit{Supervised Abstractive - intermediate finetuning}}\\
\multicolumn{5}{l}{\textbf{mLED}}\\
\text{$\text{-MLS}_{\text{tgt}}$} 	&	1.99/1.50&1.72/1.38&2.43/1.50&1.91/1.12\\
\text{$\text{-MLS}_{\text{src+tgt}}$} 	&	1.83/1.38&1.70/1.06&2.33/1.38&1.66/1.00\\
\text{-MLS-CLS}		&	1.80/1.29&1.71/1.29&2.39/1.57&1.98/1.00\\
\text{-MLS-CLS-CTS}	&	2.41/1.12&1.88/1.00&2.34/1.12&\underline{2.00}/1.00\\
\multicolumn{5}{c}{\centering \textit{Zero-shot Abstractive - ChatGPT}}\\
\multicolumn{5}{l}{\textbf{ChatGPT as summarizer}}\\
-e2e (Title) prompt &	3.31/2.19&2.25/2.25&3.28/3.19&2.31/1.88\\
-e2e prompt& 3.31/2.78&2.80/2.61&3.38/3.11&2.98/2.33\\
-pipeline prompt&	3.34/\textbf{3.25}&\textbf{3.19}/3.12&\textbf{3.47}/\textbf{3.50}&\textbf{3.31}/\textbf{3.12}\\
-$\text{retrieve-ChatGPT}_{\text{max 100}}$	&\textbf{3.38}/\textbf{3.25}&2.94/\textbf{3.38}&3.34/3.25&3.06/3.00\\
\bottomrule
\\
\multicolumn{5}{c}{\centering \textbf{(b) Direction hEn-De}}\\
\\
\toprule
\textbf{Models} & \textbf{Coh.} & \textbf{Con.} & \textbf{Flu.} & \textbf{Rel.} \\
\hline
hDe-En & 0.512 & 0.531 & 0.569 & 0.549   \\
hEn-De & 0.417 & 0.454 & 0.424 & 0.568 \\ 
\bottomrule
\\
\multicolumn{5}{c}{\centering \textbf{(c) Annotation agreement for human and ChatGPT}}\\
\end{tabular}
\caption{\label{dis:annotation}
Average human and ChatGPT (GPT-3.5-turbo) ratings for the CLCTS dataset (in Table (a) and (b), the scores from human and ChatGPT are separated by a slash (/), i.e., human-annotation/ChatGPT-annotation). Table (c) is the document level annotation agreement between humans and ChatGPT. The best result of all models is in bold font. The best score for the supervised abstractive model is indicated with an underline. }
\end{table*}
\\ \\
\noindent Table \ref{tab:agree_chatty} reports the annotation agreement between GPT-3.5-turbo and GPT-4-1106-preview.
\begin{table*}[!htb]
\fontsize{8pt}{8pt}\selectfont
\begin{tabular}{lllll}
\toprule
\textbf{Models} & \textbf{Coh.} & \textbf{Con.} & \textbf{Flu.} & \textbf{Rel.} \\
\midrule
hDe-En & 0.664 & 0.631 & 0.751 & 0.738 \\
hEn-De & 0.729 & 0.661 & 0.770 & 0.748 \\ 
\bottomrule
\end{tabular}
\caption{\label{tab:agree_chatty} Document level annotation agreement between GPT-3.5-turbo and GPT-4-1106-preview.}
\end{table*} 

\subsection{Comparison of the generated summaries and references}\label{apdx:sample}
In Section \ref{sec:exmaples}, we present references and generated samples showcasing the main errors of the generated outputs. Here, we provide more details comparing generated outputs to references according to our four evaluation criteria, namely coherence, consistency, fluency, and relevance. 

Machine-generated summaries are different from human-written summaries in that the generated summaries: 
(1) lack fluency (except for GPT4), containing grammar errors, punctuation errors, and untranslated texts as shown in Table \ref{result:example}. The automatic summaries sometimes contain nonsensical sentences such as \textit{``Once again, hedge and hedge become hedges again.''} (from mLED-Base with fluency of 1); (2) contain factual inconsistencies, especially in supervised abstractive models. For example, a generated summary from mLED-Base (supervised abstractive model) with relevance of 1 starts with \textit{``A rich man has two daughters. The first is beautiful, the second is beautiful and the third is beautiful.''} which greatly contradicts the source and reference. Indeed, the human-written reference mentions ``only daughter'' in the very first sentence. Moreover, the generated sentence by itself is also inconsistent concerning the number of daughters; (3) lack relevance. Sometimes, the generated summaries omit important plots provided in human-written references and the summaries may become overly concise in this case. For example, ChatGPT summarizes that \textit{``Her father remarries and her stepmother and stepsisters treat her cruelly. With the help of some birds, cinderella is able to attend a ball and dance with the prince.''} while the human-written reference contains the description of the event: \textit{``The king decides to proclaim a festival that will last for three days and invites all the beautiful maidens in that country to attend so that the prince can select one of them for his bride.''} Without this information, the understandability of the story drops and such omissions lead to points reduction in relevance and coherence; (4) lack coherence, especially for abstractive models. There are also cases where the texts are incomprehensible. For example, here is one generation with coherence of 1: \textit{``[...] He asks him where he was when he was at the castle [...] The third asks what he learned, and he explains the reason he did so [...] He goes to church and admonishes the giants and flies. He follows them until the world becomes filled with birds, and then the world is filled with flies and spider-smugglingbirds.''} (5) have different distributions of lengths compared to the references depending on the models. Outputs from abstractive models usually contain fewer tokens than human-written references, on average 27 tokens less for supervised abstractive models and more than 100 tokens less for zero-shot abstractive models. This also explains why zero-shot models omit plots. In contrast, the generated outputs from extractive models contain 300 more tokens on average and usually contain more information than the reference, including more details or even redundant contents such as full conversations and repeated information.

\subsection{Details on regression analysis}
\label{appdx:reg}    
\subsubsection{Variance inflation factors (VIF)}
Table \ref{apdx:vif_table} contains the variance inflation factor for numerical features.
\begin{table}[!hbt]
\fontsize{9pt}{9pt}\selectfont
\begin{tabular}{lll}
\toprule
\textbf{Feature}  & \textbf{hDe-En} &  \textbf{hEn-De}  \\
\hline
Publication year &	1.04 &2.45 \\
Length of document &	1.27	& 2.60 \\
Mean dependency distance & 1.02 & 2.19 \\
Embedding similarity &	1.30 &	1.55 \\
\bottomrule
\end{tabular}
\caption{\label{apdx:vif_table}
Variance inflation factor for numerical features. The VIF for the publication year is computed based on the numerical value of the year. 
}
\end{table}

\subsubsection{Regression coefficient for variable model}
In Table \ref{tab:reg} we list the regression coefficient for the variable \textit{model}.
\begin{table}[!hbt]
\fontsize{8pt}{8pt}\selectfont
\begin{tabular}{lll}
\toprule
\textbf{Model} & \textbf{hDe-En}  &  \textbf{hEn-De}  \\
\midrule
\multicolumn{3}{c}{\centering \textit{Supervised Extractive}}\\
\multicolumn{3}{l}{\centering \textbf{MemSum}}\\
-$\text{translation}_{\text{max 25}}$ & -0.06 & -0.94$^{***}$ \\
-$\text{Norma-translation}_{\text{max 25}}$	&-0.19$^{**}$ &  -0.98$^{***}$\\
\multicolumn{3}{c}{\centering \textit{Supervised Abstractive - Baseline CLCTS}}\\
\multicolumn{3}{l}{\textbf{mLED}}\\
\text{-Historical MT}		&	-0.0 & 0.08 \\
\text{-$\text{Bidirection}_{\text{Prefix}}$} &	0.08 & 0.08 \\
\multicolumn{3}{c}{\centering \textit{Supervised Abstractive - intermediate finetuning}}\\
\multicolumn{3}{l}{\textbf{mLED}}\\
\text{$\text{-MLS}_{\text{tgt}}$} 	& -0.01 & 0.02 \\
\text{$\text{-MLS}_{\text{src+tgt}}$} 	& -0.13$^{*}$ &	-0.07 \\
\text{-MLS-CLS}	&	-0.22$^{***}$ &  0.09 \\
\text{-MLS-CLS-CTS}	&  -0.16$^{**}$& 0.08 \\
\multicolumn{3}{c}{\centering \textit{Zero-shot Abstractive - ChatGPT}}\\
\multicolumn{3}{l}{\textbf{ChatGPT as summarizer}}\\
-e2e (Title) prompt 	& -0.37$^{***}$ & 0.06 \\
-e2e prompt & 1.47$^{***}$ & 0.64$^{***}$\\
-pipeline prompt	& 1.30$^{***}$ & 0.59$^{***}$\\ 
-$\text{retrieve-ChatGPT}_{\text{max 100}}$	& 1.30$^{***}$ & 0.55$^{***}$ \\
\bottomrule
\end{tabular}
\caption{Regression results for model features.}
\label{tab:reg}
\end{table}

\subsection{Examples of adversarial attacks}
\label{apdx:example_adversarial}
We list here one example from a short story with all three attacks. All successful cases are heightened in \textcolor{green}{green} and the failures in \textcolor{red}{red}. We also highlight the swapped entities in \textcolor{orange}{orange} in the adversarial text with the corresponding entities in the source text in \textit{italics}. 

\subsubsection{Source Text}
``A little brother and sister were once playing by a \textit{well}, and while they were thus playing, they both fell in. A water-nix lived down below, who said, "Now I have got you, now you shall work hard for me!" and carried them off with her. She gave the girl dirty tangled flax to spin, and she had to fetch \textit{water} in a bucket with a hole in it, and the boy had to hew down a tree with a blunt axe, and they got nothing to eat but dumplings as hard as stones. Then at last the children became so impatient, that they waited until one Sunday, when the nix was at church, and ran away. But when church was over, the nix saw that the birds were flown, and followed them with great strides. The children saw her from afar, and the girl threw a \textit{brush} behind her which formed an immense \textit{hill of bristles}, with thousands and thousands of spikes, over which the nix was forced to scramble with great difficulty; at last, however, she got over. When the children saw this, the boy threw behind him a \textit{comb} which made a great \textit{hill of combs} with a thousand times a thousand teeth, but the nix managed to keep herself steady on them, and at last crossed over that. Then the girl threw behind her a looking-glass which formed a hill of mirrors, and was so slippery that it was impossible for the nix to cross it. Then she thought, "I will go home quickly and fetch my \textit{axe}, and cut the hill of glass in half." Long before she returned, however, and had hewn through the glass, the children had escaped to a great distance, and the water-nix was obliged to betake herself to her well again.''

\subsubsection{Entity Swap}
``A little brother and sister were once playing by a \textcolor{orange}{cave}, and while they were thus playing, they both fell in. A water-nix lived down below, who said, "Now I have got you, now you shall work hard for me!" and carried them off with her. She gave the girl dirty tangled flax to spin, and she had to fetch \textcolor{orange}{wine} in a bucket with a hole in it, and the boy had to hew down a tree with a blunt axe, and they got nothing to eat but dumplings as hard as stones. Then at last the children became so impatient, that they waited until one Sunday, when the nix was at church, and ran away. But when church was over, the nix saw that the birds were flown, and followed them with great strides. The children saw her from afar, and the girl threw a \textcolor{orange}{durian} behind her which formed an immense \textcolor{orange}{hill of durians}, with thousands and thousands of spikes, over which the nix was forced to scramble with great difficulty; at last, however, she got over. When the children saw this, the boy threw behind him a \textcolor{orange}{pineapple} which made a great \textcolor{orange}{hill of pineapples} with a thousand times a thousand spikes, but the nix managed to keep herself steady on them, and at last crossed over that. Then the girl threw behind her a looking-glass which formed a hill of mirrors, and was so slippery that it was impossible for the nix to cross it. Then she thought, "I will go home quickly and fetch my \textcolor{orange}{butcher knife}, and cut the hill of glass in half." Long before she returned, however, and had hewn through the glass, the children had escaped to a great distance, and the water-nix was obliged to betake herself to her well again.''
\newline

\noindent\textbf{Successful case from ChatGPT:}
``A brother and sister were playing near a \textcolor{green}{cave} when they fell in. They were captured by a water nymph who made them work hard and gave them only hard dumplings to eat. The children waited until the nymph went to church on a Sunday and then escaped. The nymph chased them, but the children threw \textcolor{green}{prickly fruits} and a mirror at her, causing her to struggle to climb over them. The children managed to escape while the nymph went to get \textcolor{green}{a tool} to break the mirror.''
\newline

\noindent\textbf{Failure case from ChatGPT:}
``A brother and sister were playing near a \textcolor{green}{cave} and fell in. They were captured by a water nymph who made them work hard, giving the girl bad flax to spin and making the boy chop a tree with a blunt axe, while only feeding them hard dumplings. One Sunday, they managed to escape when the nymph went to church. The nymph chased them, but the children threw different \textcolor{green}{fruit with spiky exteriors} behind them, making it hard for the nymph to catch up. Finally, the girl threw a smooth mirror, forcing the nymph to go home for an \textcolor{red}{axe} to break through it. By the time she returned, the children had already escaped.''

\subsubsection{Negation}
``\textbf{A water-nix} was once playing by a well, and while it was thus playing, it fell in. \textbf{A little brother and sister} lived down below, who said, "Now we have got you, now you shall work hard for \textbf{us}!" and carried it off with \textbf{them}. \textbf{They gave the nix} dirty tangled flax to spin, and \textbf{the water-nix had to} fetch water in a bucket with a hole in it, and \textbf{the water-nix had to also} hew down a tree with a blunt axe, and \textbf{the water-nix got nothing to eat} but dumplings as hard as stones. Then at last it became so impatient, that it waited until one Sunday, when \textbf{the brother and sister were at church}, and ran away. But when church was over, \textbf{they} saw that the birds were flown, and followed them with great strides. The \textbf{water-nix saw them} from afar, and threw a brush behind her which formed an immense hill of bristles, with thousands and thousands of spikes, over which they were forced to scramble with great difficulty; at last, however, \textbf{they got over}. When \textbf{the nix saw} this, \textbf{the nix threw} behind it a comb which made a great hill of combs with a thousand times a thousand teeth, but they managed to keep themselves steady on the combs, and at last crossed over that. Then \textbf{the nix threw} behind her a looking-glass which formed a hill of mirrors, and was so slippery that it was impossible for them to cross it. Then they thought, "\textbf{We will go} home quickly and fetch our axes, and cut the hill of glass in half." Long before they returned, however, and had hewn through the glass, \textbf{the nix had escaped to a great distance, and the brother and sister were obliged to betake themsleves to their well again}.''
\newline

\noindent\textbf{Successful case from ChatGPT:}
``\textcolor{green}{A water-nix fell into a well and was captured by a brother and sister who made it work hard for them}. \textcolor{green}{The water-nix had to} spin flax, fetch water with a leaking bucket, chop down a tree with a blunt axe, and eat hard dumplings. \textcolor{green}{The water-nix escaped when the siblings went to church}, but they chased after it. \textcolor{green}{The water-nix threw} obstacles in their way, including a hill of bristles, a hill of combs, and a hill of mirrors. \textcolor{green}{The siblings managed to cross} the first two hills but were unable to cross the hill of mirrors. They decided to go home and get their axes to cut the hill in half, but by the time they returned, \textcolor{green}{the water-nix had escaped}.''
\newline

\noindent\textbf{Failure case from ChatGPT:}
``\textcolor{green}{A water-nix falls into a well and is captured by a brother and sister} who force it to spin dirty flax, fetch water in a leaky bucket, and do other difficult tasks. \textcolor{green}{The water-nix escapes} on a Sunday when \textcolor{green}{the siblings are at church}, but they follow and the water-nix hinders them with a hill of bristles, a hill of combs, and a hill of mirrors. They plan to cut the glass hill in half but \textcolor{green}{the nix escapes before they can return with their axes}. \textcolor{red}{The story ends with the water-nix back in the well and the brother and sister empty-handed.}''

\subsubsection{Omission}   
``A little brother and sister were once playing by a well, and while they were thus playing, they both fell in. A water-nix lived down below, who said, "Now I have got you, now you shall work hard for me!" and carried them off with her. She gave the girl dirty tangled flax to spin, and she had to fetch water in a bucket with a hole in it, and the boy had to hew down a tree with a blunt axe, and they got nothing to eat but dumplings as hard as stones. Then at last the children became so impatient, that they waited until one Sunday, when the nix was at church, and ran away. But when church was over, the nix saw that the birds were flown, and followed them with great strides. The children saw her from afar, and the girl threw a brush behind her which formed an immense hill of bristles, with thousands and thousands of spikes, over which the nix was forced to scramble with great difficulty; at last, however, she got over. When the children saw this, the boy threw behind him a comb which made a great hill of combs with a thousand times a thousand teeth, but the nix managed to keep herself steady on them, and at last crossed over that. Then the girl threw behind her a looking-glass which formed a hill of mirrors, and was so slippery that it was impossible for the nix to cross it. Then she thought, "I will go home quickly and fetch my axe, and cut the hill of glass in half.''
\newline

\noindent\textbf{Successful case from ChatGPT:}
``A brother and sister fall into a well and are taken by a water-nix. They are forced to work hard and eat only hard dumplings. One Sunday when the nix is at church, the children run away. The nix chases them, but the children throw objects that create obstacles for her to cross. \textcolor{green}{The nix eventually crosses them all except for a hill of mirrors. She decides to go home and fetch her axe to cut the hill in half.}''
\newline

\noindent\textbf{Failure case from ChatGPT:}
``A brother and sister were playing by a well when they both fell in and were captured by a water nymph. The nymph made the girl spin flax and carry water while the boy had to chop a tree with a blunt axe. They had nothing to eat but hard dumplings. They waited for the nymph to go to church one Sunday and then fled. The nymph chased them but they threw a brush, a comb, and a mirror behind them which turned into impenetrable obstacles. \textcolor{red}{The nymph eventually gave up and the children escaped.}''

\subsection{Examples of failure cases for adversarial attacks}
\label{failure}
\noindent\textbf{Omission}
\begin{table}[H]
\fontsize{8pt}{8pt}\selectfont
    \begin{tabular}{p{\textwidth}}
    \toprule
    \textbf{Attack details}  \\
    In the story of Cinderella (``Aschenputtel'' in German), we omit the last part of the story where Cinderella attends the ball and meets the prince. \\
    \textbf{Generated outputs} \\
    ``Once upon a time, there was a wealthy man who lived happily with his wife and their daughter. When the wife became sick and died, she instructed her daughter to plant a tree on her grave and promised to help her if she shook the tree when in need. The man remarried and his new wife treated the daughter poorly, making her work in the kitchen and giving her the nickname "Cinderella". When the King held a ball to find a wife for his son, Cinderella's stepmother made her stay home. However, with the help of two doves, \textcolor{red}{Cinderella was able to attend the ball} and see her stepmother dancing with the Prince. The next day, the stepmother was angry that Cinderella had read the lentils and couldn't scold her.'' \\
    \bottomrule
    \end{tabular}
    \caption{Example of failure cases of omission attack. The failure is highlighted in \textcolor{red}{red}.}
    \label{res:example}
\end{table} 

\noindent\textbf{Entity swap}
\begin{table}[H]
\fontsize{8pt}{8pt}\selectfont
    \begin{tabular}{p{\textwidth}}
    \toprule
    \textbf{Attack details}  \\
    In this story, we replace (1) the well (which the siblings fall into) --> cave; (2) comb and brushes (with which the siblings use to create obstacles) --> pineapples and durian; (3) the axe (which the nymph fetches) --> butcher knife.  \\
    \textbf{Generated outputs} \\
    ``A brother and sister were playing near a \textcolor{green}{cave} and fell in. They were captured by a water nymph who made them work hard, giving the girl bad flax to spin and making the boy chop a tree with a blunt axe, while only feeding them hard dumplings. One Sunday, they managed to escape when the nymph went to church. The nymph chased them, but the children threw different \textcolor{green}{fruit with spiky exteriors} behind them, making it hard for the nymph to catch up. Finally, the girl threw a smooth mirror, forcing the nymph to go home for an \textcolor{red}{axe} to break through it. By the time she returned, the children had already escaped.'' \\
    \bottomrule
    \end{tabular}
    \caption{Example of failure cases of entity swap attack. The successful swap is heightened in \textcolor{green}{green} and the failure is highlighted in \textcolor{red}{red}.}
    \label{res:example_swap}
\end{table} 

\noindent\textbf{Negation}
\begin{table}[H]
\fontsize{8pt}{8pt}\selectfont
    \begin{tabular}{p{\textwidth}}
    \toprule
    \textbf{Attack details}  \\
    In the story of the frog prince (``Der Froschkönig oder der eiserne Heinrich'' in German), we negate the story by drastically changing the ending where the princess refuses to keep her promise and kills the frog. \\
    \textbf{Generated outputs} \\
    ``Once upon a time, a princess was playing with her golden ball when it fell into a deep well. She cried and wished for someone to retrieve it for her. A frog offered to help her in exchange for becoming her companion and eating from her plate and sleeping in her bed. The princess agreed and the frog retrieved the ball. However, when the frog came to claim his reward, the princess was scared and refused to keep her promise. The king then made her keep her word and \textcolor{red}{the frog turned into a prince}.'' \\
    \bottomrule
    \end{tabular}
    \caption{Example of failure cases of negation attack. The failure is highlighted in \textcolor{red}{red}.}
    \label{res:example_negation}
\end{table}

\end{document}